\documentclass[acmlarge,screen,authorversion,nonacm,review=false,timestamp=false]{acmart}

\usepackage{graphicx}
\usepackage{hyperref}
\usepackage{subcaption}
\usepackage{multirow} 
\newcommand{\splime}{SP-LIME}

\AtBeginDocument{%
  }

\setcopyright{acmlicensed}
\copyrightyear{2025}
\acmYear{2025}
\acmDOI{XXXXXXX.XXXXXXX}

\acmISBN{978-1-4503-XXXX-X/18/06}

\begin{document}

\title{Explanations as Bias Detectors: A Critical Study of Local Post-hoc XAI Methods for Fairness Exploration}

\author{Vasiliki Papanikou}
\affiliation{
  \institution{\textit{University of Ioannina}}
  \institution{\textit{Archimedes / Athena RC}}
  \country{Greece}
}
\email{v.papanikou@athenarc.gr}

\author{Danae Pla Karidi}
\affiliation{
  \institution{\textit{Archimedes / Athena RC}}
  \country{Greece}
}
\email{danae@athenarc.gr}

\author{Evaggelia Pitoura}
\affiliation{
  \institution{\textit{University of Ioannina}}
  \institution{\textit{Archimedes / Athena RC}}
  \country{Greece}
}
\email{pitoura@uoi.gr}

\author{Emmanouil Panagiotou}
\affiliation{
  \institution{\textit{Freie Universität Berlin}}
  \country{Germany}
}
\email{emmanouil.panagiotou@fu-berlin.de}

\author{Eirini Ntoutsi}
\affiliation{
  \institution{\textit{Universität der Bundeswehr München}}
  \country{Germany}
}
\email{eirini.ntoutsi@unibw.de}


\renewcommand{\shortauthors}{Papanikou et al.}


\begin{abstract}
As Artificial Intelligence (AI) is increasingly used in areas that significantly impact human lives, concerns about fairness and transparency have grown, especially regarding their impact on protected groups. 
Recently, the intersection of explainability and fairness has emerged as an important area to promote responsible AI systems. 
This paper explores how explainability methods can be leveraged to detect and interpret unfairness. We propose a pipeline that integrates local post-hoc explanation methods to derive fairness-related insights. 
During the pipeline design, we identify and address critical questions arising from the use of explanations as bias detectors such as the relationship between distributive and procedural fairness, the effect of removing the protected attribute, the consistency and quality of results across different explanation methods, the impact of various aggregation strategies of local explanations on group fairness evaluations, and the overall trustworthiness of explanations as bias detectors. Our results show the potential of explanation methods used for fairness while highlighting the need to carefully consider the aforementioned critical aspects.  
\end{abstract}

\ccsdesc[500]{Computing methodologies~Artificial intelligence}
\ccsdesc[500]{Information systems~Decision support systems}
\ccsdesc[500]{Human-centered computing~Interactive systems and tools}


\keywords{Explainable AI, algorithmic fairness}


\maketitle
\section{Introduction}
\label{sec:intro}
Artificial Intelligence (AI) is increasingly being deployed in critical areas that affect our daily lives. In the financial sector, AI plays a crucial role in evaluating credit scores or approving loans. In healthcare, it aids in diagnosing medical conditions, recommending treatment plans, and optimizing patient care management. Similarly, in education, it is reshaping processes like student admissions and personalizing learning experiences. As AI systems become increasingly embedded in such critical domains, concerns about fairness and transparency have grown, particularly regarding their effects on protected groups defined by gender, race, or other protected attributes. For example, studies have shown that many AI-driven hiring systems exhibit bias against women, reflecting historical inequalities~\cite{fabris2024fairness}. Similarly, the COMPAS system, used for recidivism prediction, has been found to assign higher risk scores to black defendants and lower risk scores to white defendants compared to their actual scores~\cite{larson2016we}, highlighting the potential for discriminatory outcomes.  

One major challenge in addressing these issues stems from the nature of AI systems themselves. Many are black-box models trained on vast and often poorly understood datasets. These datasets, collected from diverse sources, may encode historical biases~\cite{bolukbasi2016man}, data imbalances~\cite{le2022survey}, or spurious correlations~\cite{bhardwaj2024machine} that inadvertently propagate unfair outcomes. The combination of opaque model behavior and limited insight into the underlying data makes it difficult to trace the origins of biased decisions or to ensure fairness in AI predictions. Moreover, there is often a need to assess deployed models~\cite{casper2024black}, where fairness interventions are no longer feasible or practical. In such cases, auditing the model for biases becomes essential, both to understand its impact and to inform future improvements.

Explainable AI (XAI)~\cite{guidotti2018survey} has emerged as a critical tool for tackling these challenges, enabling transparency in model behavior and supporting fairness exploration~\cite{fragkathoulas2024explaining}. By shedding light on the ``decision mechanism" of AI systems, XAI facilitates the detection of biases and helps understand the relationships between protected attributes and target outcomes. In some studies, it has also been employed for bias mitigation by reducing the contribution of protected attributes to  decisions~\cite{kennedy2020contextualizing}.
This has led to growing attention on the intersection of explainability and fairness. However, despite its potential, the application of XAI to fairness has been questioned due to inconsistent terminology \cite{deck2024critical} and the absence of a standardized pipeline. Furthermore, XAI methods themselves rely on algorithmic processes that can inherit biases from the underlying data or models they aim to explain. This raises concerns about the reliability and trustworthiness of explanations, especially when biased data may lead to biased explanations~\cite{jain2020biased}. These challenges underscore the need for robust evaluation frameworks and systematic methodologies to ensure that XAI can effectively support fairness assessments.

To address these concerns, we propose a pipeline that integrates local post-hoc explanation methods to derive fairness-related insights. We identify and address critical questions arising during the design of such a pipeline, such as the relationship between distributive and procedural fairness, the effect of removing the protected attribute,
the consistency and quality of results across different explanation methods, the impact of various aggregation strategies of local
explanations on group fairness evaluations, and the overall trustworthiness of explanations as bias detectors. Our extensive empirical evaluation demonstrates the potential of explanations for bias detection and exploration. However, our study also highlights the important role of responsible evaluation and the need to carefully address the aforementioned critical aspects of the pipeline design.

The rest of the article is organized as follows: Section \ref{preliminaries} provides the necessary background,  Section \ref{motivation-methodology} introduces the proposed pipeline and critical design aspects formulated as research questions, in Section \ref{results} we present and discuss our experimental results, Section \ref{related_work} reviews related work on fairness and XAI, and Section \ref{conclusion} concludes the paper. 












\section{Preliminaries}
\label{preliminaries}
This section provides the necessary background on fairness definitions and explanation methods employed in our study.

 \textbf{Fairness metrics}
 Fairness in machine learning \cite{pitoura2022fairness, mehrabi2021survey, caton2024fairness, processfairness1, processfairness2} is often approached through two key approaches: distributive or statistical fairness~\cite{pitoura2022fairness, mehrabi2021survey, caton2024fairness} and procedural or process fairness \cite{processfairness1, processfairness2}.  \textit{Distributive fairness} focuses on the outcomes of models, while \textit{procedural fairness} evaluates the fairness of the decision-making process itself rather than just its outcomes~\cite{processfairness1, processfairness2}. The majority of fairness metrics in machine learning focus on distributive fairness, while procedural fairness remains relatively underexplored. 

The distributive fairness approaches can be further categorized into individual and group fairness. \emph{Individual fairness} requires that similar individuals are treated similarly, meaning receiving similar outcomes from the model. On the other hand, \emph{group fairness} assumes that individuals are partitioned into groups based on the value of one or more protected attributes and requires that these groups are treated similarly by the model. In this work, we focus on group fairness, which is commonly evaluated using three popular metrics: Demographic Parity, Equal Opportunity, and Equalized Odds. While DP focuses on the predictions of the model, Equal Opportunity and Equalized Odds focus on the errors of the model. We consider a fully supervised learning setting with two groups defined on the basis of some protected attribute(s), the protected group $G^+$ and the non-protected group $G^-$. Let also a binary classifier $f: \mathcal{X} \to \{0, 1\}$, where $1$ represents the favorable outcome. 
Let $y$ denote the ground truth label and $\hat{y}$ the predicted output of the classifier. 

\textit{Demographic Parity} requires that the positive prediction rates are similar across the groups, ensuring that the probability of an individual $v$ receiving a favorable outcome is independent of the group membership. Formally: 
\begin{equation}
P(\hat{y} = 1 \mid v \in G^+) = P(\hat{y} = 1 \mid v \in G^-)
\label{eq:demographic_parity}
\end{equation}

\textit{Equal Opportunity} requires the true positive rate (TPR) to be equal across groups. This ensures that individuals who truly belong to the favorable class are treated equitably, regardless of group membership. More formally:
\begin{equation}
P(\hat{y} = 1 \mid y = 1, v \in G^+) = P(\hat{y} = 1 \mid y = 1, v \in G^-)
\label{eq:equal_opportunity}
\end{equation}

 \textit{Equalized Odds} requires both the true positive rate (TPR) and the false positive rate (FPR) to be the same across the groups. More formally:
\begin{equation}
P(\hat{y} = 1 \mid y = 1, v \in G^+) = P(\hat{y} = 1 \mid y = 1, v \in G^-), 
 P(\hat{y} = 1 \mid y = 0, v \in G^+) = P(\hat{y} = 1 \mid y = 0, v \in G^-)
 \label{eq:equalized_odds}
\end{equation}

\textbf{Explanation Methods}
Explanations can be broadly categorized into: intrinsic, pre-process, and post-hoc types. \emph{Intrinsic explanations} come from models designed with built-in transparency. \emph{Pre-process explanations} use unsupervised techniques to uncover data patterns. \emph{Post-hoc explanations} are applied after model training  to clarify decision making~\cite{dwivedi2023explainable, arrieta2020explainable, adadi2018peeking, molnar2020interpretable, bodria2023benchmarking}. 
Post-hoc explanations are further divided into \emph{global explanations}, which explain the overall model logic, and \emph{local explanations}, which focus on individual predictions. In this study, we use local post-hoc explanations as we aim to analyze the behavior of the decision-making model on specific subpopulations defined by protected attributes, in contrast to global explanations that focus on the overall logic of the model. 
Among the variety of local explanation methods available, we selected the most widely used approaches, choosing one from each category: LIME \cite{ribeiro2016should}, an \emph{approximation-based method}, SHAP\cite{lundberg2017unified}, a \emph{feature-based method}, and DiCE \cite{mothilal2020explaining}, an \emph{example-based explanation method}.

\noindent\textit{LIME (Local Interpretable Model-agnostic Explanations)} \cite{ribeiro2016should} is a local, post-hoc, and model-agnostic that approximates black-box model predictions. The explanations include the contributions of the features to the prediction. LIME approximates the behavior of a complex black-box model around a specific instance using an interpretable surrogate model. To build this model, it generates a neighborhood around the instance to be explained by perturbing the features of the instance. Weights $\pi_x$ are assigned to neighborhood samples based on their proximity to the original instance. These samples are passed through the black-box model to obtain predicted labels. LIME then trains a surrogate model, such as linear regression or a decision tree, on the labeled weighted samples to approximate locally the black box. The optimization objective is: $\xi(x) = \arg\min_{g \in G} \, L(f, g, \pi_x) + \Omega(g)$, where $L(f, g, \pi_x)$ measures how well the surrogate model $g$ approximates the black-box model $f$, and $\Omega(g)$ controls the complexity of $g$. Finally, the prediction for the instance is explained by analyzing the feature contributions in the surrogate model.

\noindent\textit{SHAP (SHapley Additive exPlanations)} \cite{lundberg2017unified} is a feature-based explanation method that quantifies the contribution of each attribute to the final prediction of a model. Inspired by game theory, Shapley values are used to determine the contribution of each player in a cooperative game or coalition. In the context of machine learning, this concept is applied to evaluate how much each feature influences the prediction.

 \noindent\textit{DiCE (Diverse Counterfactual Explanations)} \cite{mothilal2020explaining} is a post hoc, example-based explanation method that generates counterfactual explanations. A counterfactual explanation identifies the smallest change to feature values that alters the prediction of a model. Given a prediction model $f$ and a data point $\mathbf{x}$, a counterfactual explanation provides an alternative point $\mathbf{x'}$, where the outcome $f(\mathbf{x'})$ differs from the initial prediction $f(\mathbf{x})$ while $x'$ remains similar to the original input $x$. DiCE focuses on two key aspects of generating counterfactuals: feasibility, ensuring that changes are realistic and actionable, and diversity, offering multiple plausible alternatives. 


\section{Motivation and Methodology}
\label{motivation-methodology}

\begin{figure}[H] 
\centering 
    \includegraphics[width=0.75\textwidth]{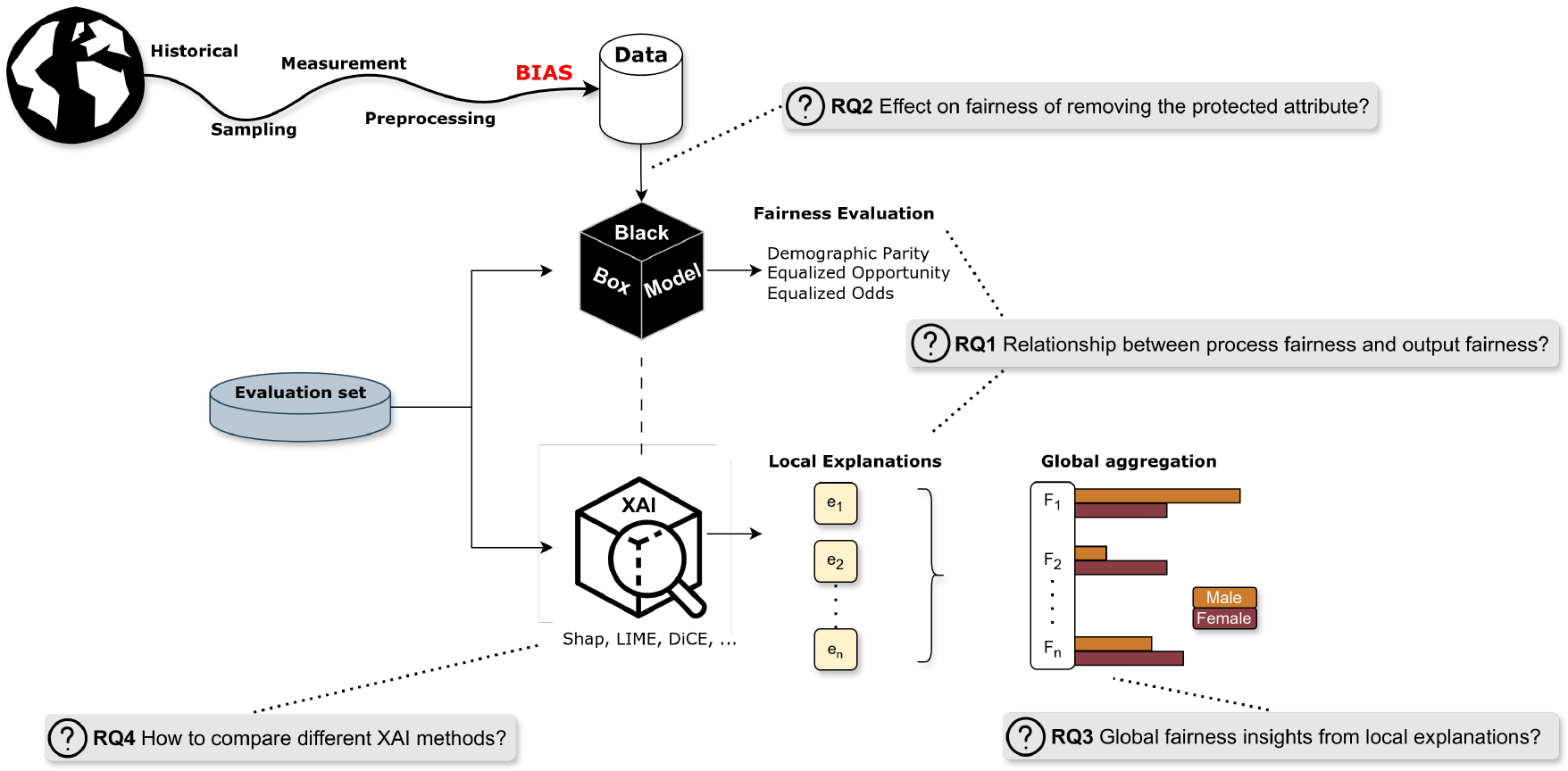}
\caption{An overview of the proposed pipeline integrating local post-hoc explanations and addressing key design/research questions for deriving fairness insights.} 
\label{fig:pipeline} 
\end{figure}

Since it is not clear how explanations can be effectively utilized for fairness analysis, we propose a general pipeline that integrates local post-hoc explanation methods to uncover fairness-related insights and identify potential biases in the decision-making process. Within this pipeline, we explore multiple questions regarding the appropriate application of each step for fairness analysis. As illustrated in Fig.~\ref{fig:pipeline}, given the output of a black-box model, we first apply a local post-hoc explanation technique to generate individual explanations for all instances within a specific group. Next, we use an aggregation method to derive the overall feature attributions for the group, representing how features contribute to the decision outcomes for that group. Finally, we compare these aggregated explanations across different demographic groups to identify potential disparities.

\noindent\textbf{RQ1. How does feature attribution relate to distributive fairness? What is the relationship, if any, between process fairness and output fairness?}

Procedural or process fairness focuses on the fairness of the decision-making process by examining the input features used in the model. Deciding which features are appropriate to be used involves considering various factors, including legal and ethical considerations. These factors include feature volitionality, which examines whether a feature reflects voluntary choices made by an individual or is influenced by circumstances beyond their control, feature reliability, which assesses the accuracy and consistency of feature measurement, feature privacy, which considers whether the inclusion of certain features violates the privacy rights of individuals, and feature relevance, which evaluates whether a feature is causally linked to the decision outcome \cite{processfairness2}. Additionally, legal issues must be considered, as highlighted in \cite{grabowicz2022marrying}. Unfair influence refers to the illegal impact of protected features on decisions, whereas a fair relationship describes a legally permissible association between protected and non-protected features. Current procedural fairness assessments rely heavily on subjective human judgment, which may overlook complex feature interactions. In this study, we apply post-hoc local XAI methods, aggregating results by demographic group to analyze feature contributions and improve fairness evaluation.

Our first research question examines how procedural fairness relates to distributive fairness by exploring the relationship between feature attribution provided by XAI and fairness metrics derived from model outcomes. We aim to examine whether the insights from distributive fairness metrics can also be detected through feature attribution. For instance, if an explanation method reveals that the decisions of a model heavily depend on protected attributes or features strongly correlated with them, do distributive fairness metrics also signal bias or unfairness?

To explore the relationship between distributive and procedural fairness, the basic idea is to compare the outcomes derived from Demographic Parity, Equalized Opportunity, and Equalized Odds with insights about feature attribution gained from procedural fairness results. 
To assess procedural fairness, we create local post-hoc explanations using the methods LIME, SHAP and DiCE. 
For LIME and SHAP, we generate explanations as follows:
\begin{itemize}
    \item For Demographic Parity, which focuses on all positive instances (Eq.~\ref{eq:demographic_parity}), we generate explanations for all positive instances within each protected group
    \item For Equalized Opportunity, which focuses on True Positives (Eq.~\ref{eq:equal_opportunity}), we generate explanations by analyzing True Positives (TP)
    \item{For Equalized Odds, which focuses on both True Positives and False Positives (Eq.~\ref{eq:equalized_odds}), we generate explanations by analyzing both True Positives (TP) and False Positives (FP)}
\end{itemize}
In contrast, for the DiCE model as it identifies the minimal changes required for an instance to alter its classification, so we generate explanations as follows:
\begin{itemize}
    \item For Demographic Parity (Eq.~\ref{eq:demographic_parity}), we generate counterfactuals for instances in the negative class to uncover the features that would require the most significant adjustments for each group to transition to the positive class
    \item For Equalized Opportunity (Eq.~\ref{eq:equal_opportunity}), we generate counterfactuals for False Negatives (FN) to identify the features preventing positive outcomes
    \item For Equalized Odds (Eq.~\ref{eq:equalized_odds}), we generate counterfactuals for both False Negatives (FN) and True Negatives (TN)
\end{itemize}

\noindent\textbf{RQ2. What is the effect of removing the protected attribute in fairness? How does this relate to direct (dependency on the protected attribute) vs indirect discrimination (dependency on proxy attributes)?}

Discrimination can occur in two forms: direct discrimination, where individuals are treated unfairly based on their membership in a protected class (disparate treatment), and indirect discrimination, where members of a protected class are negatively affected even if their membership in that class is not explicitly used (disparate impact). While removing the protected attribute might seem like a straightforward way to mitigate direct discrimination, it does not necessarily eliminate unfair outcomes. Instead, it can lead to induced discrimination, where the absence of the protected attribute increases the influence of correlated proxy features. A critical question arising from the use of explanation methods to identify procedural unfairness is whether we can identify indirect discrimination using explanation methods, even when the model does not explicitly use protected attributes. Using explanations after removing the protected attribute allows for evaluating the result of this removal on procedural fairness. Users can see what are the features that are used in decision making and whether their use is considered fair.

To explore this, we remove the protected attribute from the model and examine the resulting changes in both distributive and procedural fairness. We compare the accuracy and distributive fairness of the model before and after the removal of the protected attribute, enabling an examination of the trade-offs between performance and fairness. Using explanation methods, we then analyze how the contributions of different features change following the removal of the protected attribute. This allows us to observe whether the influence of sensitive features is redirected to other features. Specifically, we aim to identify whether high contributions are retained by features unrelated to the task, features that act as proxies for the protected attribute, or features strongly correlated with it. So the related questions we are going to explore are: Can the redistribution of the protected attribute contribution be identified through explanation methods? Furthermore, does the observed shift in feature contributions correspond to changes in distributive fairness?

\noindent\textbf{RQ3. Can aggregated individual explanations provide global fairness insights?}


In our experimental study, we use local post-hoc explanations as we focus on specific demographic groups. Since local explanations provide individual insights, there is a need to aggregate them. In our case, we aggregate the explanations according to demographic groups. This raises the question of what is the most appropriate method for aggregating individual explanations in a way that helps draw conclusions about fairness. Several aggregation methods exist in the literature. For example, the \splime~\cite{ribeiro2016should} algorithm for LIME explanations selects a representative set of explanations with features that have high global importance, defined as the square root of the sum of absolute attributions. This method assumes that frequently appearing features with high local attributions are globally significant. However, this approach is not ideal for our case, as we aim to analyze all instances within each group rather than a selected subset.

Different aggregation methods influence the fairness analysis by either amplifying or smoothing out disparities in feature importance across groups. For LIME and SHAP methods, we selected and applied two different techniques. For a feature $f_j$  we computed the aggregated importance $I_{\text{abs}}(f_j)$ as the mean of the absolute sum of contributions $c_{ij}$: $I_{\text{abs}}(f_j) = \frac{1}{N} \sum_{i=1}^{N} \left| c_{ij} \right|$. This method sums the magnitude of feature attributions across instances, highlighting features that significantly contribute to model decisions. While this approach identifies disproportionately influential features, it may obscure differences in positive and negative contributions across groups, potentially hiding oppositional biases (e.g., a feature benefiting one group while harming another). Another option we use is summing the contributions without taking the absolute values: $I_{\text{abs}}(f_j) = \frac{1}{N} \sum_{i=1}^{N}  c_{ij}$. In this case, by preserving the signs of the contributions, we can identify whether the contributions of each group tend toward negative values, indicating that the feature pushes the group toward the unfavorable class, or it pushes the group toward the favorable class.

Aggregating individual counterfactual explanations across different subgroups of a sensitive attribute poses challenges. One way to aggregate counterfactuals is by counting the number of changed features. If a particular group consistently requires more modifications to achieve the same prediction outcome, it suggests that the decision boundary of the model is more rigid for that subgroup. An alternative approach is to measure the magnitude of feature changes, capturing the extent to which the model expects individuals to alter their attributes. Significant shifts in key features, such as a large increase in income for loan approval, may reveal potential bias affecting specific groups. Another method we explore is the Burden metric, which is defined as follows for all instances $N$ of the group: $\text{Burden} = \frac{1}{N} \sum_{i=1}^{N} c(x_i, x_i') $, for some distance metric $c$  such as the Euclidean distance.

\noindent\textbf{RQ4. How do different explanation methods compare in terms of robustness, consistency and explanation quality, and can they be trusted?}

While explainability in AI has achieved significant success in analyzing model behavior, it has also faced criticism regarding its robustness, consistency, and trustworthiness \cite{rawal2021recent}. Recent studies suggest that while many existing methods provide reliable local explanations, these can be misleading when attempting to derive a global understanding of models \cite{mittelstadt2019explaining}. Additionally, other research highlights vulnerabilities in XAI methods, such as SHAP and LIME, which can be exploited to conceal biases present in models \cite{slack2020fooling}. Similarly, counterfactual explanations are susceptible to manipulation \cite{slack2021counterfactual}, allowing small perturbations to produce outcomes that unfairly favor certain subgroups. This has led to a growing emphasis on incorporating \emph{robustness} as a critical criterion for counterfactual generation \cite{guyomard2023generating}. Furthermore, challenges also emerge at the data level. For example, gradient-based counterfactual generation methods may exhibit feature-type bias, favoring changes in continuous features over discrete ones \cite{panagiotou2024tabcf}. Given these challenges, it is critical to approach the interpretation, testing, and validation of explanations with care. 

Ensuring the reliability and fairness of explainability methods requires rigorous evaluation and a comprehensive understanding of their limitations, enabling more trustworthy and actionable insights. However, explainable AI methods lack ground truth, making it challenging to evaluate their performance and compare them effectively. This remains a significant open problem in the XAI field. Recent studies have highlighted this issue, demonstrating that the same method can have different results depending on factors such as data normalization and the reference values used during evaluation \cite{koenen2024toward}. One well-established metric to evaluate feature attribution methods, mainly in the textual domain \cite{local-global}, but also used for tabular data \cite{hameed2022based} is the Area Under the Pertubation Curve (AOPC). This metric sequentially perturbs (removes) features according to their ranked (global) importance and measures the resulting impact on model performance. While AOPC has faced criticism regarding the practical implementation of feature removal \cite{hameed2022based} and concerns about generating out-of-distribution samples during the perturbation process \cite{hase2021out}, it remains a valuable tool for evaluation due to its interpretability and simplicity. Therefore, we employ this metric in our experiments to compare and evaluate the global feature rankings outputted by the XAI methods.

\section{Experimental Evaluation}
\label{results}
In this section, we describe the experimental setup and discuss our results. 
\subsection{Experimental setup}
For our experiments, we used the \textit{Adult}\footnote{\href{https://archive.ics.uci.edu/dataset/2/adult}{Adult dataset: \url{https://archive.ics.uci.edu/dataset/2/adult}}}
 dataset, a common benchmark for fairness studies derived from the 1994 US Census survey. This dataset contains the demographic characteristics of 48,842 individuals and is used to predict whether their annual income exceeds \$50,000. The dataset consists of 14 attributes and a target variable. We followed the preprocessing of related work \cite{zhang2018mitigating, kamiran2012data} and excluded some attributes. 
Since the Adult dataset is relatively old and there has been criticism regarding its suitability for fairness evaluation \cite{ding2021retiring}, we also incorporated more recent datasets derived from US Census surveys, the so-called \textit{{ACS PUMS dataset}}~\cite{ding2021retiring}. We chose to utilize data from 2023, as it is the most recent available. To explore potential cultural and political differences, we chose datasets from California (AdultCA) and Louisiana (AdultLA), with 203,278 and 20,970 instances, respectively. 
Further details about the attributes of the datasets can be found in Appendix \ref{appendix}.

The protected attributes considered in this study are \texttt{sex} and \texttt{race}. The \texttt{sex} attribute is restricted to two categories \{male, female\} as they are the only ones available in the dataset. Although the \texttt{race} attribute includes additional racial groups, we focus on two of them \{Black, White\} due to limited sample sizes of other groups.

We train a Random Forest classifier using scikit-learn library \cite{scikit-learn}, to act as a Black-box. For LIME \footnote{\href{https://github.com/marcotcr/lime/tree/master}{LIME: https://github.com/marcotcr/lime/tree/master}}, we use the default parameter settings. For SHAP \footnote{\href{https://github.com/shap/shap}{SHAP: https://github.com/shap/shap}}, we employ the exact explainer version. For DiCE \footnote{\href{https://github.com/interpretml/DiCE}{DiCE: https://github.com/interpretml/DiCE}}, we allow counterfactuals to modify all features while setting the diversity parameter to zero, therefore producing a single counterfactual.

\begin{table}
  \centering
  \small
  \caption{Differences in PR, TPR, and FPR for \texttt{sex} (male-female) and \texttt{race} (White-Black) across datasets, incl. statistical significance scores. Higher values indicate worse disparities.}
\label{tab:fairness_diff_sex_race}
  \begin{tabular}{lcccccc}
    \toprule
    Metric & \multicolumn{2}{c}{Adult} & \multicolumn{2}{c}{AdultCA} & \multicolumn{2}{c}{AdultLA} \\
    \cmidrule(r){2-3} \cmidrule(r){4-5} \cmidrule(r){6-7}
           & Gender & Race & Gender & Race & Gender & Race \\
    \midrule
    PR     & 0.165 (15.91) & 0.099 (6.01)  & 0.108 (21.88) & 0.137 (10.80) & 0.256 (17.06) & 0.261 (13.51) \\
    TPR    & 0.093 (2.65)  & 0.112 (1.84)  & 0.040 (7.27)  & 0.055 (4.20)  & 0.195 (8.73)  &0.207 (6.4)  \\
    FPR    & 0.084 (10.26) & 0.025 (2.0)   & 0.064 (11.51) & 0.074 (4.65)  & 0.143 (9)  & 0.13 (7.29)   \\
    \midrule
    Accuracy & \multicolumn{2}{c}{0.82} & \multicolumn{2}{c}{0.807} & \multicolumn{2}{c}{0.769} \\
    \bottomrule
  \end{tabular}
\end{table}

We generate explanations for a representative number of instances in each demographic (sub)group. Namely, for the Adult and AdultCA datasets, we generate explanations for 100 instances per demographic group across all outcome categories (P, TP, FP, N, TN, FN). For the AdultLA dataset, due to a smaller sample size, we generate explanations for 50 instances for every category.

\begin{figure}[H]
    \centering
    \includegraphics[width=0.65\linewidth]{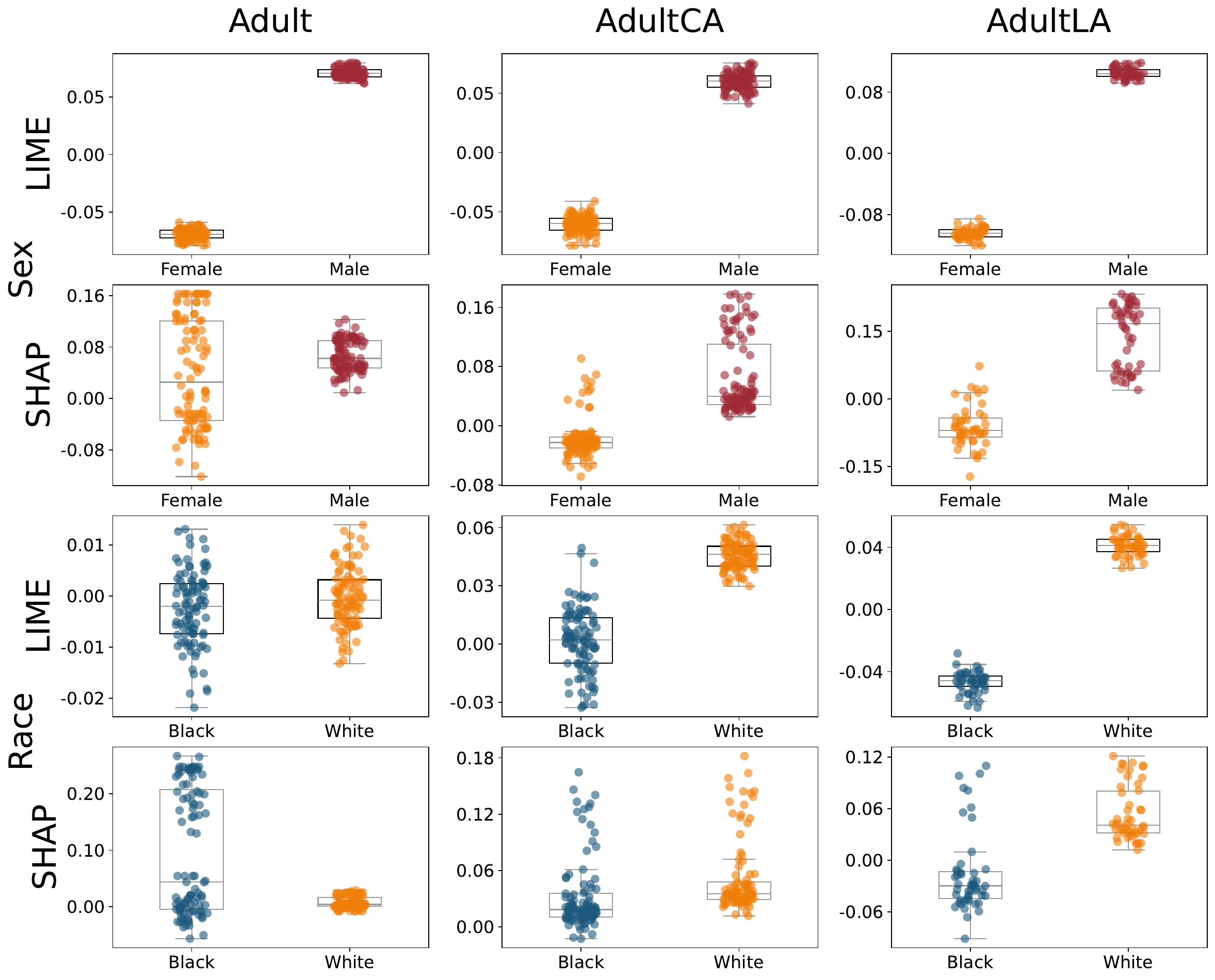}
    \caption{LIME and SHAP feature contributions for \texttt{sex} and \texttt{race} across datasets.}
    \label{fig:contributions_distributions}
\end{figure}


\subsection{Results for RQ1: How does feature attribution relate to distributive fairness? What is the relationship, if any, between process fairness and output fairness?}
 Table \ref{tab:fairness_diff_sex_race} shows the differences in PR, TPR, and FPR between male and female groups and White and Black groups, with statistical z-tests for significance.  We observe that all distributive fairness metrics are violated across all datasets, as indicated by significant differences with high z-test values (where higher values imply greater statistical significance). The largest disparities are found in the AdultLA dataset, which is consistent with the conservative nature of this state.

Next, we look at procedural fairness by studying the contribution of the protected attribute to the positive class. We first report on LIME and SHAP.  Fig.~\ref{fig:contributions_distributions} illustrates the distribution of contributions for the \texttt{sex} attribute across male and female groups and White and Black for positive instances. 
For LIME, we observe that the contribution of \texttt{sex} is consistently negative for females and positive for males across all datasets. This indicates that \texttt{sex} drives males towards favorable outcomes while it pushes females toward unfavorable outcomes. The difference in contributions among groups is more pronounced in the AdultLA dataset. For SHAP, this phenomenon is most evident in the AdultCA and AdultLA datasets, where the contributions for \texttt{sex} are more strongly polarized. For \texttt{race}, the disparity is again more pronounced in the AdultLA dataset. Overall, in most cases, the contribution of the protected attribute tends to favor non-protected groups (males, White). The results are consistent across datasets for \texttt{sex} but show slight variations for \texttt{race}. Disparities are more evident with the LIME method, especially in the AdultLA dataset, aligning with its major distributive fairness violations.
 
\begin{table*}[htbp]
\centering
\small
\renewcommand{\arraystretch}{1.5} 
\caption{Difference in mean contributions for \texttt{sex} between males and females. For LIME and SHAP we report on differences w.r.t. P, TP, FP. For DiCe we report on the feature change percent differences w.r.t. N, FN and TN.}
\label{tab:contributions_dif_sex}
\vspace{-1em}
\begin{tabular}{p{1cm}|p{0.52cm}|p{0.52cm}|p{0.52cm}|p{0.52cm}|p{0.52cm}|p{0.52cm}|p{0.52cm}|p{0.52cm}|p{0.4cm}|p{0.4cm}|p{0.4cm}|p{0.4cm}}
\hline
\multirow{2}{*}{Dataset} & \multicolumn{3}{c|}{\textbf{LIME}} & \multicolumn{3}{c|}{\textbf{SHAP}} & \multicolumn{3}{c|}{\textbf{DiCE}} \\
\cline{2-10}
& P & TP & FP & P & TP & FP & N & FN & TN \\
\hline

Adult & 0.132 & 0.131 & 0.132 & 0.014 & 0.032 & 0.004 & 43 & 47 &  41\\
\hline
AdultCA & 0.116 & 0.117 & 0.115 & 0.076 & 0.075 & 0.1 & 10 & 14 & 4 \\
\hline
AdultLA & 0.216 & 0.214 & 0.216 & 0.194 & 0.182 & 0.225 & 16 & 2 & 2 \\
\hline
\end{tabular}
\end{table*}

Next, we utilize the explanation method for the rest of the distributive fairness metrics by explaining TP and FP using LIME and SHAP, and accordingly for DiCE by explaining N, FN, and TN. Table \ref{tab:contributions_dif_sex} presents the differences in the contribution for \texttt{sex}. A similar table for the \texttt{race} attribute can be found in the Appendix \ref{appendix}. The differences are calculated by subtracting the contribution of the non-protected group from that of the protected group for the LIME and SHAP methods. In contrast, for the DiCE method, the subtraction is reversed, as we aim to observe the additional feature changes required by the protected group. We observe that the contribution of the protected attribute remains consistent across P, TP, and FP when using LIME, with only slight variations when using SHAP. In AdultLA we observe pronounced differences with both methods. 

\begin{figure}[H]
    \centering
    \includegraphics[width=\linewidth]{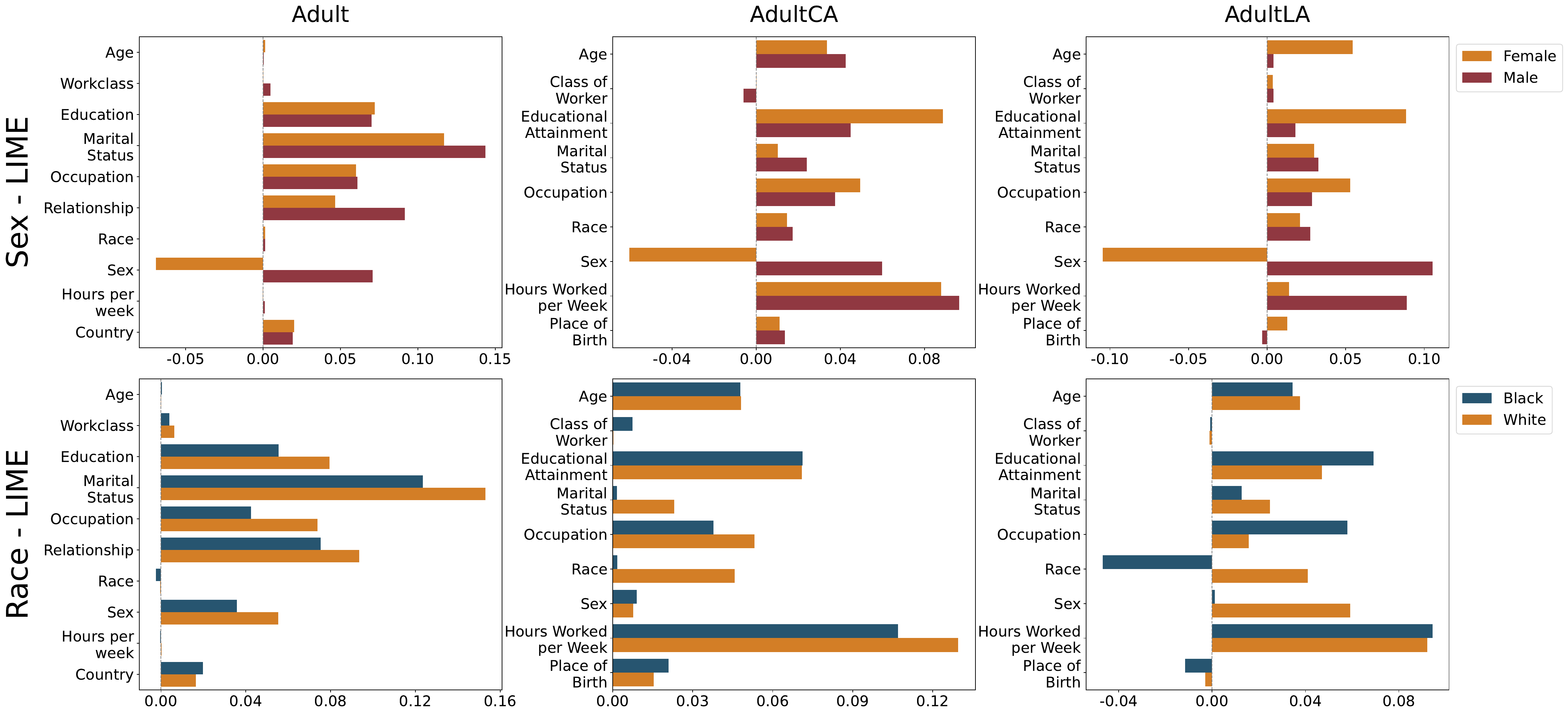}
     \caption{LIME mean feature contributions for \texttt{sex} and \texttt{race} across datasets.}
    \label{label:mean_contr_LIME}
\end{figure}

\begin{figure}[H]
    \centering
    \includegraphics[width=\linewidth]{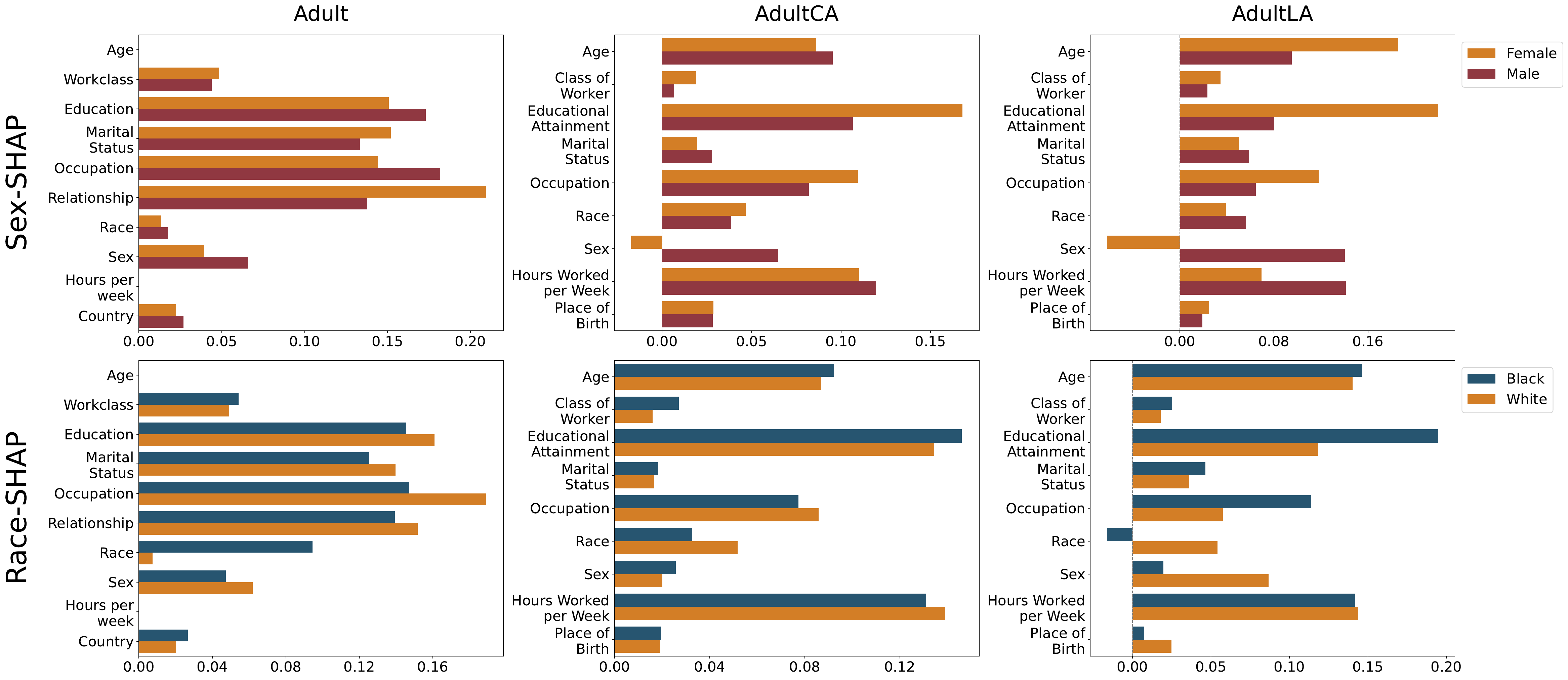} 
     \caption{Mean contributions for features in the Adult, AdultCA, and AdultLA datasets.}
    \label{label:mean_contrSHAP}
\end{figure}

\begin{figure}[H]
        \begin{minipage}{0.32\textwidth}
        \centering
        \includegraphics[width=\linewidth]{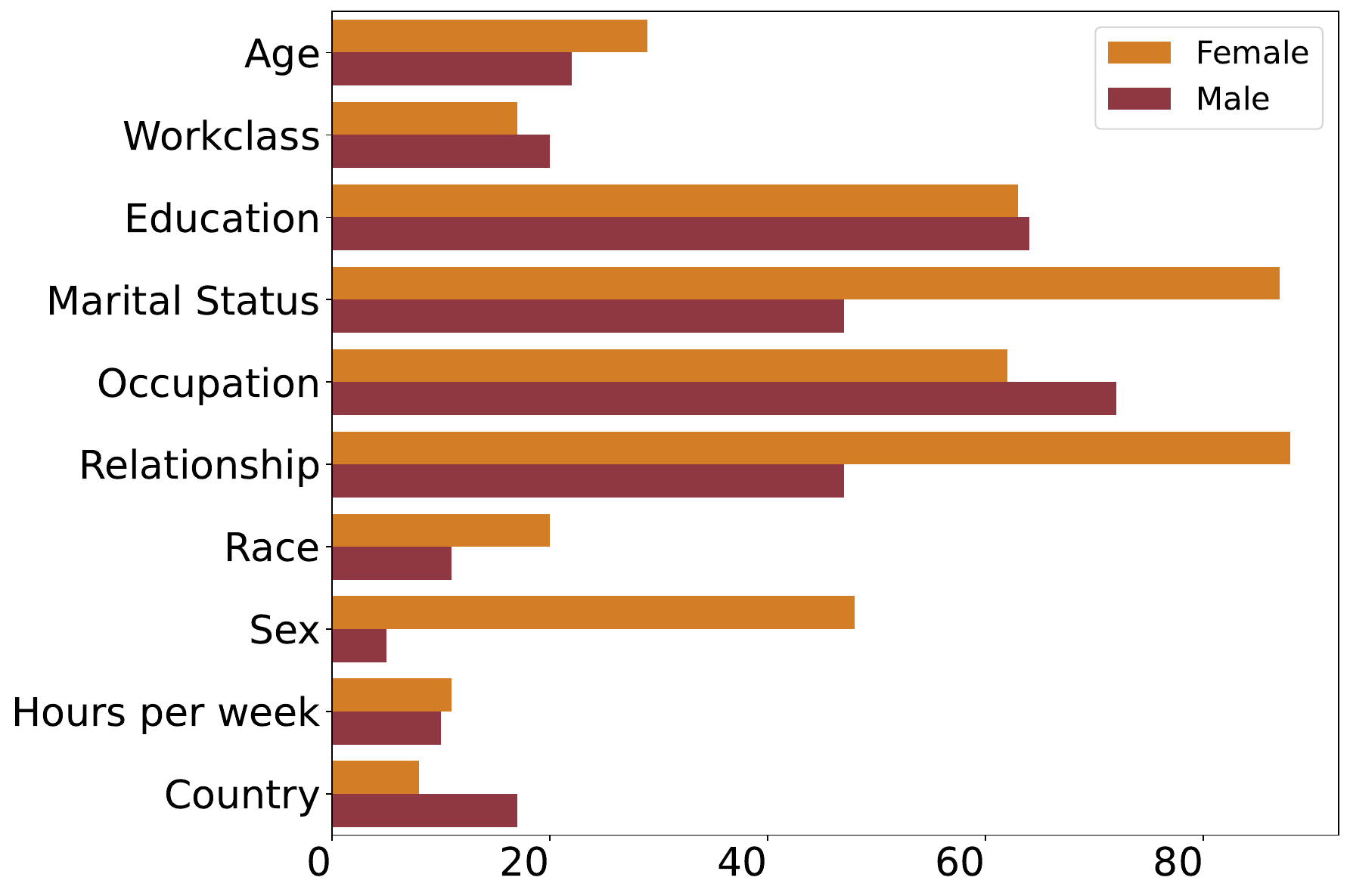}
    \end{minipage}
    \hfill
    \begin{minipage}{0.32\textwidth}
        \centering
        \includegraphics[width=\linewidth]{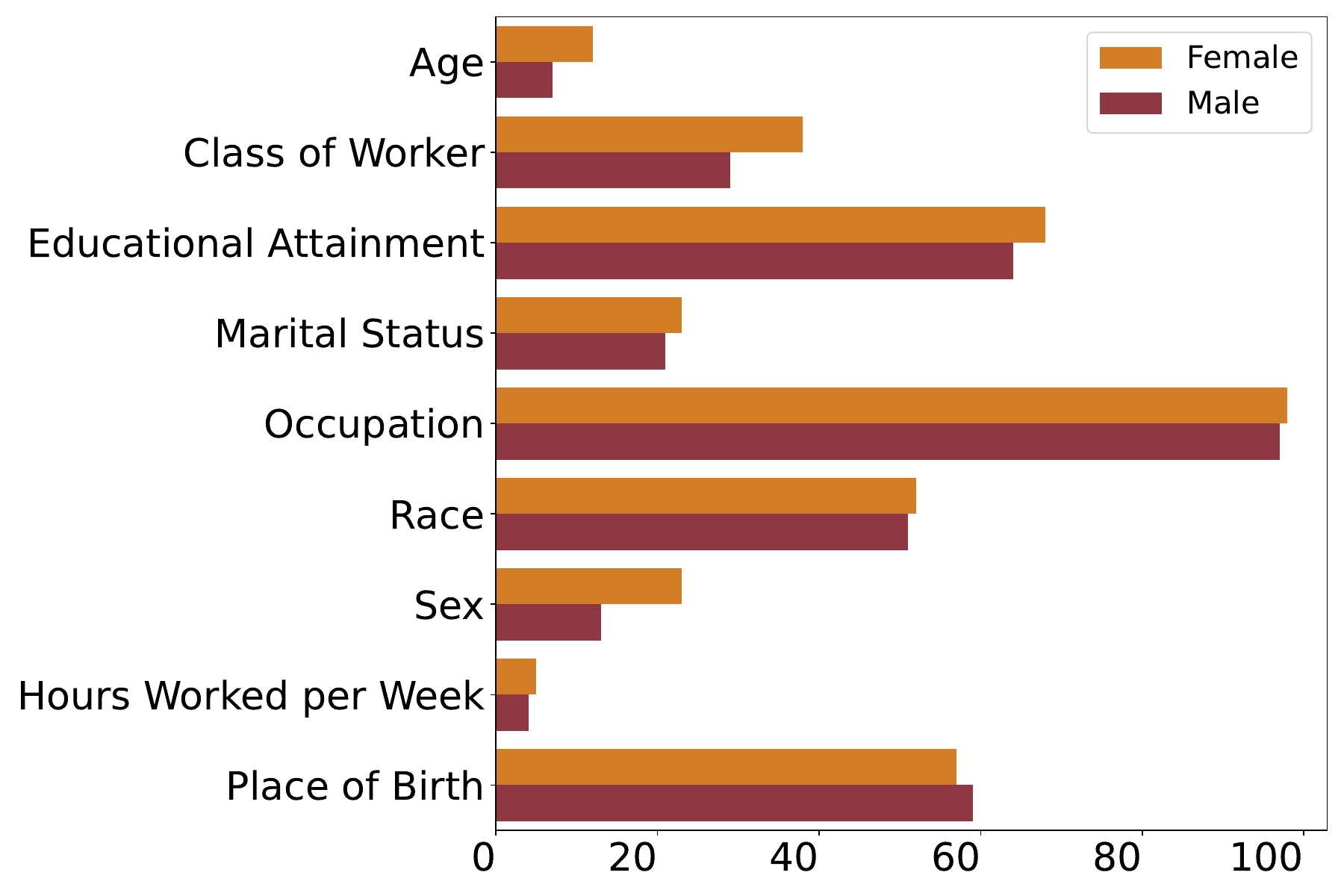}      
    \end{minipage}
    \hfill
    \begin{minipage}{0.32\textwidth}
        \centering
        \includegraphics[width=\linewidth]{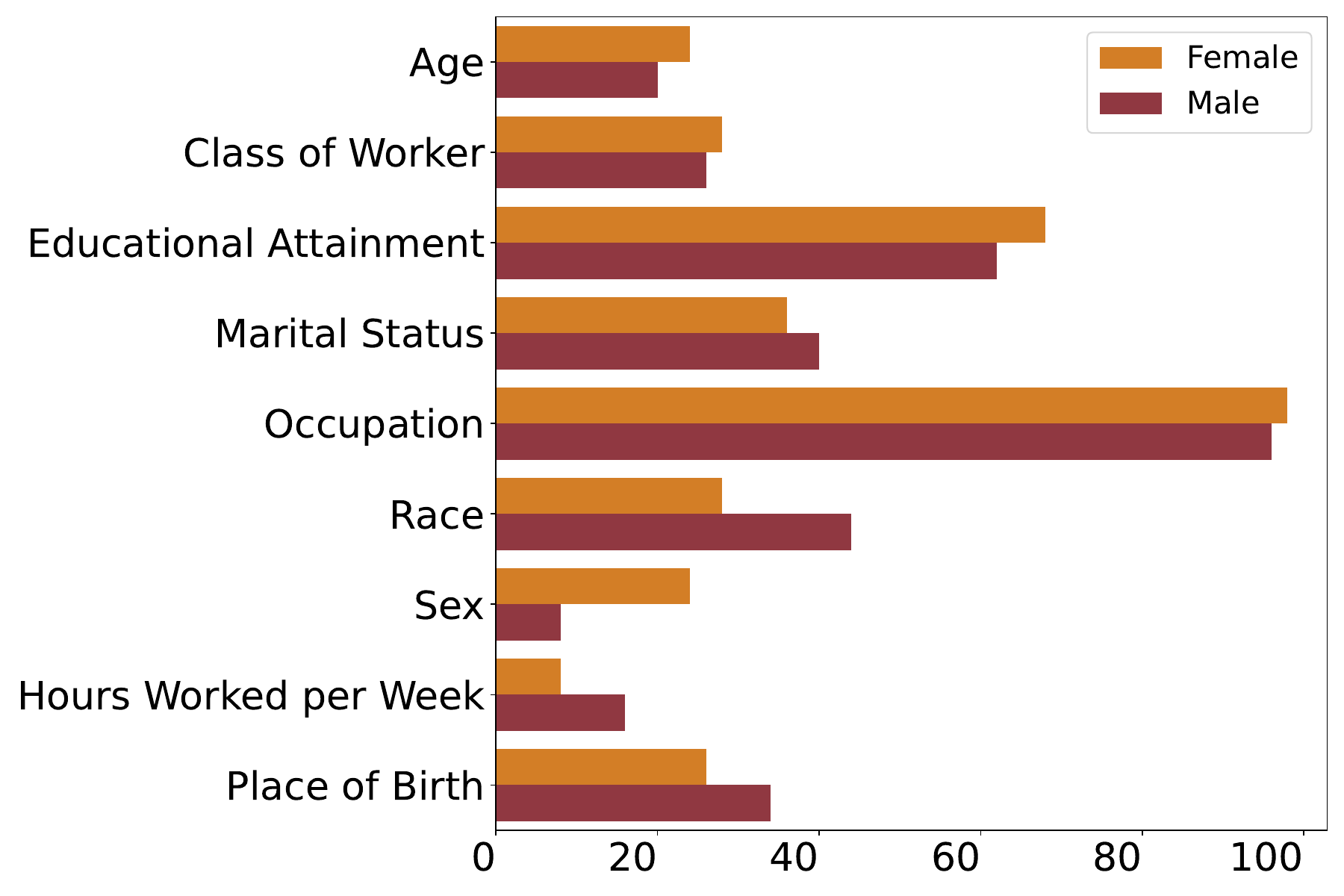}
         
    \end{minipage}
   \hfill
     \begin{minipage}{0.32\textwidth}
        \centering
        \includegraphics[width=\linewidth]{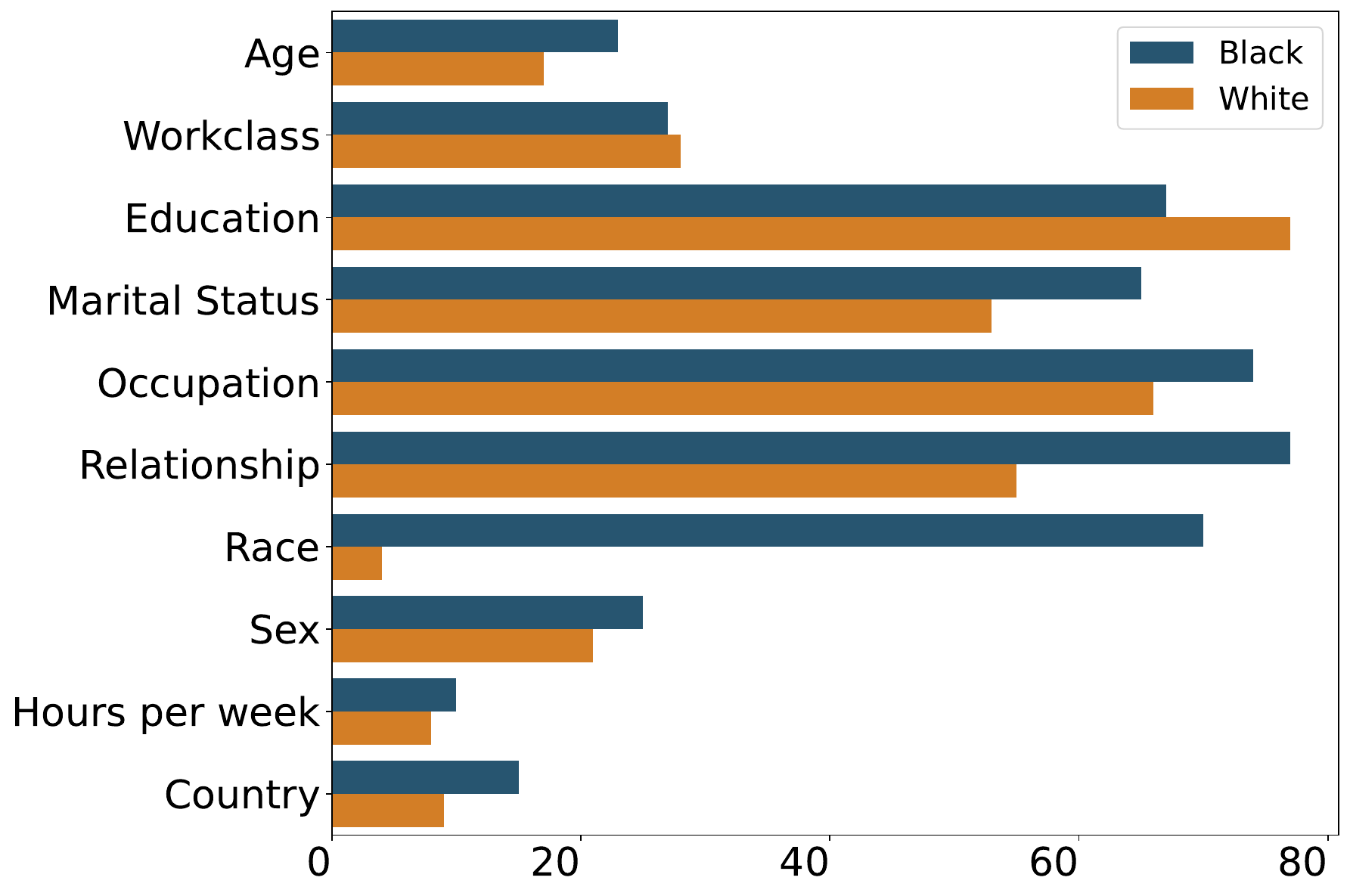}
    \end{minipage}
    \hfill
    \begin{minipage}{0.32\textwidth}
        \centering
        \includegraphics[width=\linewidth]{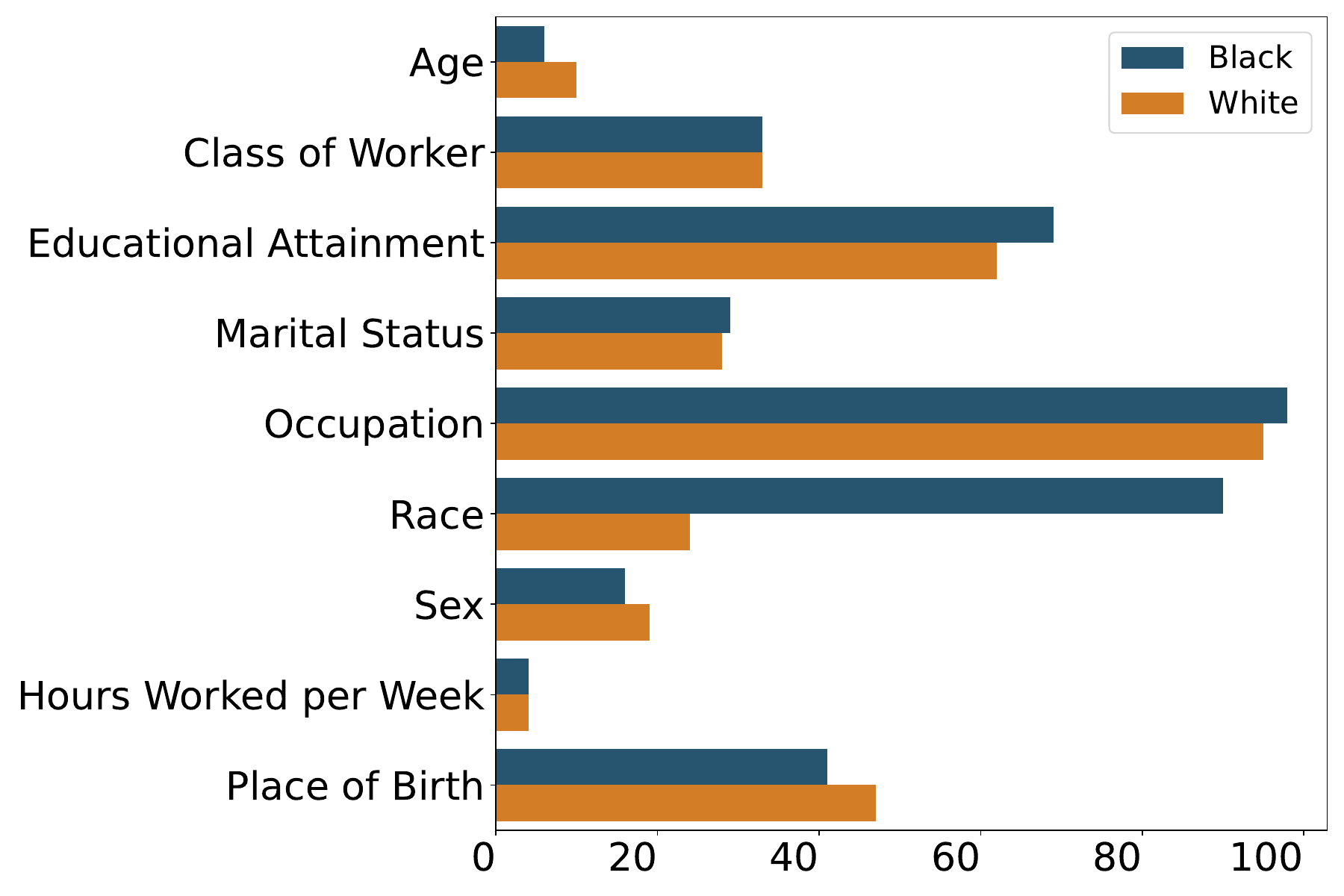} 
    \end{minipage}
    \hfill
    \begin{minipage}{0.32\textwidth}
        \centering
        \includegraphics[width=\linewidth]{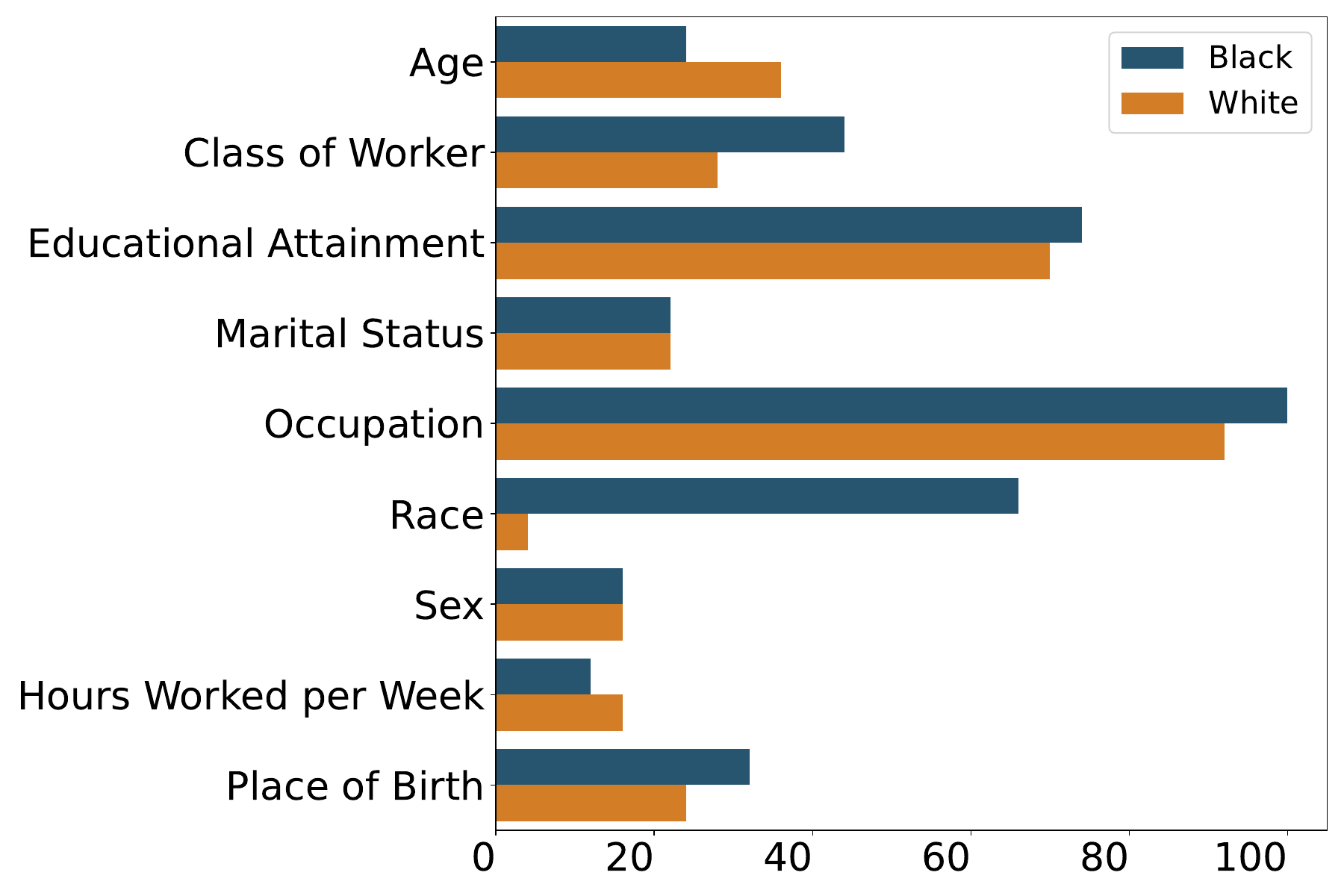}
    \end{minipage}
    \caption{Percentage of feature changes from the DiCE method per group for the Adult, AdultCA, and AdultLA datasets.}
    \label{fig:DICE}
\end{figure}
Finally, we look at other feature contributions. We aggregate individual feature explanations by calculating the mean, allowing us to observe both positive and negative impacts on each group. Alternatively, using the mean absolute contributions provides insight into the magnitude of each feature influence. In Figures \ref{label:mean_contr_LIME}, \ref{label:mean_contrSHAP}, we observe that the most significant features include features not directly related to the task, such as marital status, relationship, \texttt{sex}, age, and \texttt{race} and that most of these features tend to benefit non-protected groups more. For both methods, \texttt{sex} is included in the features with the highest contributions across all datasets, with a consistent negative contribution for the female group. Only in California, we observe that the most significant feature is directly related to the task, namely Hours worked per week and this is consistent both per \texttt{sex} and per \texttt{race} with both LIME and SHAP methods.

In Fig.~\ref{fig:DICE}, we present the results for the DiCE explanation method. Since DiCE is a counterfactual approach, we calculate and compare the percentage of feature changes required across the different groups. Across all datasets, we observe that the female and Black groups need to change more features than their privileged counterparts. In Adult, females frequently need to change features such as \texttt{sex}, relationship, and marital status in almost every instance to achieve the desired outcome. Notably, in the contributions per \texttt{race}, \texttt{race} attribute is consistently one of the top attributes across all datasets. This indicates that individuals from the Black group are disproportionately required to change their \texttt{race} to achieve a favorable outcome.

\begin{table}[H]
  \centering
  \small
  \caption{Differences in PR, TPR, and FPR between gender and racial groups across datasets after removing protected attributes}
   \label{tab:fairness_diff_no_sex_race}
  \begin{tabular}{lcccccc}
    \toprule
    Metric & \multicolumn{2}{c}{Adult} & \multicolumn{2}{c}{AdultCA} & \multicolumn{2}{c}{AdulLA} \\
    \cmidrule(r){2-3} \cmidrule(r){4-5} \cmidrule(r){6-7}
           & Gender & Race & Gender & Race & Gender & Race \\
    \midrule
    PR     & 0.152 (14.82) & 0.102 (6.25)  & 0.064 (13.07) & 0.124 (9.72)  & 0.129 (8.51) & 0.155 (8.03) \\
    TPR    & 0.063 (1.81)  & 0.016 (0.26)  & 0.001 (0.19)  & 0.051 (3.79)  & 0.043 (1.94)  &0.041(1.25)   \\
    FPR    & 0.075 (9.29)  & 0.042 (3.4)   & 0.017 (3.04)  & 0.053 (3.38)  & 0.029 (1.79)  & 0.05 (3.13)   \\
    \midrule
    Accuracy & 0.81 & 0.81 & 0.78 & 0.79 & 0.75 & 0.76 \\
    \bottomrule
  \end{tabular}
\end{table}

\begin{figure}[H] 
\centering 
    \includegraphics[width=0.85\textwidth]{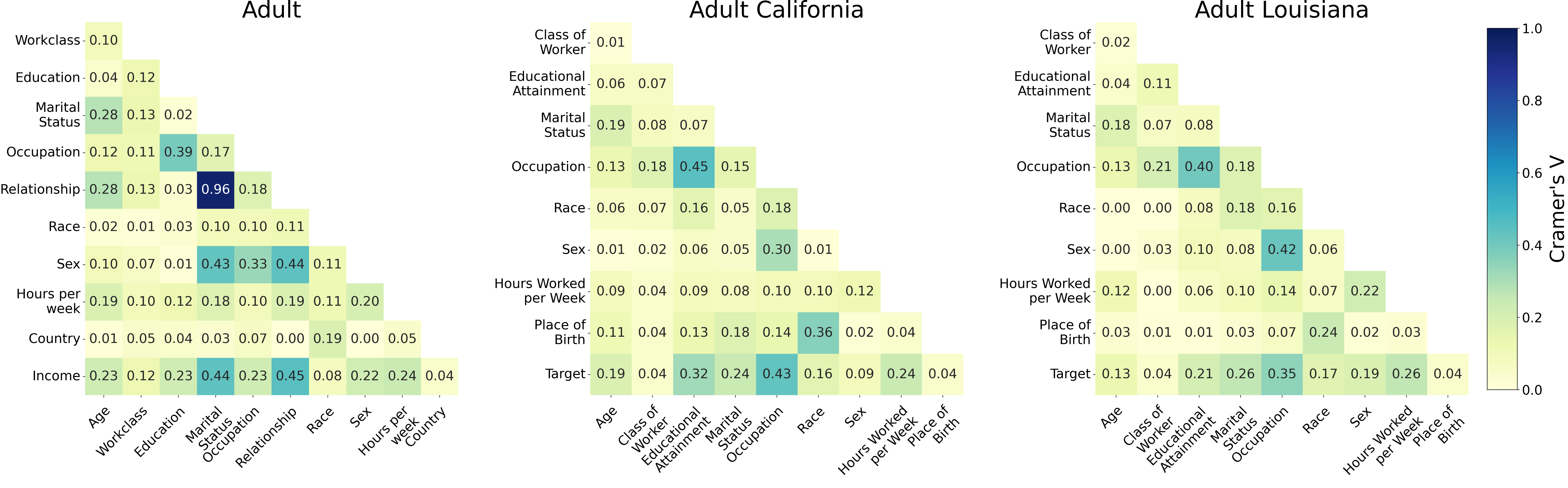}
\caption{Correlation matrices for the Adult, AdultCA, and AdultLA datasets.} 
\label{fig:cor_matrices} 
\end{figure}

\subsection{Results for RQ2: What is the impact of removing protected attributes on fairness, and how does it relate to direct vs. indirect discrimination}

\begin{figure}[H]
    \centering

    \begin{subfigure}[b]{0.32\textwidth}
        \includegraphics[width=\linewidth]{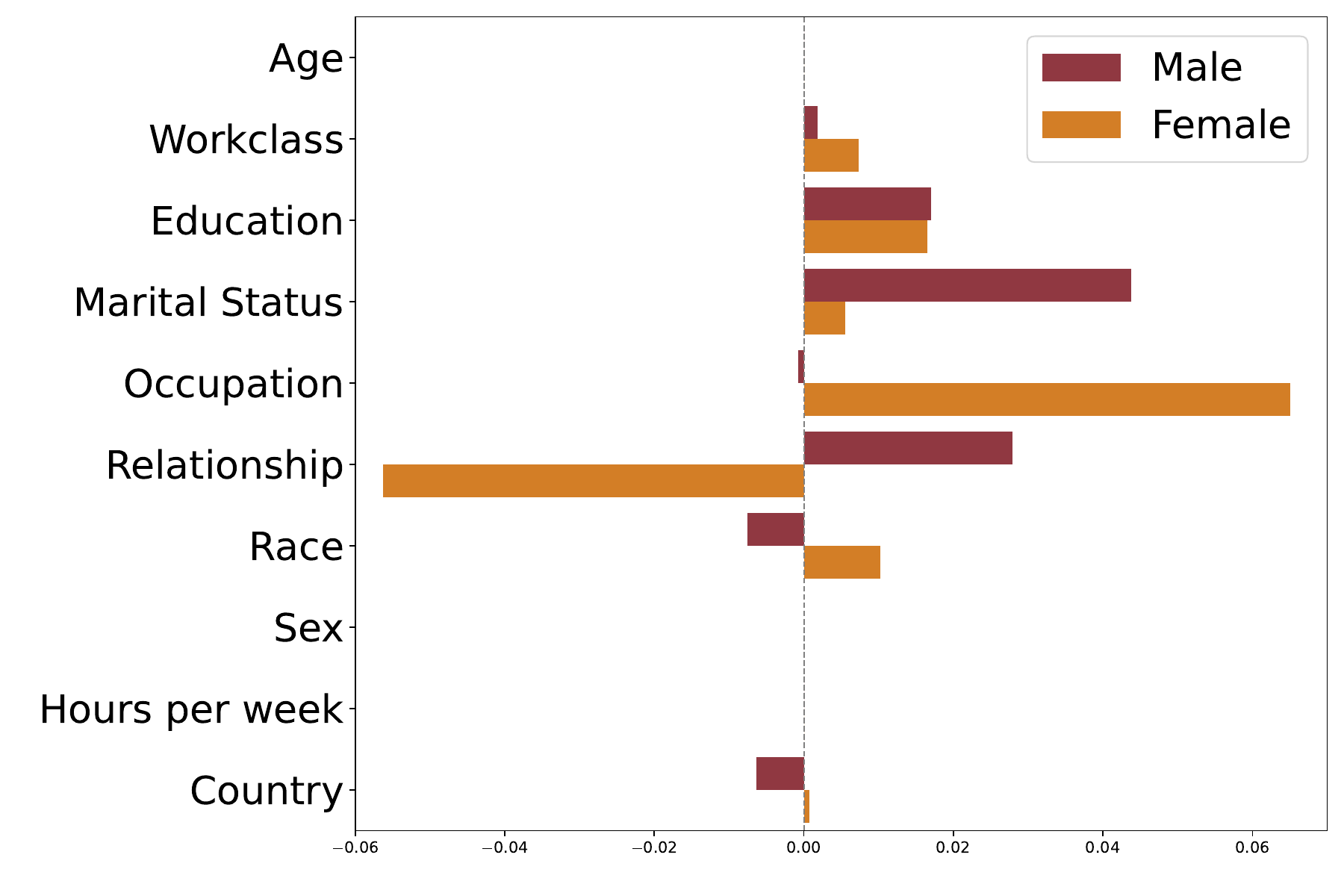}
    \end{subfigure}
    \begin{subfigure}[b]{0.32\textwidth}
        \includegraphics[width=\linewidth]{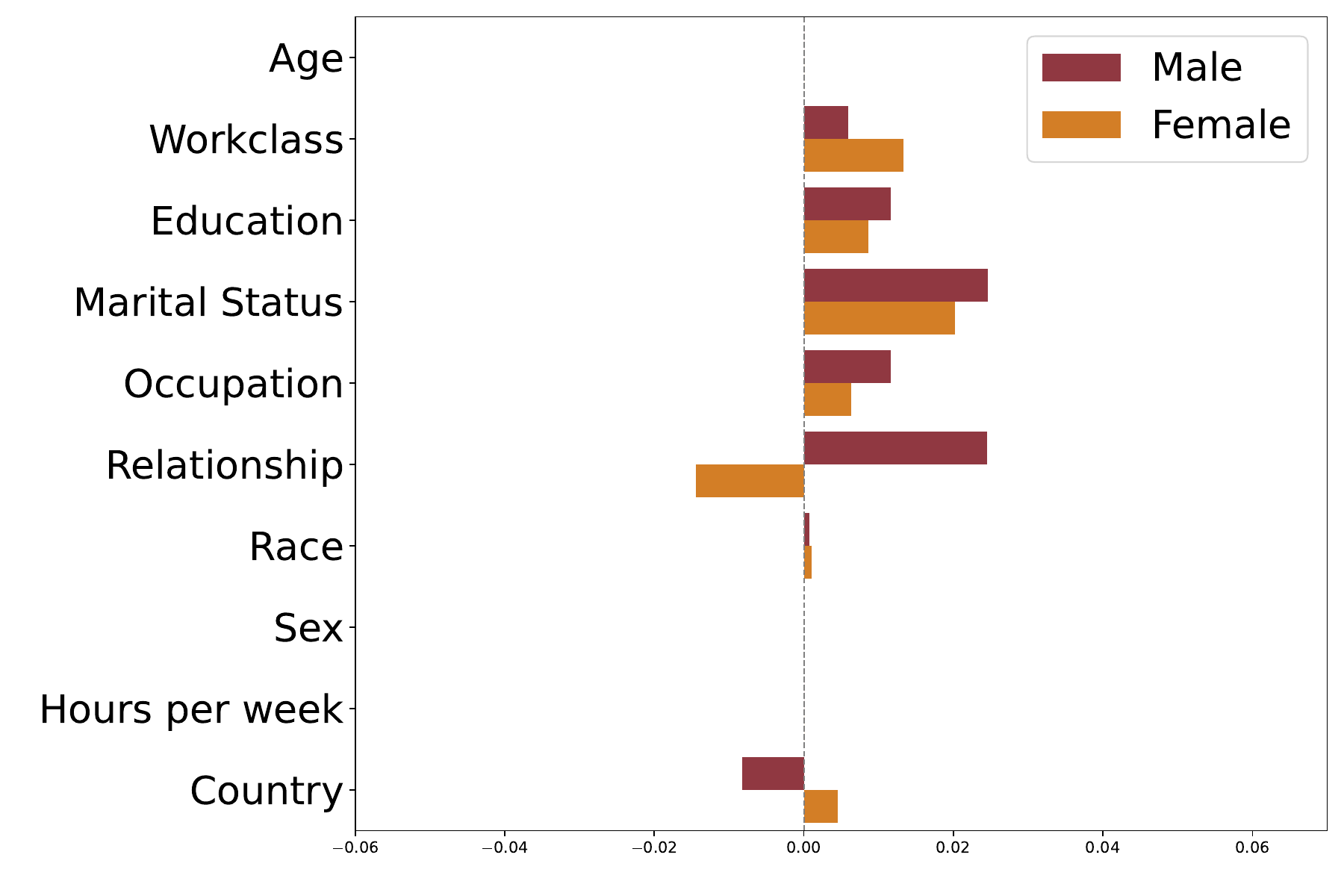}
    \end{subfigure}
    \begin{subfigure}[b]{0.32\textwidth}
        \includegraphics[width=\linewidth]{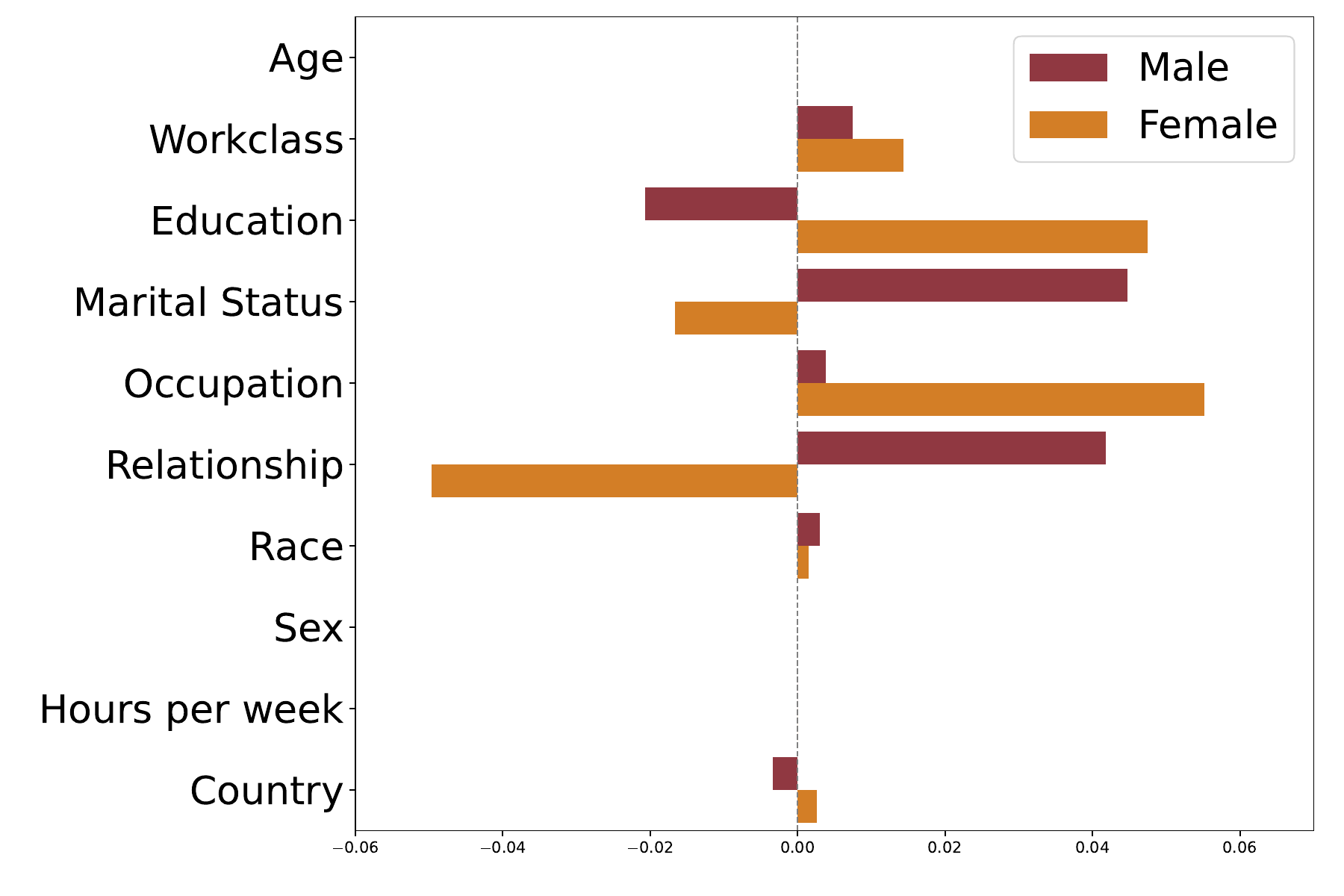}
    \end{subfigure}
    \begin{subfigure}[b]{0.32\textwidth}
        \includegraphics[width=\linewidth]{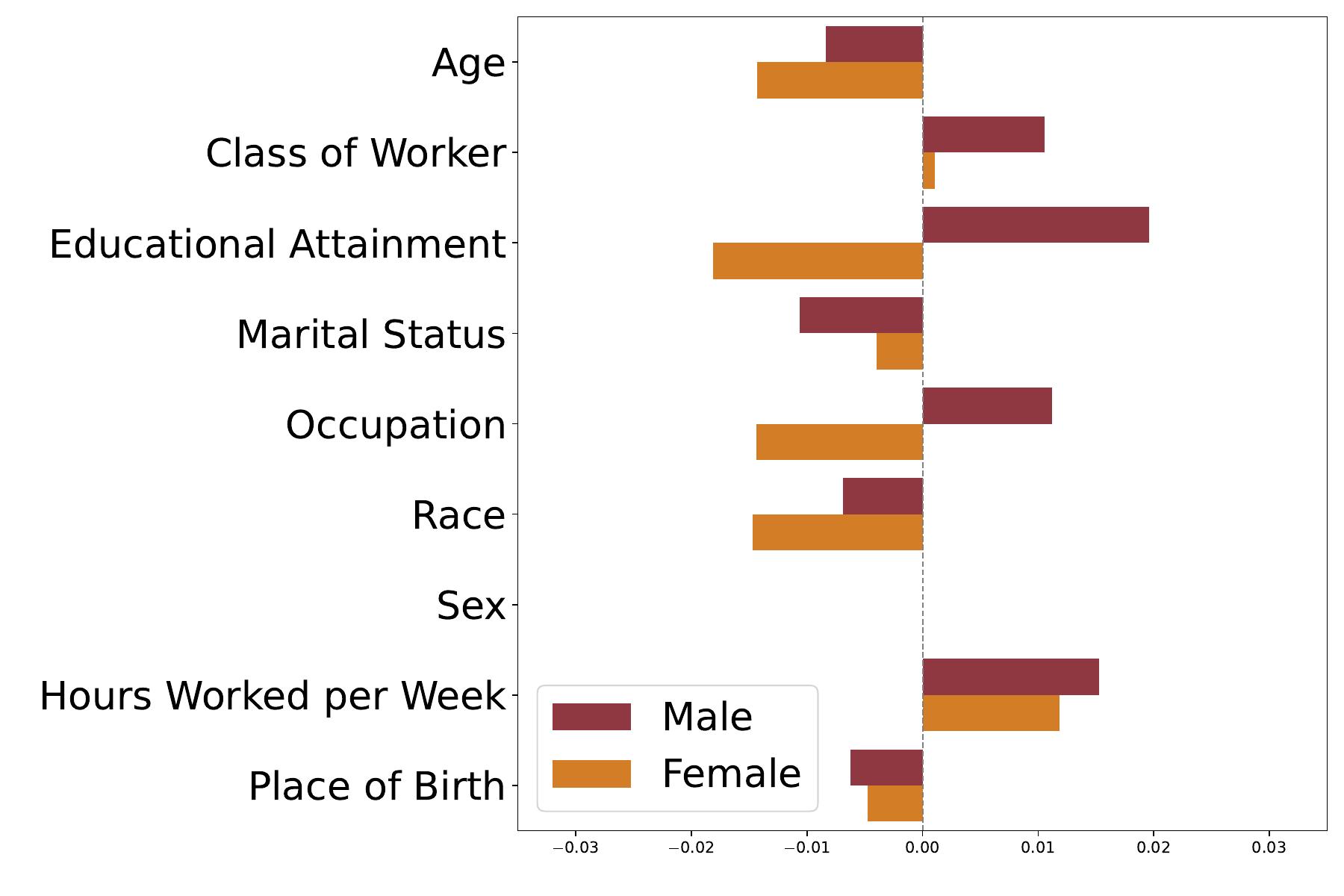}
    \end{subfigure}
    \begin{subfigure}[b]{0.32\textwidth}
        \includegraphics[width=\linewidth]{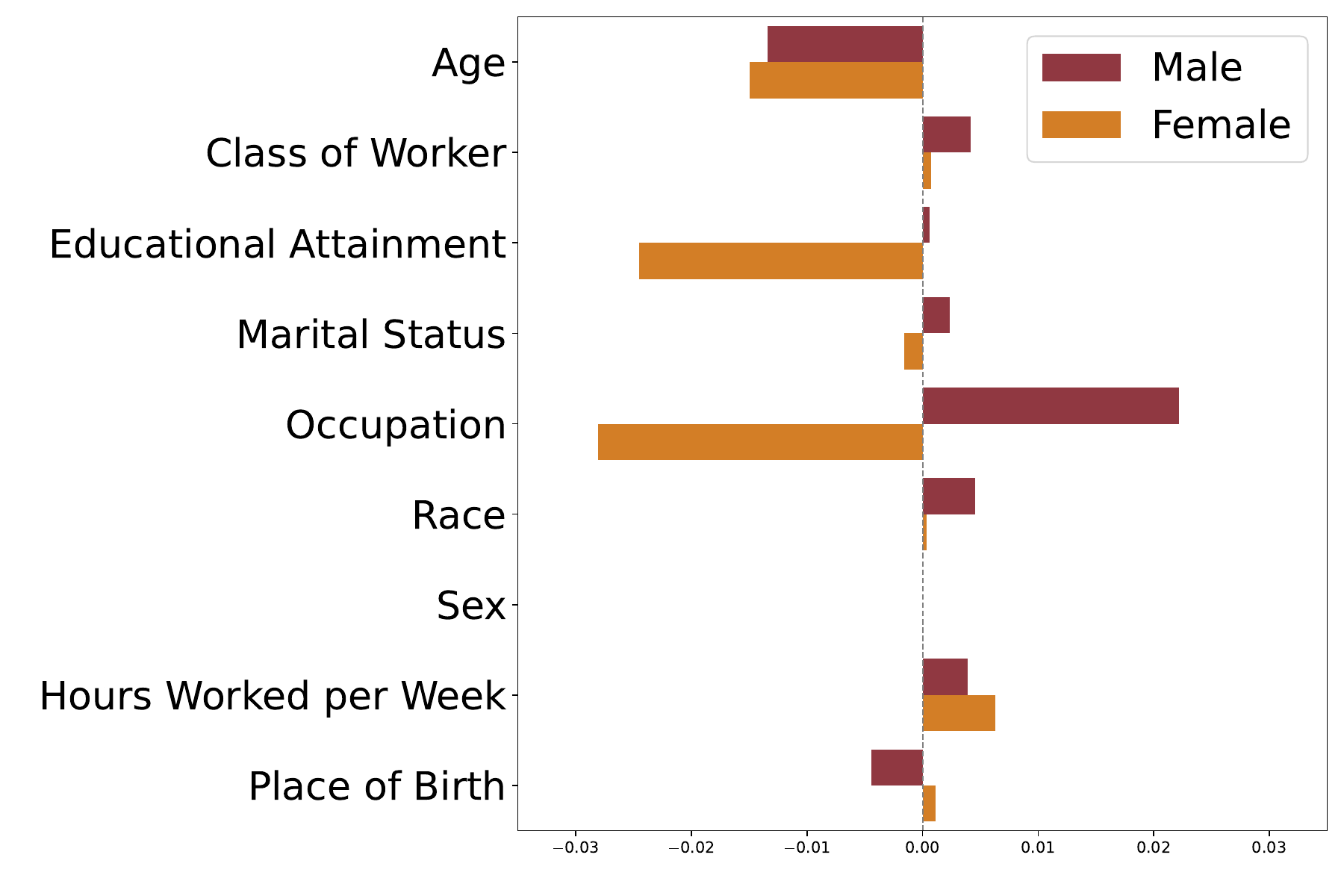} 
    \end{subfigure}
    \begin{subfigure}[b]{0.32\textwidth}
        \includegraphics[width=\linewidth]{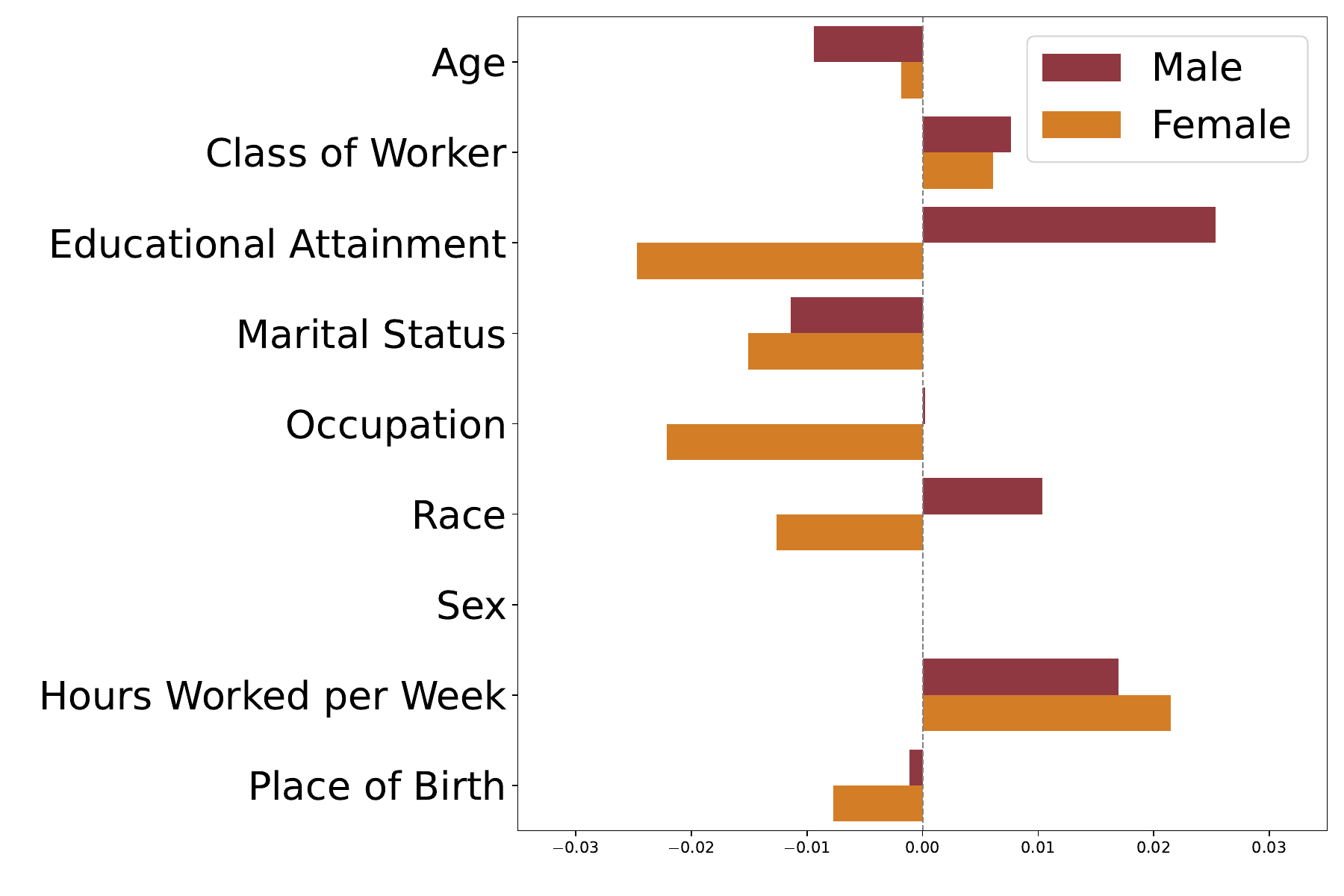}
    \end{subfigure}
    \begin{subfigure}[b]{0.32\textwidth}
        \includegraphics[width=\linewidth]{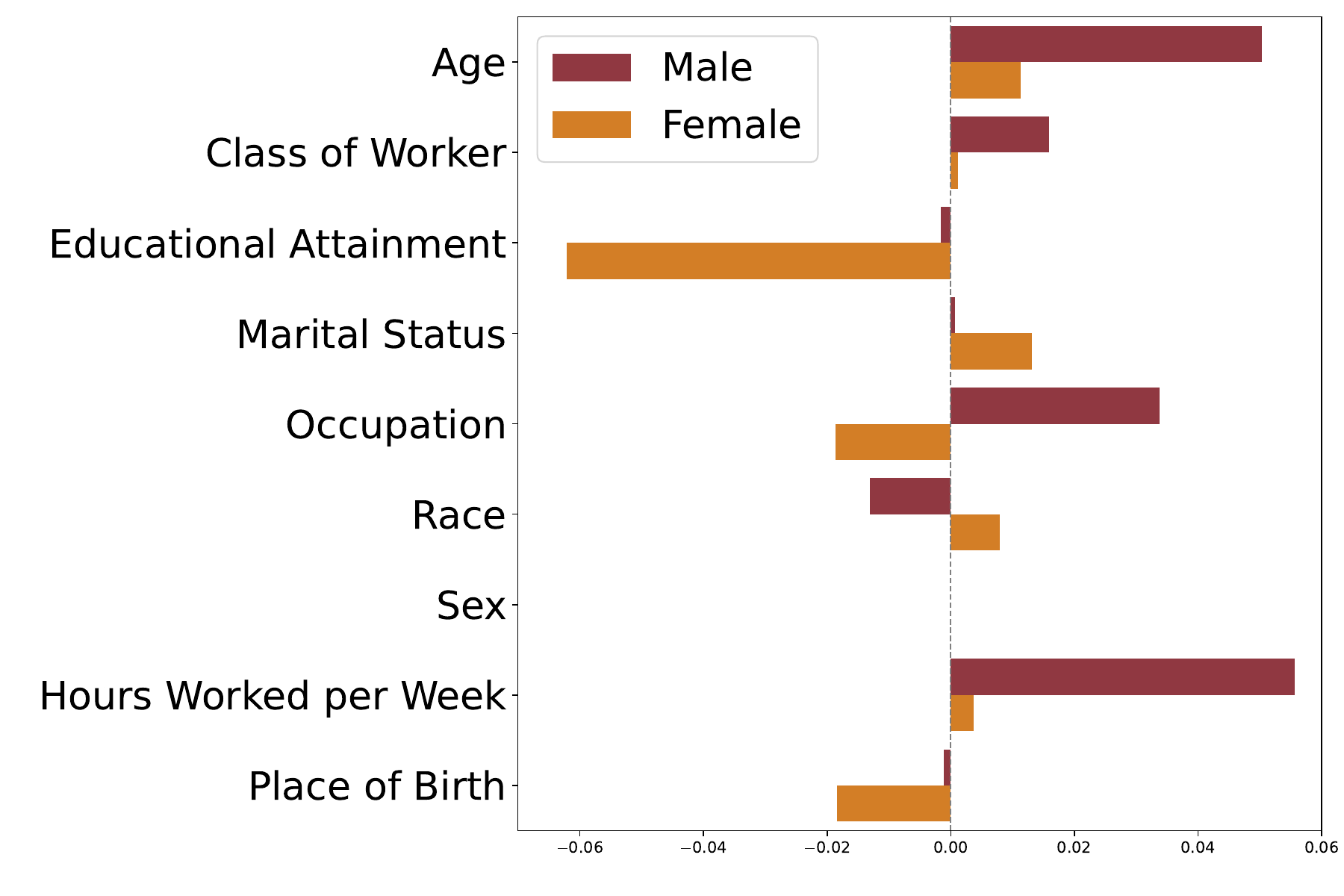} 
    \end{subfigure}
    \begin{subfigure}[b]{0.32\textwidth}
        \includegraphics[width=\linewidth]{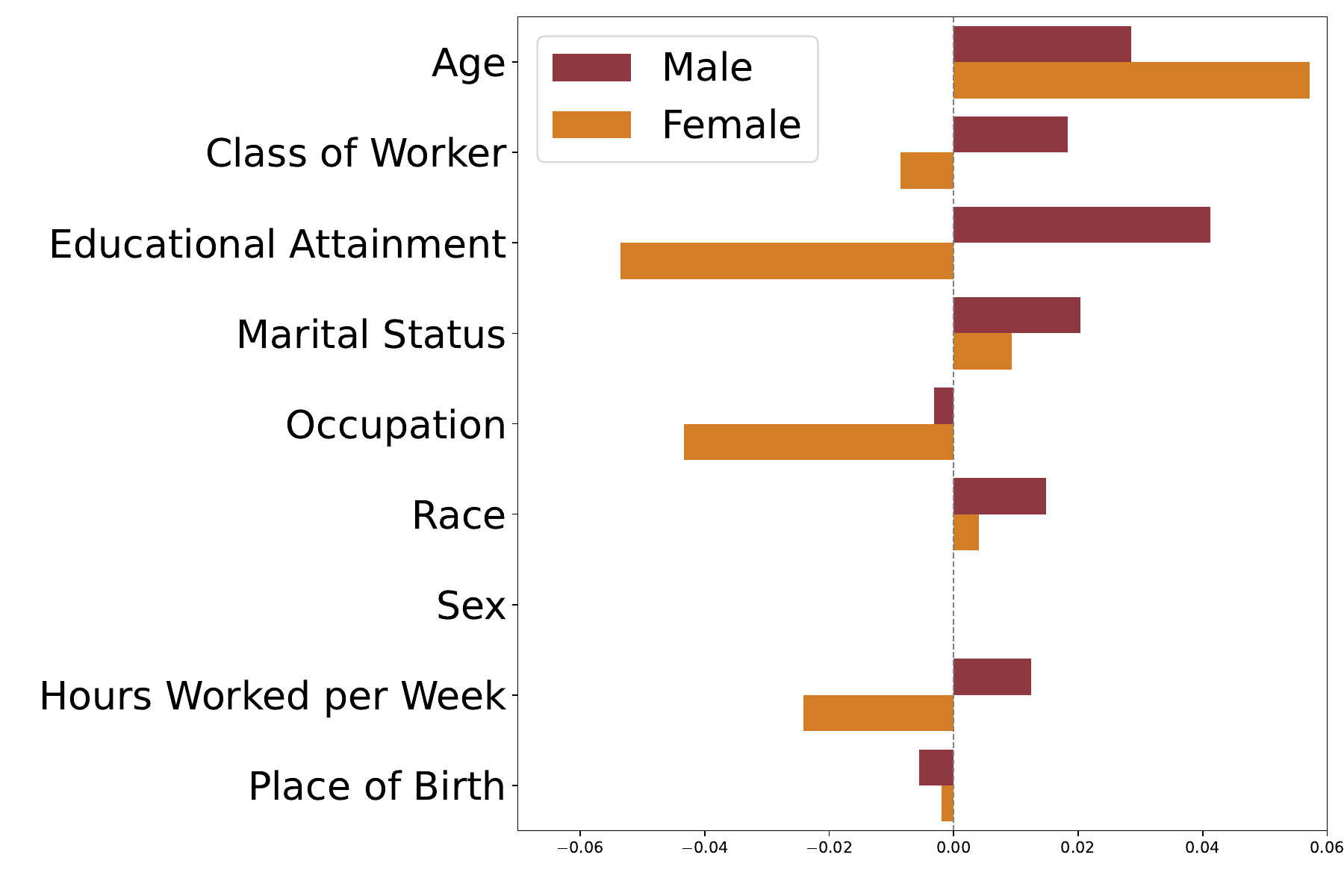}
    \end{subfigure}
    \begin{subfigure}[b]{0.32\textwidth}
        \includegraphics[width=\linewidth]{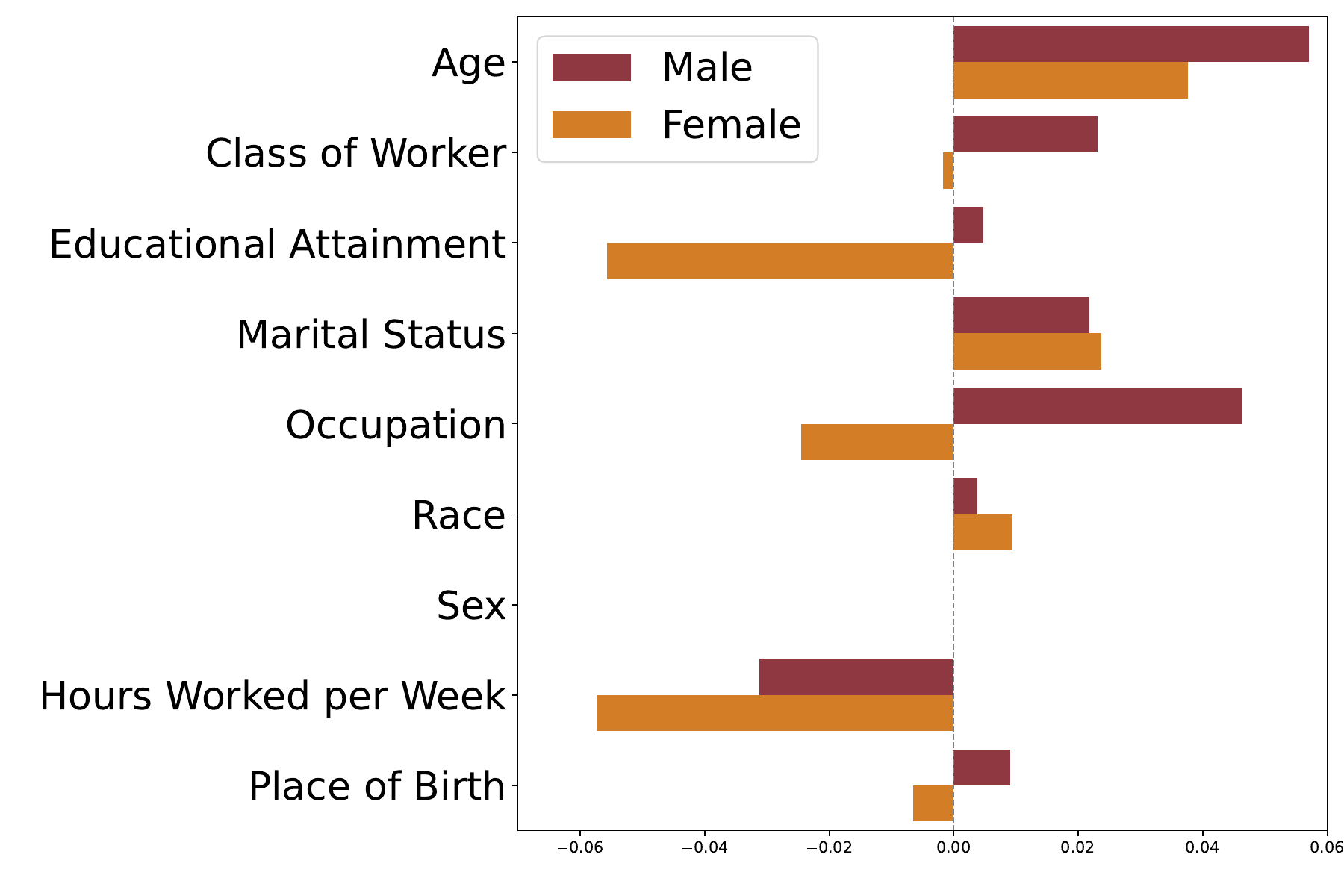}
    \end{subfigure}
    \caption{Differences of mean contributions with SHAP when the protected attribute is removed for the Adult, AdultCA, and AdultLA datasets. Rows represent the datasets, while columns correspond to P, TP, and FP.}
    \label{fig:diff_SHAP}
\end{figure}

Now, we investigate indirect discrimination by excluding the protected attribute from the model. Specifically, we train the model twice, once without the attribute \texttt{sex} and once without \texttt{race}. Table \ref{tab:fairness_diff_no_sex_race} presents the distributive fairness results for these updated models. A comparison between Table \ref{tab:fairness_diff_sex_race} and Table \ref{tab:fairness_diff_no_sex_race} shows that accuracy has a slight decrease. Also, while the disparities in PR, TPR, and FPR fairness have been reduced, violations of the distributive fairness metrics still persist. Fig~ \ref{fig:cor_matrices} shows the contributions of features for the three datasets. We observe that the protected attribute \texttt{sex} is highly correlated with Marital Status, Relationship, and Occupation, while in the other two datasets, it is primarily correlated with Occupation. In Fig.~ \ref{fig:diff_SHAP}, which illustrates the feature contributions for the model trained without the \texttt{sex} attribute for SHAP, we notice an increase in the contributions of features that are correlated with \texttt{sex}. In the Adult dataset, the contributions of features such as Marital Status and those indirectly reflecting gender, such as Relationship, show a significant increase. Additionally, there is a slight rise in the contribution of \texttt{race} for female instances, particularly in positive predictions. When the model includes the protected attribute \texttt{sex}, it exhibits a bias favoring males by pushing them toward the positive class, even in cases where they should belong to the negative class (FP).  However, upon the removal of the \texttt{sex} attribute, this positive bias shifts to other features, such as Marital Status and Relationship, resulting in their increased influence.

\subsection{Results for RQ3: Can aggregated individual/local explanations provide global fairness insights?}

\begin{figure}[H]
  \centering
  \begin{minipage}{0.48\textwidth}
    \centering
    \includegraphics[width=0.8\linewidth]{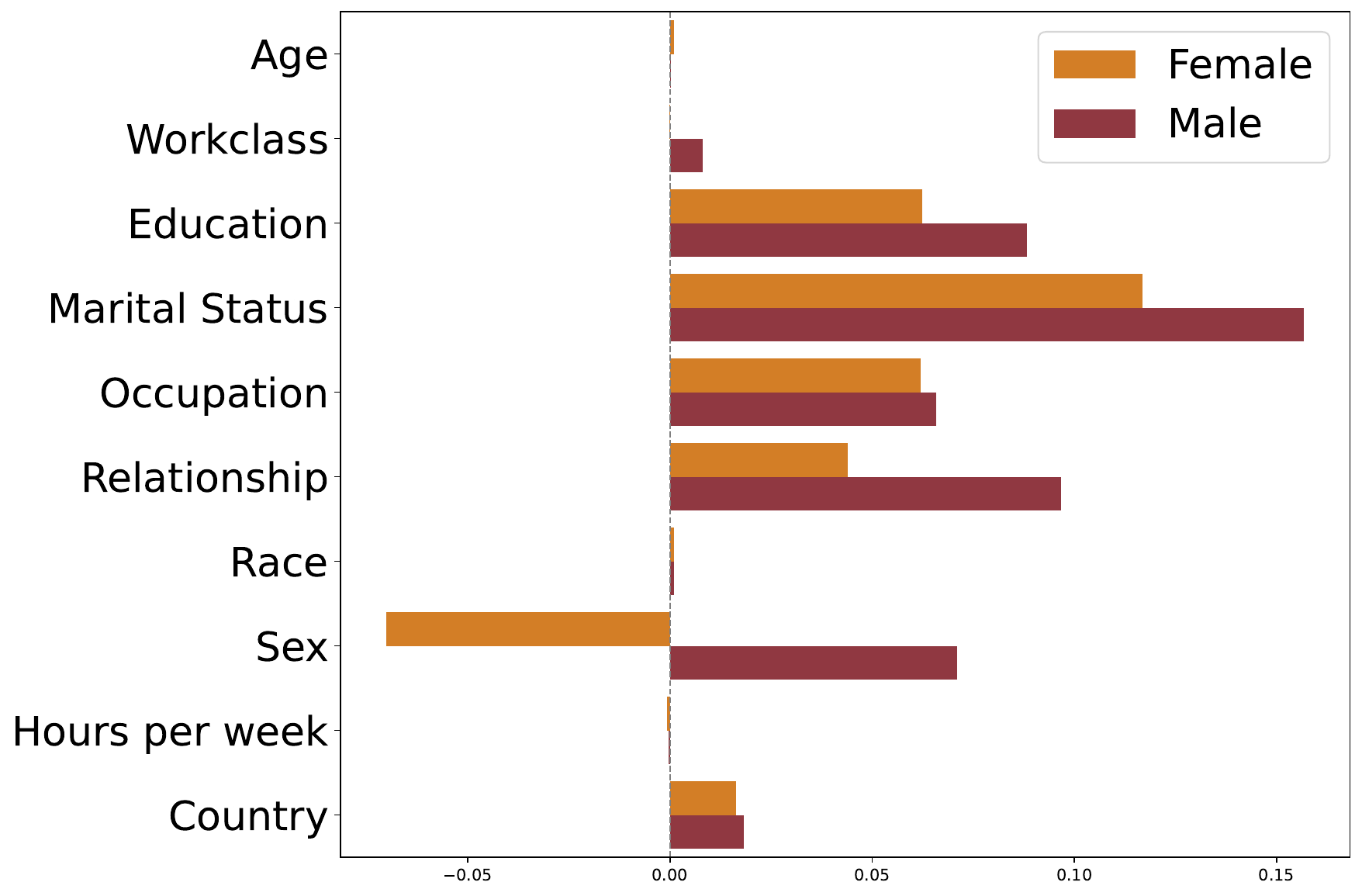}
  \end{minipage}
  \begin{minipage}{0.48\textwidth}
    \centering
    \includegraphics[width=0.8\linewidth]{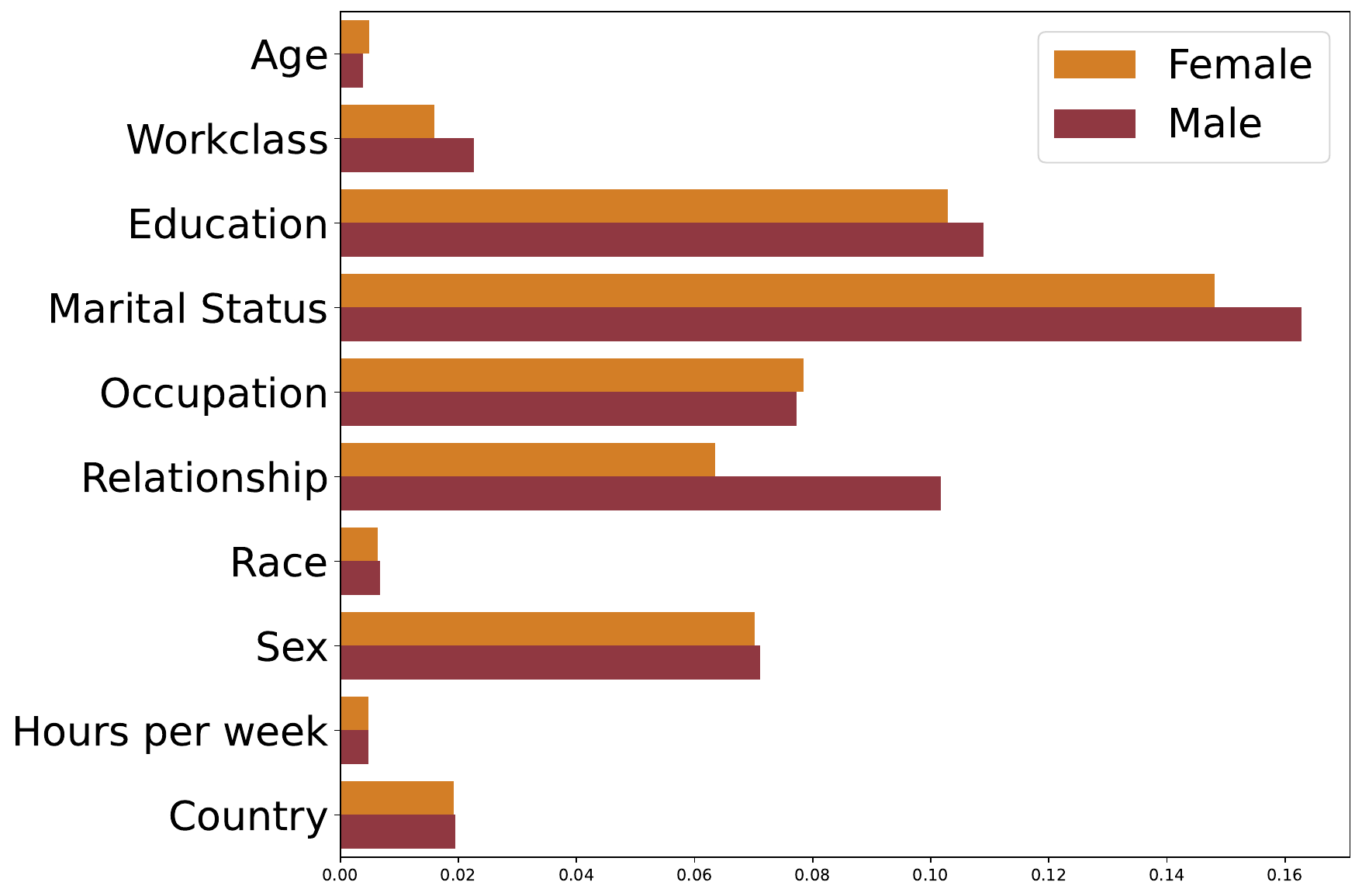}
  \end{minipage}
  \caption{Mean feature contributions vs mean of absolute sum of feature contributions in Adult dataset.}
  \label{label:mean_vs_abs_mean}
\end{figure}

To address the question of aggregation, we explored different approaches for aggregating local explanations within each group and examined the fairness-related insights that can be derived in each case. Fig.~ \ref{label:mean_vs_abs_mean} presents the feature contributions for the Adult dataset using the LIME method. The left side of the figure illustrates the results when aggregating without taking the absolute values, while the right side shows the results when using the mean of the absolute sum. When using the mean of the absolute sum, we observe that attribute \texttt{sex} has almost the same contributions for both male and female, however, this approach fails to capture the direction of the contributions. This hides that female contributions are predominantly negative, meaning that \texttt{sex} attribute negatively influences females by pushing them toward the unfavorable outcome. 

For the aggregation of counterfactual explanations, we used the percentage of feature changes, as shown in Fig.~ \ref{fig:DICE}. This approach allows us to examine the differences in effort required for individuals from different groups in terms of the features that need to be modified. In Fig.~ \ref{fig:DICE}, we observe that in the Adult dataset, the features Relationship and Marital Status need to be changed in almost every counterfactual for both males and females.
Additionally, we computed the burden by calculating the Euclidean distance between factual and counterfactual instances and averaging it across groups. The results for N, FN, and TN instances are presented in Table \ref{tab:fairness_diff_male_female}. We observe that consistently, the protected groups of females and Black people have the largest burden distances. 

\begin{figure}[H]
    \centering
    \begin{subfigure}[b]{0.50\textwidth}
        \includegraphics[width=\linewidth]{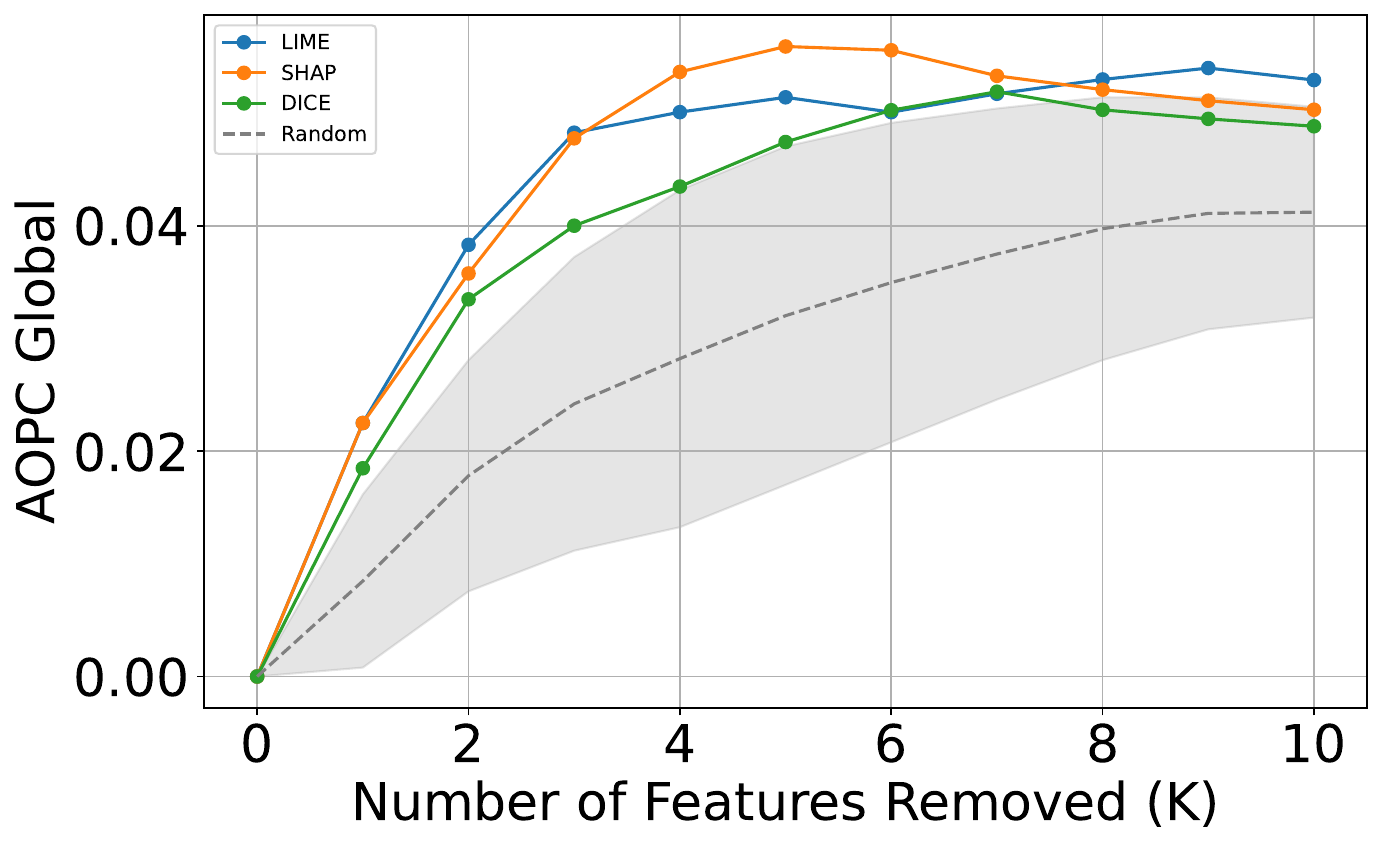}
    \end{subfigure}
    
    \caption{AOPC curves for LIME, SHAP and DICE explanations.}
    \label{fig:AOPC}
\end{figure}

\textbf{RQ4. How do different explanation methods compare in terms of robustness, consistency and explanation quality, and can they be trusted?} 
To compare explanation methods and assess their reliability, we use the AOPC curve. We sample 500 instances from the test set, generate explanations, rank features by contribution for every method, and replace features with their mean according to the ranked order. The AOPC score is computed as the average cumulative drop in predicted class probability up to the given rank, averaged across all instances. We also include a curve based on random ranking for comparison. Fig.~ \ref{fig:AOPC} shows the AOPC curves for the Adult dataset using LIME, SHAP, and DiCE. A higher and steeper curve indicates better feature ranking. We observe that all methods outperform the random baseline, with LIME and SHAP having steeper slopes when removing the most important features. This suggests that the features identified by the explanations are indeed important for the model, providing evidence that we can trust the feature rankings.

\begin{table}[H]
  \centering
  \caption{Mean distance of factuals to DiCE counterfactuals by \texttt{sex} (M/F) and \texttt{race} (W/B) for N, FN, and TN.}
  \label{tab:fairness_diff_male_female}
  \begin{tabular}{lcccccccccccc}
    \toprule
     & \multicolumn{4}{c}{Adult} & \multicolumn{4}{c}{AdultCA} & \multicolumn{4}{c}{AdultLA} \\
    \cmidrule(r){2-5} \cmidrule(r){6-9} \cmidrule(r){10-13}
           & M & F & W & B & M & F & W & B & M & F & W & B \\
    \midrule
    N     & 3.91 & 5.11 & 4.05 & 5.62 & 3.53& 3.91 & 3.31 & 4.01 & 4.14 & 4.34 & 3.98 & 5.18 \\
    FN    & 2.03 & 3.56 & 2.29 & 3.08 & 3.31 & 3.34 & 2.9 & 22.98 & 3.62 & 3.2 & 2.88 & 3.4 \\
    TN    & 3.75 & 5.24 & 4.08 & 5.04 & 3.53 & 4.03  & 3.51 &4.16 & 20.64 & 5.42 & 4.0 & 4.66 \\
  \end{tabular}
\end{table}

\section{Related Work}
\label{related_work}
Fairness explanation methods \cite{fragkathoulas2024explaining} focus on detecting biases, defining fairness metrics by quantifying mitigation costs, and mitigating unfairness by reducing dependence on sensitive attributes or recommending modifications.
To this end, feature attribution methods such as LIME \cite{ribeiro2016should} and SHAP \cite{lundberg2017unified} have been used to uncover unfair behavior by quantifying feature importance, while counterfactual techniques like DiCE \cite{mothilal2020explaining} identify minimal feature changes needed to achieve different outcomes, providing actionable recourse.
Specifically, LimeOut \cite{LIMEOUT1, LIMEOUT2} leverages LIME to detect and mitigate unfairness by generating diverse explanations using the Submodular Pick algorithm \cite{ribeiro2016should}.
The work in \cite{biased-models} uses SHAP values to define fairness, assuming minimal impact from protected attributes. Specifically, demographic parity requires minor mean absolute values, equality of opportunity demands similar distributions for positives, and equalized odds extend this to both positive and negative instances.
Additionally, \cite{CesaroC19} propose a two-step method: applying pre-processing like re-weighting to reduce bias and comparing SHAP values between original and bias-corrected models, showing lower SHAP values indicate improved fairness.
Moreover, \cite{grabowicz2022marrying} measures the influence of protected and proxy attributes on decisions, modeling outcomes as shaped by non-protected and protected variables. 
In addition, \cite{begley2020explainability} defines demographic parity as the Shapley value difference between protected and non-protected groups. Using the additive property, they capture unfairness via contribution summation and propose a meta-algorithm that adjusts the model for fairness.

Regarding counterfactual explanations, the work in \cite{goethals2023precof} presents PreCoF, a counterfactual method that identifies features disproportionately contributing to negative outcomes for protected groups. It generates counterfactuals for negatively classified instances across groups, analyzing feature changes to identify those with the largest impact differences as indicators of unfairness.
Group counterfactuals \cite{rawal2020beyond} extend counterfactual explanations to groups, providing conditions for favorable outcomes instead of identifying key features by formulating the problem as a constraint optimization problem.
Building on this, several approaches have been proposed: GLOBE-CE \cite{ley2023globe} defines a global direction along which a group can adjust its features, counterfactual explanation trees \cite{kanamori2022counterfactual} assign actions to multiple instances, and FACTS \cite{kavouras2023fairness} employs a frequent itemset approach to analyze fairness at the subgroup level.

\section{Conclusion and Future Work}

\label{conclusion}

Our study explores the use of explanations to analyze fairness by proposing a pipeline that integrates local post-hoc methods for fairness insights and evaluating various approaches in each step, considering the most appropriate ones for studying fairness. 
Through our study, we observe that when distributive fairness is violated we get similar signs of procedural unfairness, such as unequal contributions of protected attributes across groups or disproportionate use of irrelevant features. We find that when the protected attribute is removed, the contributions of features correlated with it, those containing hidden information about it, or those unrelated to the task tend to increase, often favoring the non-protected group. We observe that different aggregation methods offer varying visibility into feature importance, leading to diverse interpretations and insights into fairness. Lastly, using the AOPC curve, we conclude that the different explanation methods show consistency. This suggests that, to some extent, we can trust these explanations for understanding fairness but it is essential to proceed with caution, considering the decisions made at each step of the pipeline and accounting for the properties of each method. In future work, we aim to explore additional explanation methods, such as group counterfactuals and global explanations, and also to evaluate the fairness of the explanations themselves. Finally, conducting a user study would be valuable for assessing procedural unfairness from the users' perspective and for understanding the effect of explanations on the users' perception of fairness.

\bibliographystyle{ACM-Reference-Format}
\bibliography{main}

\appendix
\section{Appendix}
\label{appendix}
\subsection{Datasets}
\begin{table}[H]
  \centering
  \caption{Dataset description with features, feature descriptions and feature types.}
  \label{tab:dataset_features}
  \begin{tabular}{l l l l}
    \toprule
    Dataset & Feature & Description & Type \\
    \midrule
    \multirow{10}{*}{Adult} & Age & The age of individual & Numeric \\
                             & Workclass & The employment status & Categorical \\
                             & Education & The highest level of education & Categorical \\
                             & Marital Status & The marital status of the individual & Categorical \\
                             & Occupation & The general type of occupation & Categorical \\
                             & Relationship & What this individual is relative to others & Categorical \\
                             & Race & Racial background of the individual & Categorical \\
                             & Sex & Gender of the individual & Categorical \\
                             & Hours-per-week & Number of hours worked per week & Numeric \\
                             & Native-country & The country of origin for an individual & Categorical \\
                             & Income & If an individual makes more than \$50,000 annually & Binary \\
    \midrule
    \multirow{9}{*}{AdultCA \& AdultLA}& Age & The age of individual & Numeric \\
                             & Class of worker & The employment status & Categorical \\
                             & Educational attainment & The highest level of education & Categorical \\
                             & Marital Status & The marital status of the individual & Categorical \\
                             & Occupation & The general type of occupation & Categorical \\
                             & Race & Racial background of the individual & Categorical \\
                             & Sex & Gender of the individual & Categorical \\
                             & Hours worked per week & Number of hours worked per week & Numeric \\
                             & Place of birth & The country of origin for an individual & Categorical \\
                             & Income & If an individual makes more than \$50,000 annually & Binary \\
    \bottomrule
  \end{tabular}
\end{table}

\subsection{RQ1. How does feature attribution relate to statistical group fairness? What is the relationship, if any,
between process fairness and output fairness?}

Table \ref{tab:contributions_dif_race} presents the differences in mean contributions of the protected attribute race between the non-protected (White) and protected (Black) groups across three explanation methods: LIME, SHAP, and DiCE.For LIME and SHAP, these differences are computed by subtracting the average contribution for the non-protected group from that of the protected group. In contrast, for DiCE, the direction of subtraction is reversed to capture the additional feature changes required for members of the protected group to receive favorable outcomes. Overall, the disparities in race contributions are smaller than those observed for the protected attribute sex, as reported in Table \ref{tab:contributions_dif_sex}, suggesting that sex may play a more significant role in shaping model behavior and potential bias.Among the explanation methods, LIME generally shows larger differences compared to SHAP. Notably, the AdultLA dataset exhibits pronounced disparities in race contributions under both LIME and SHAP, indicating stronger potential bias.

\begin{table*}[h]
\centering
\small
\renewcommand{\arraystretch}{1.5} 
\caption{Difference in mean contributions for the protected attribute race between White and Black using LIME and SHAP for Positives, True Positives and False Positives and feature change percent differences in DiCE for Negatives, False Negatives and True Negatives}
\vspace{-1em}
\begin{tabular}{p{1cm}|p{0.52cm}|p{0.52cm}|p{0.52cm}|p{0.75cm}|p{0.75cm}|p{0.75cm}|p{0.75cm}|p{0.52cm}|p{0.4cm}|p{0.4cm}|p{0.4cm}|p{0.4cm}}
\hline
\multirow{2}{*}{Dataset} & \multicolumn{3}{c|}{\textbf{LIME}} & \multicolumn{3}{c|}{\textbf{SHAP}} & \multicolumn{3}{c|}{\textbf{DiCE}} \\
\cline{2-10}
& P & TP & FP & P & TP & FP & N & FN & TN \\
\hline
Adult & 0.002 & 0 & 0.001& -0.092 & -0.096 & -0.105 & 66 & 75 & 73 \\
\hline
AdultCA& 0.03 & 0.033 & 0.033 & 0 & 0.001 & 0.013 & 52 & 43 & 51 \\
\hline
AdultLA & 0.095 & 0.095 & 0.098 & 0.085 & 0.076 & 0.09 & 62 & 50 & 58 \\
\hline
\end{tabular}
\label{tab:contributions_dif_race}
\end{table*}

\subsection{RQ2. What is the effect in fairness of removing the protected attribute? How this relates to direct
(dependency on the protected attribute) vs indirect discrimination (dependency on proxy attributes)?}

\begin{figure}[h]
    \centering

    \begin{subfigure}[b]{0.32\textwidth}
        \includegraphics[width=\linewidth]{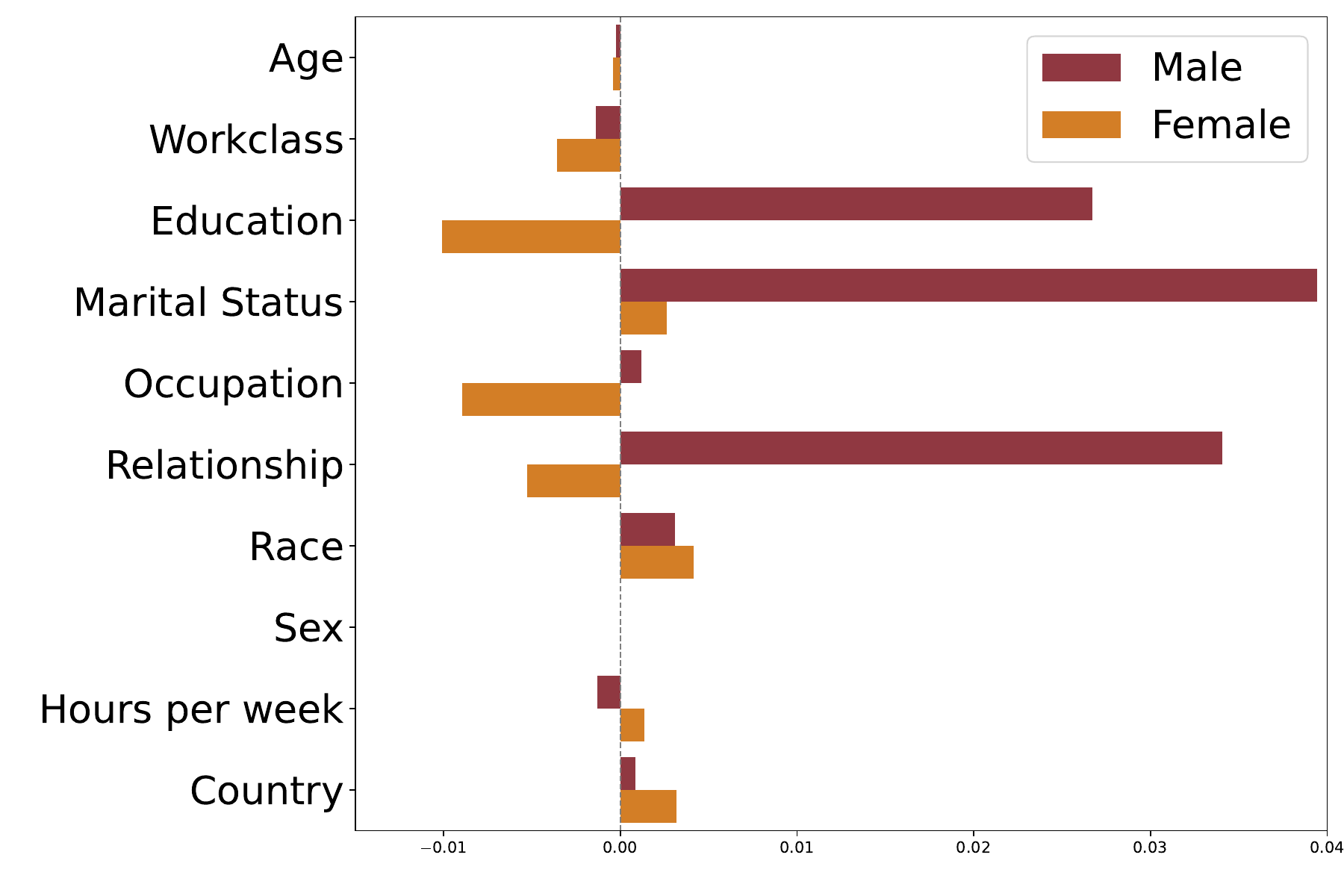}
    \end{subfigure}
    \begin{subfigure}[b]{0.32\textwidth}
        \includegraphics[width=\linewidth]{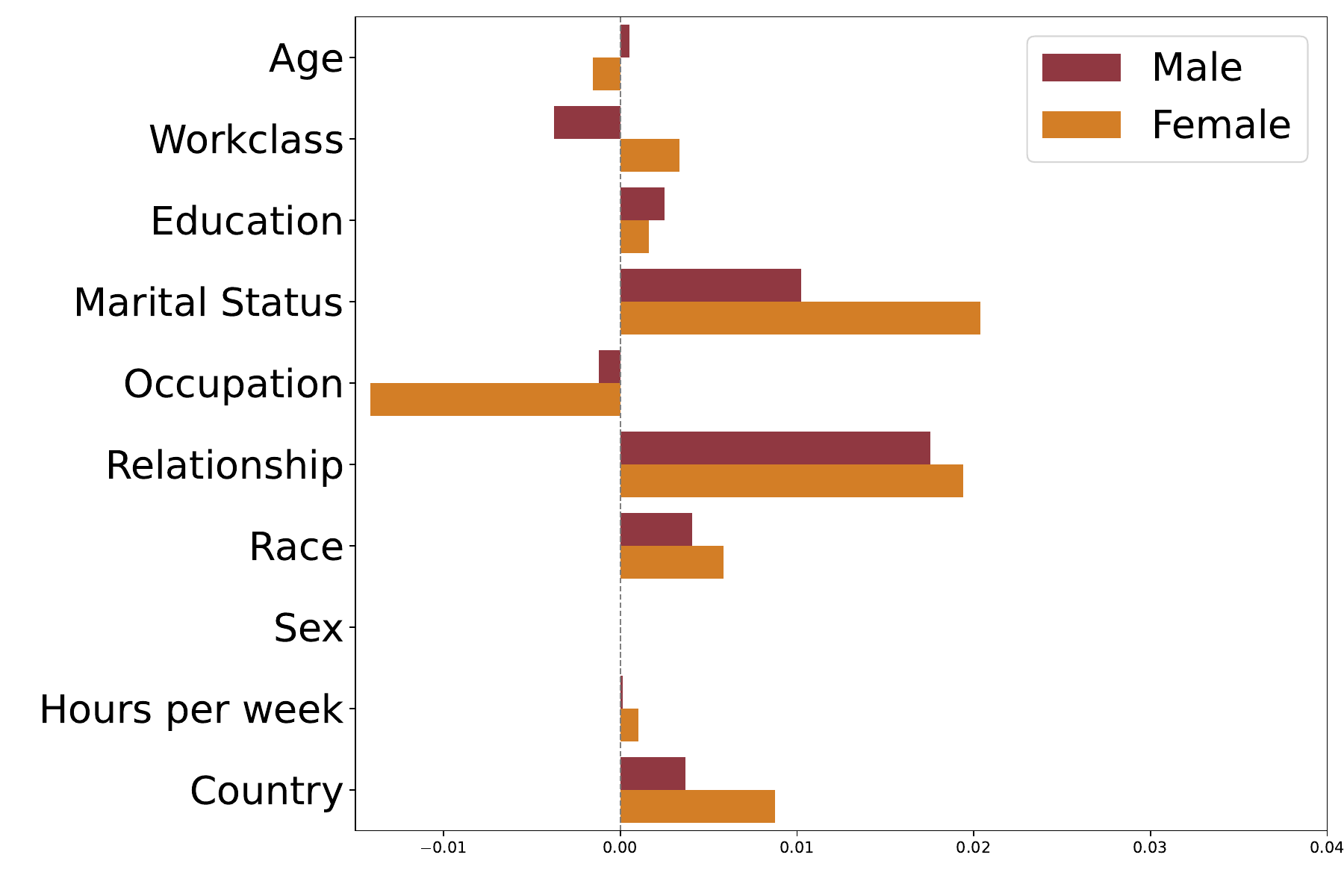}
       
    \end{subfigure}
    \begin{subfigure}[b]{0.32\textwidth}
        \includegraphics[width=\linewidth]{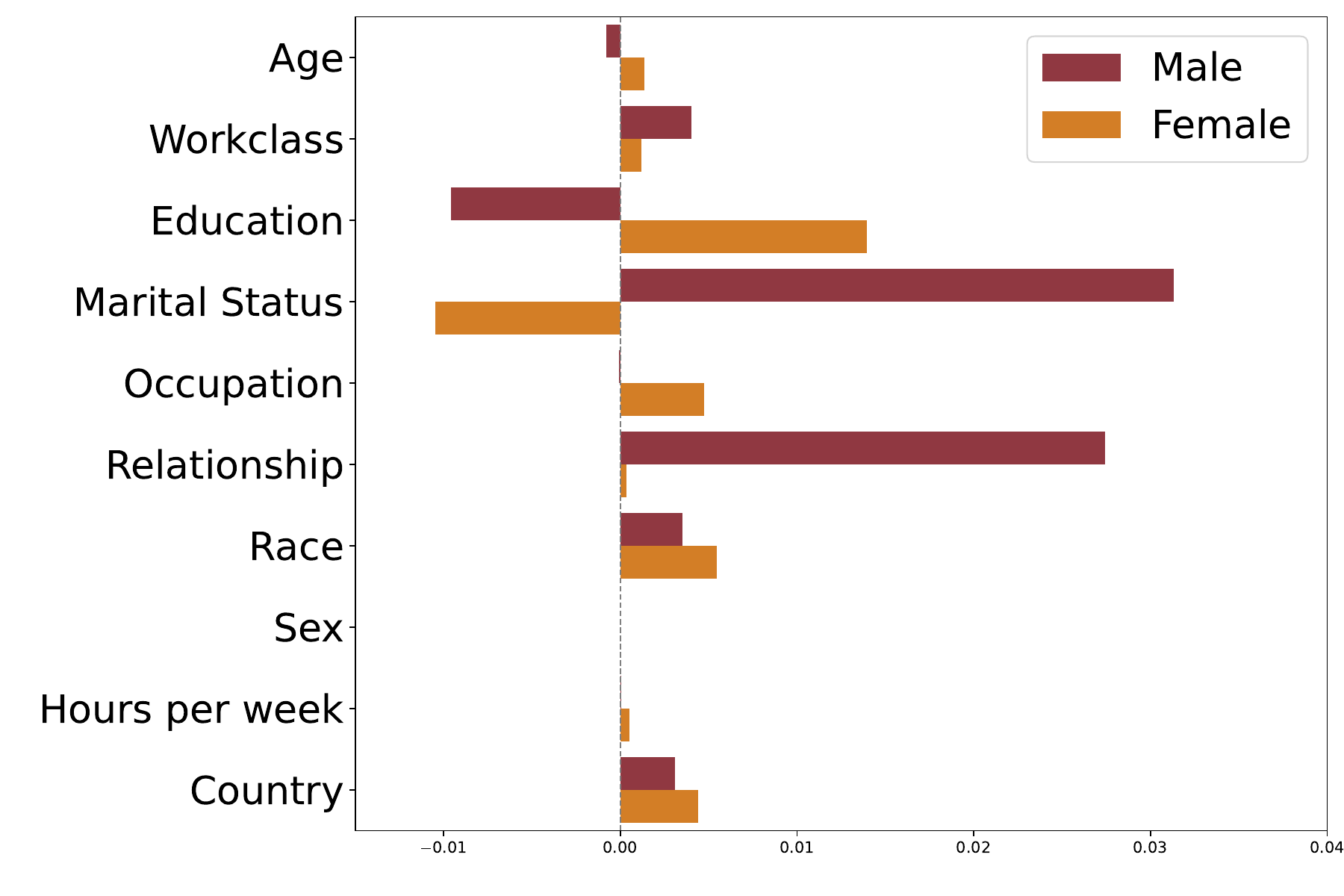}   
    \end{subfigure}
    \begin{subfigure}[b]{0.32\textwidth}
        \includegraphics[width=\linewidth]{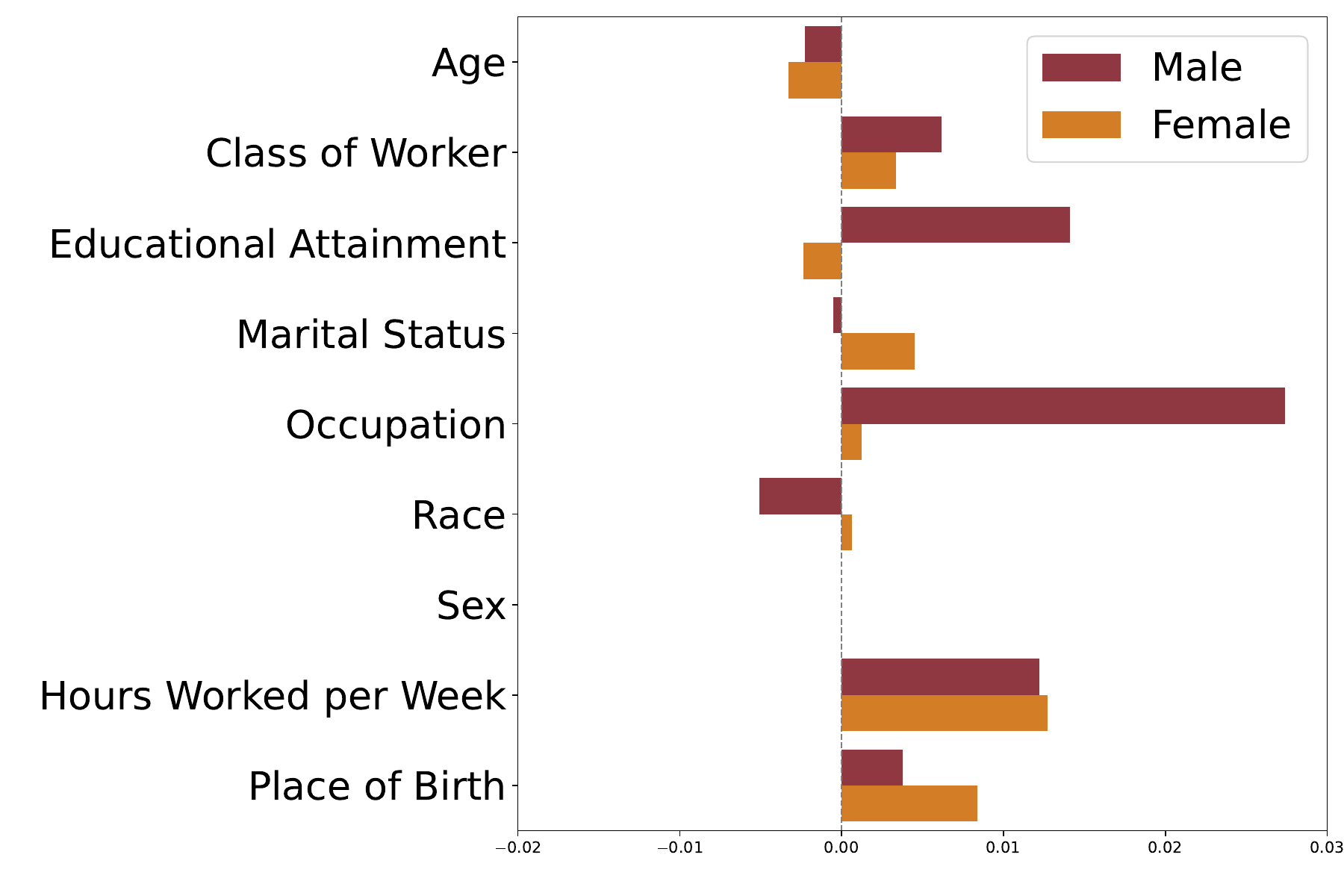}
    \end{subfigure}
    \begin{subfigure}[b]{0.32\textwidth}
        \includegraphics[width=\linewidth]{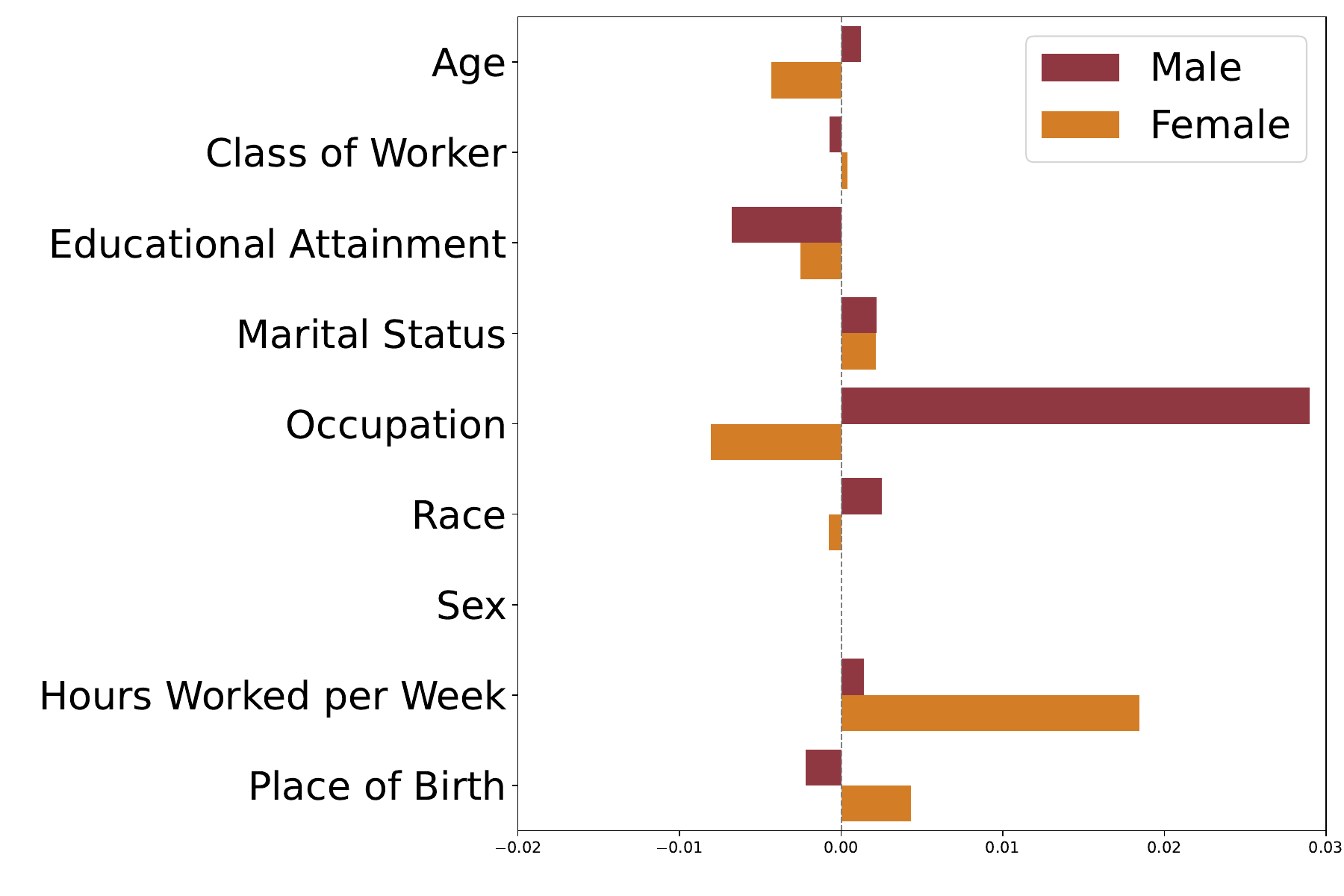}
    \end{subfigure}
    \begin{subfigure}[b]{0.32\textwidth}
        \includegraphics[width=\linewidth]{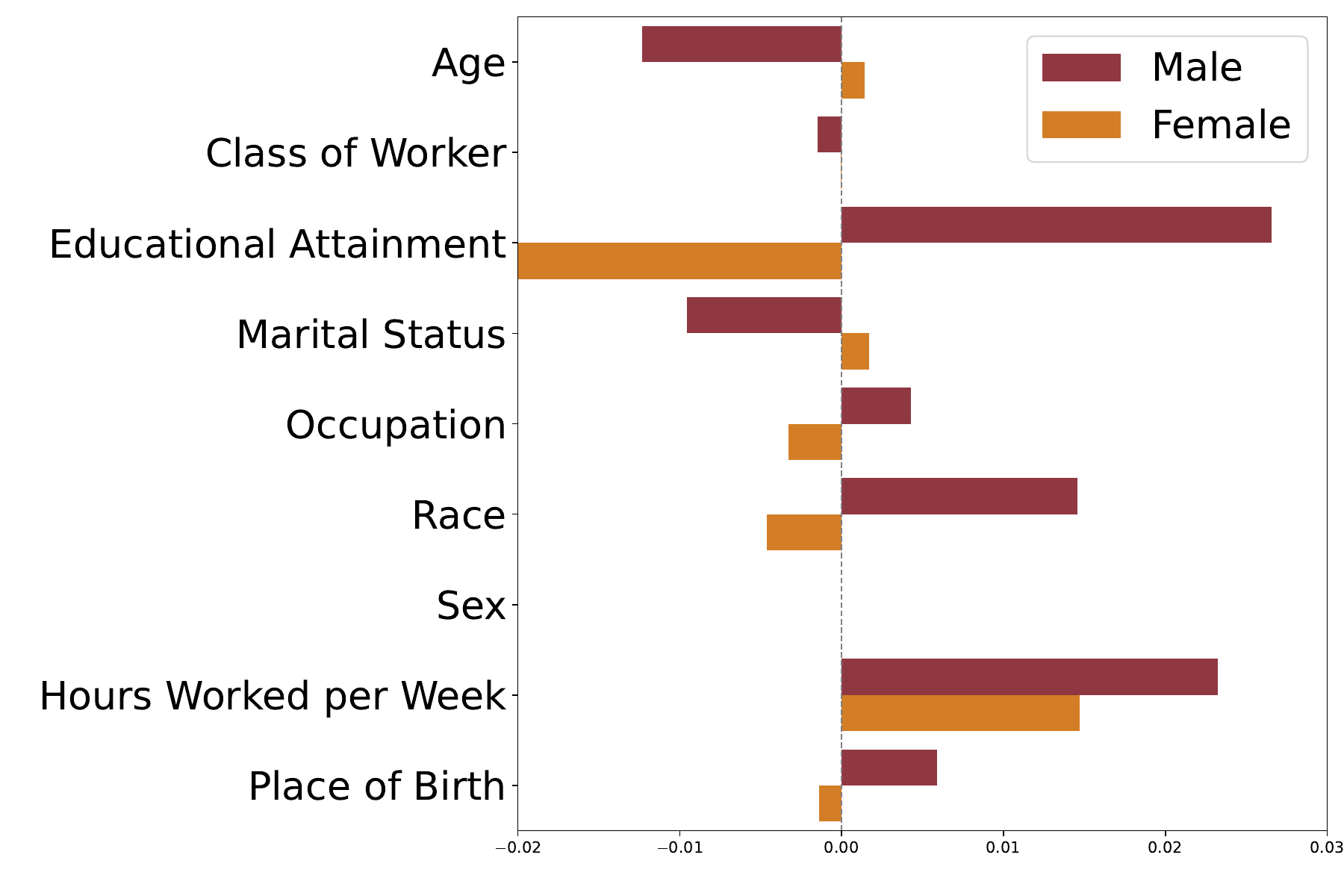}
    \end{subfigure}
    \begin{subfigure}[b]{0.32\textwidth}
        \includegraphics[width=\linewidth]{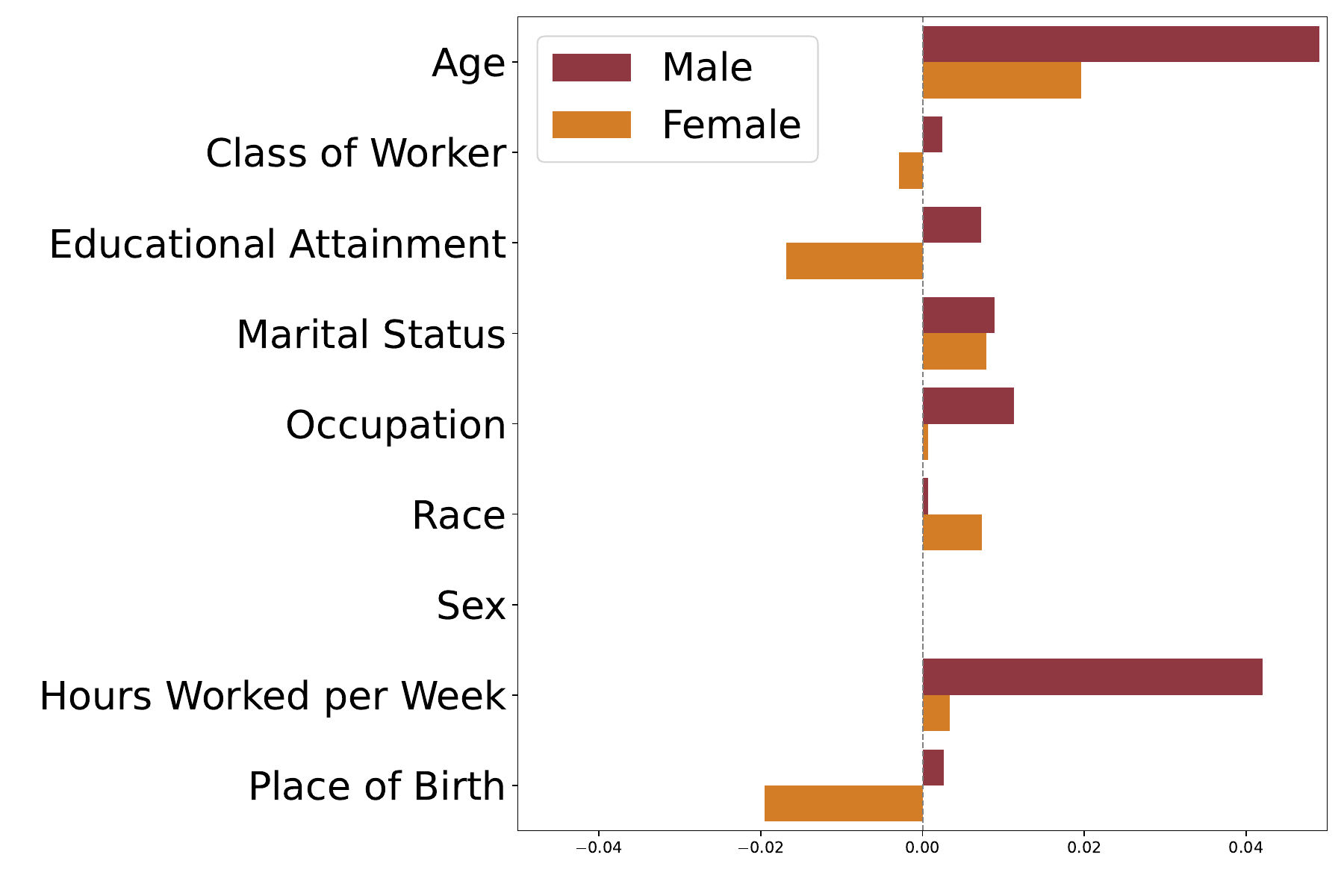}
    \end{subfigure}
    \begin{subfigure}[b]{0.32\textwidth}
        \includegraphics[width=\linewidth]{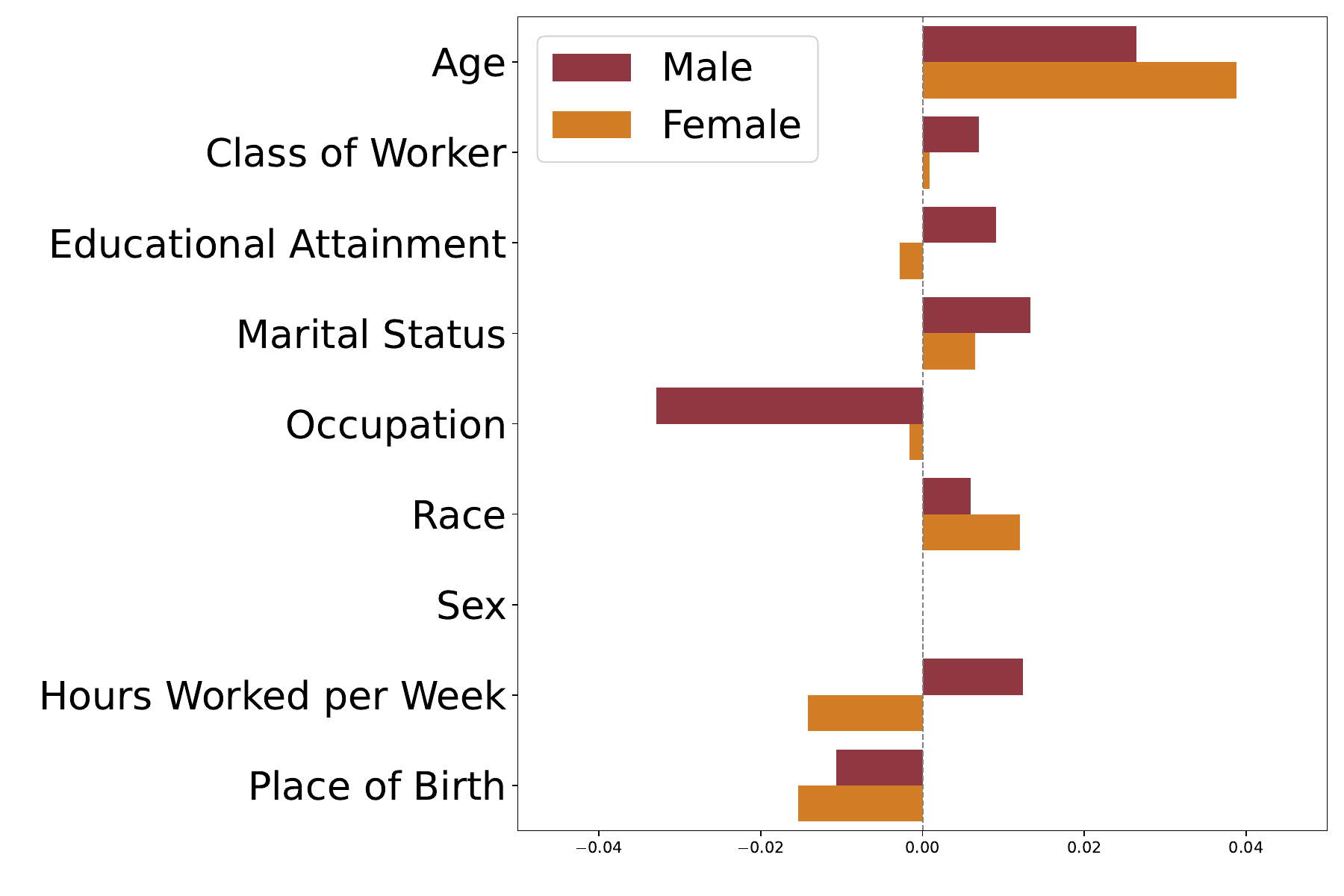}
    \end{subfigure}
    \begin{subfigure}[b]{0.32\textwidth}
        \includegraphics[width=\linewidth]{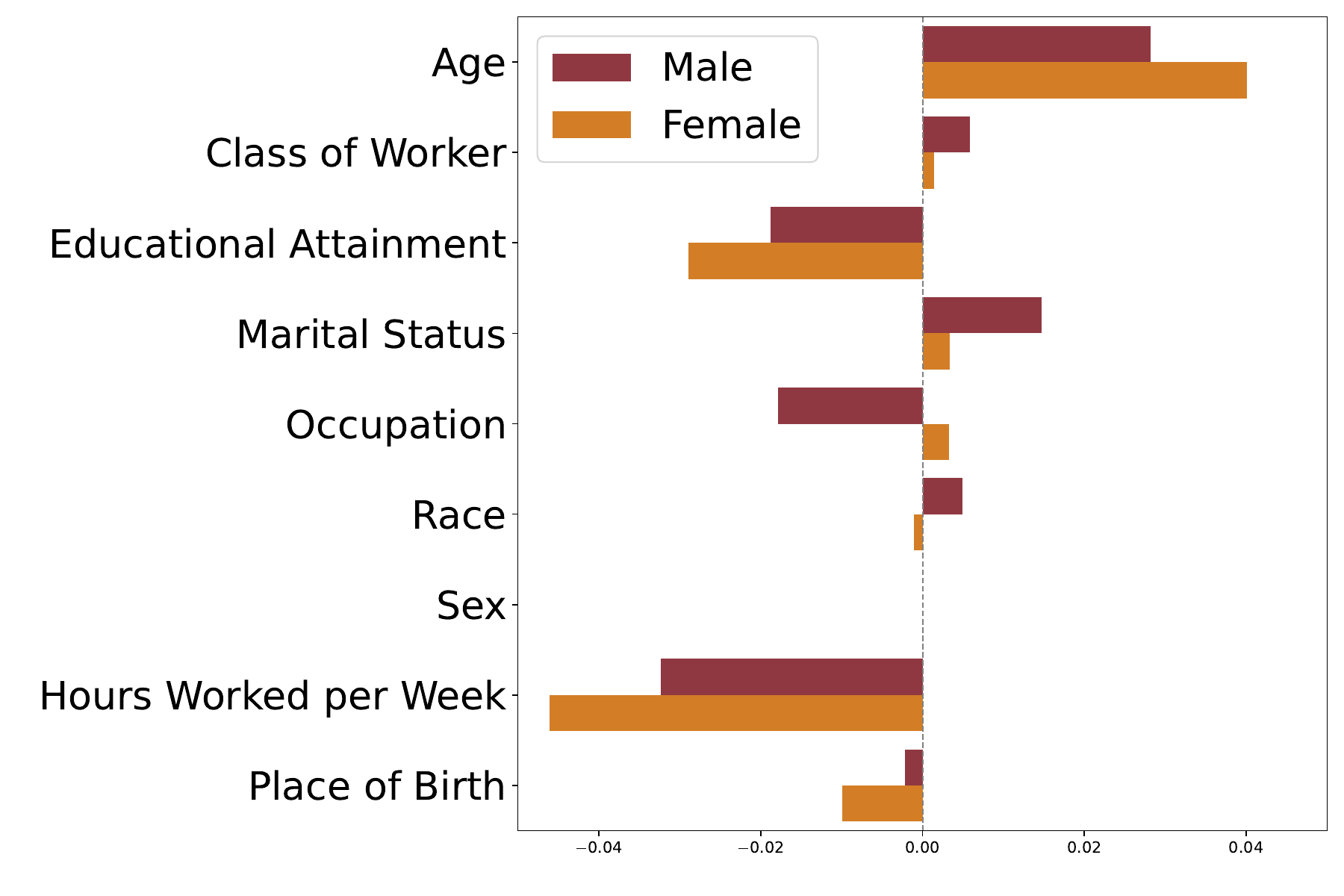}
    \end{subfigure}
    \caption{Differnces of mean contributions with LIME when the protected attribute sex is removed for the Adult, AdultCA, and AdultLA datasets. Rows represent the datasets, while columns correspond to P, TP, and FP.}
    \label{fig:diff_LIME}
\end{figure}

\begin{figure}[h]
    \centering
    \begin{subfigure}[b]{0.32\textwidth}
        \includegraphics[width=\linewidth]{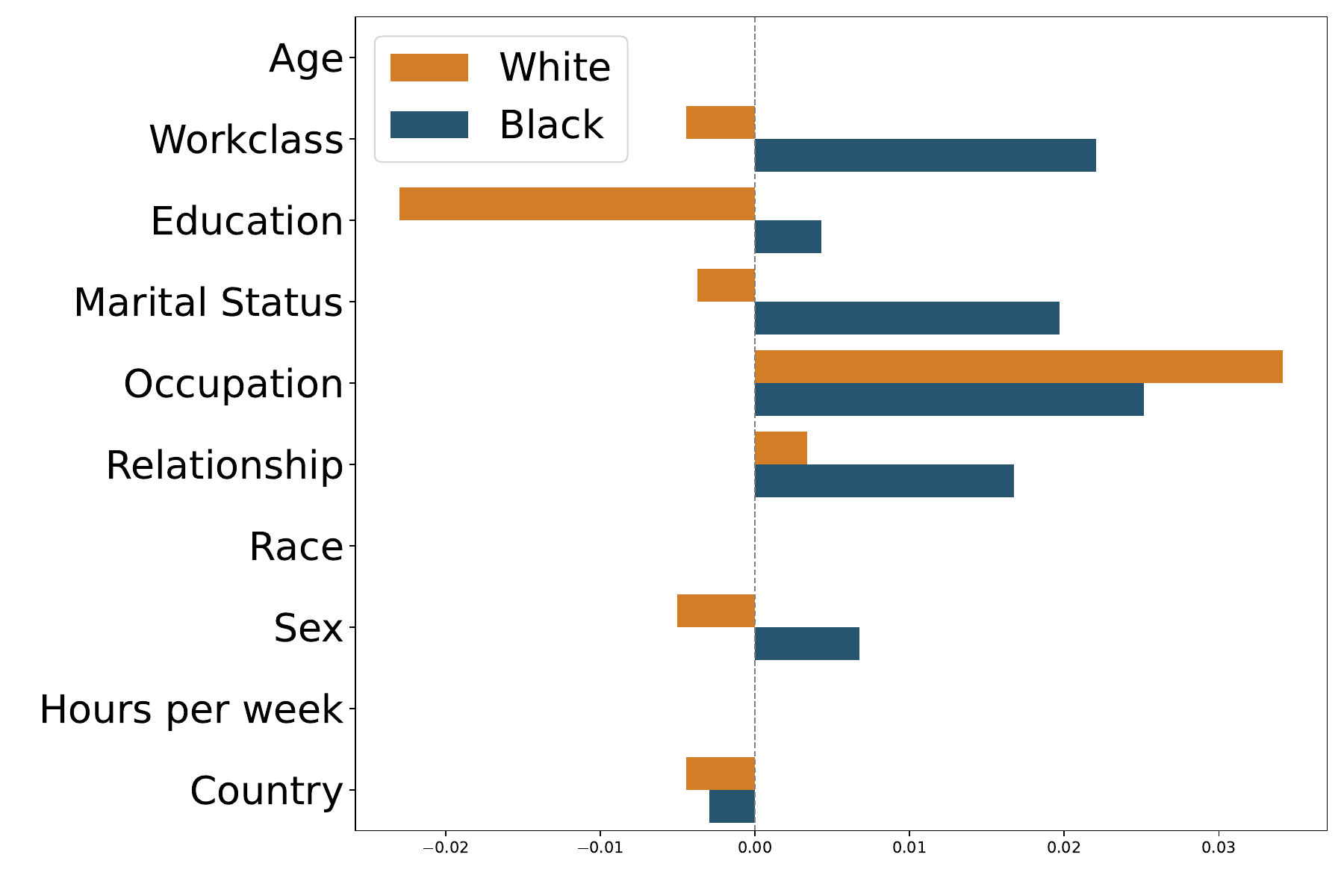}
    \end{subfigure}
    \begin{subfigure}[b]{0.32\textwidth}
        \includegraphics[width=\linewidth]{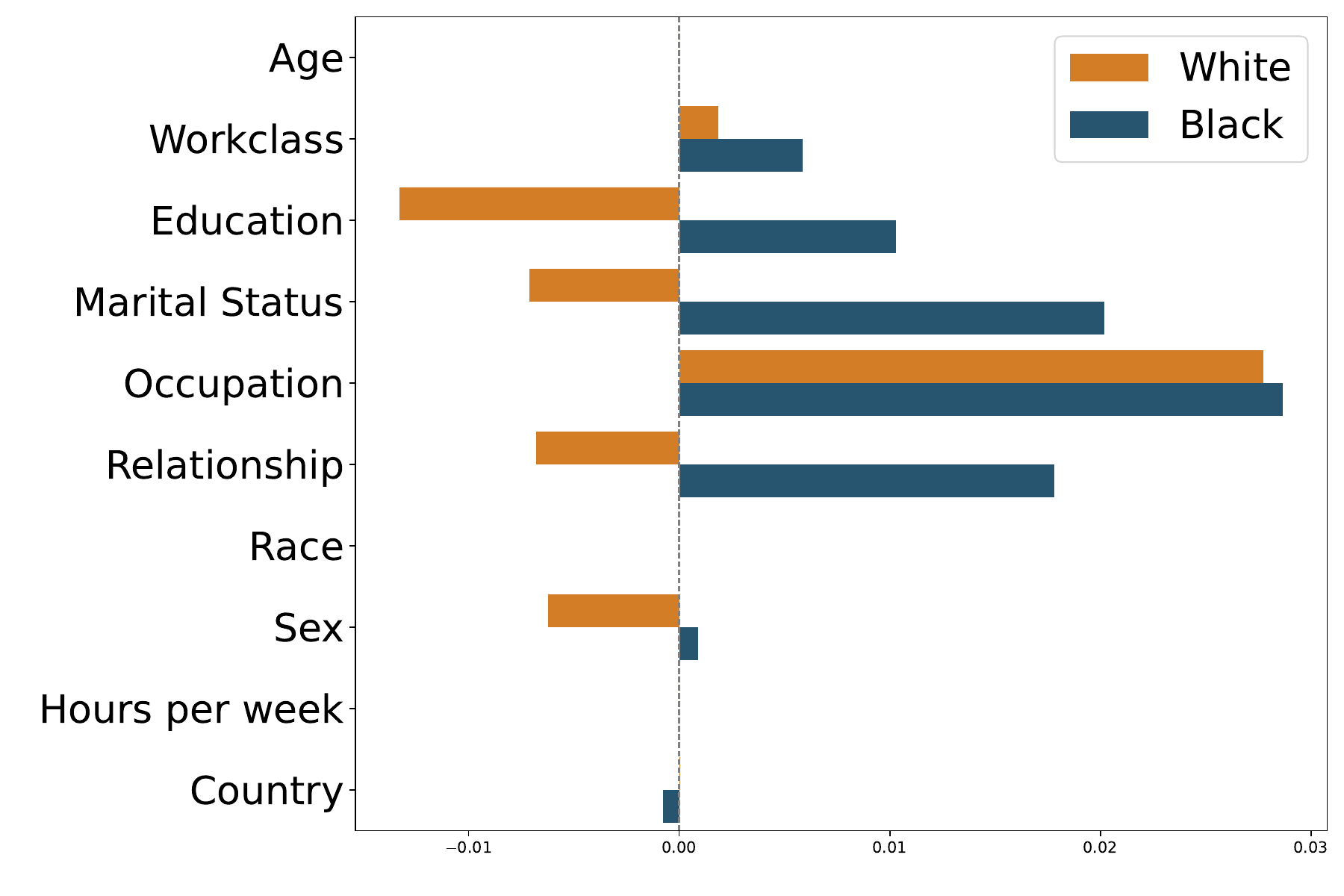}
    \end{subfigure}
    \begin{subfigure}[b]{0.32\textwidth}
        \includegraphics[width=\linewidth]{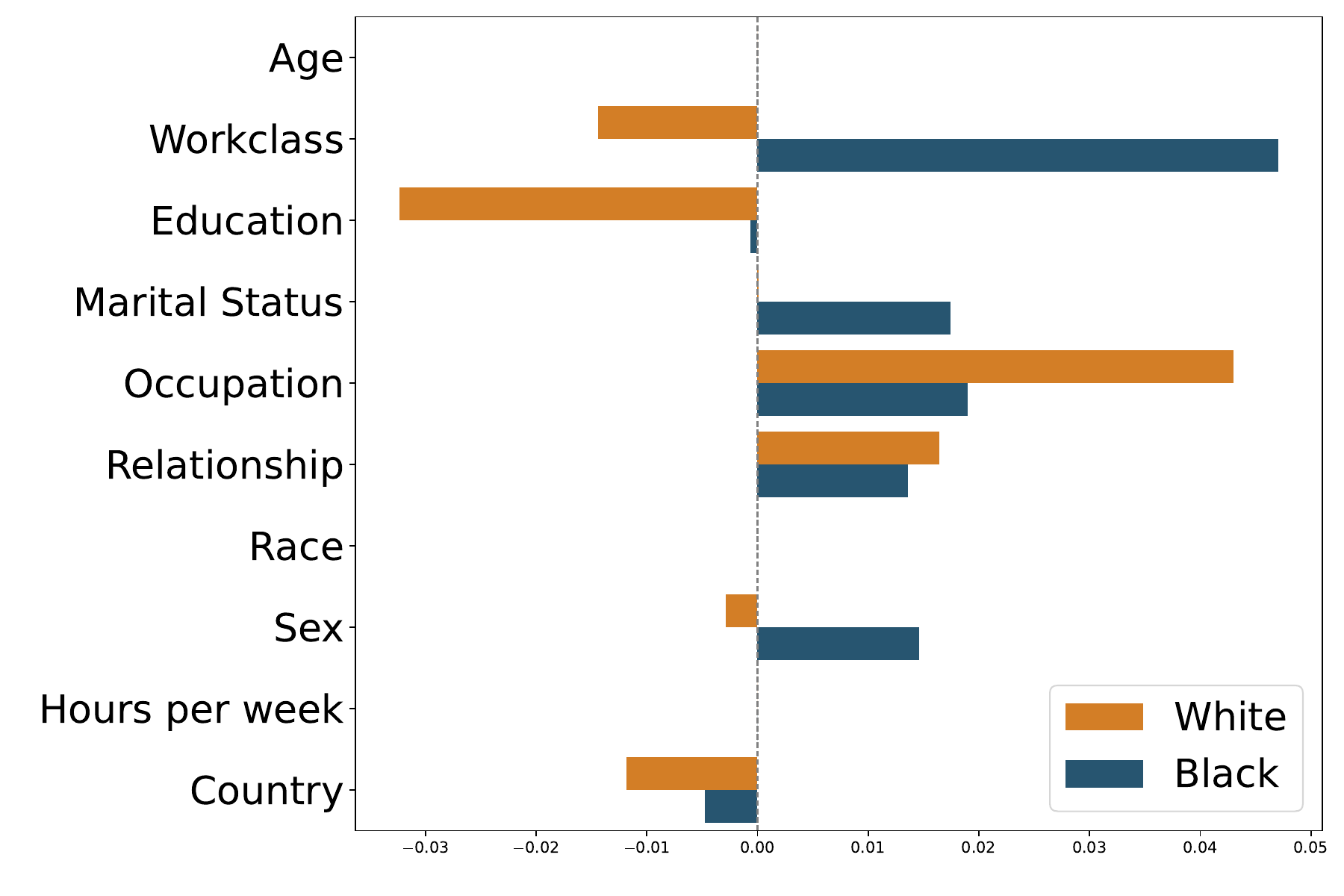}
    \end{subfigure}
    \begin{subfigure}[b]{0.32\textwidth}
        \includegraphics[width=\linewidth]{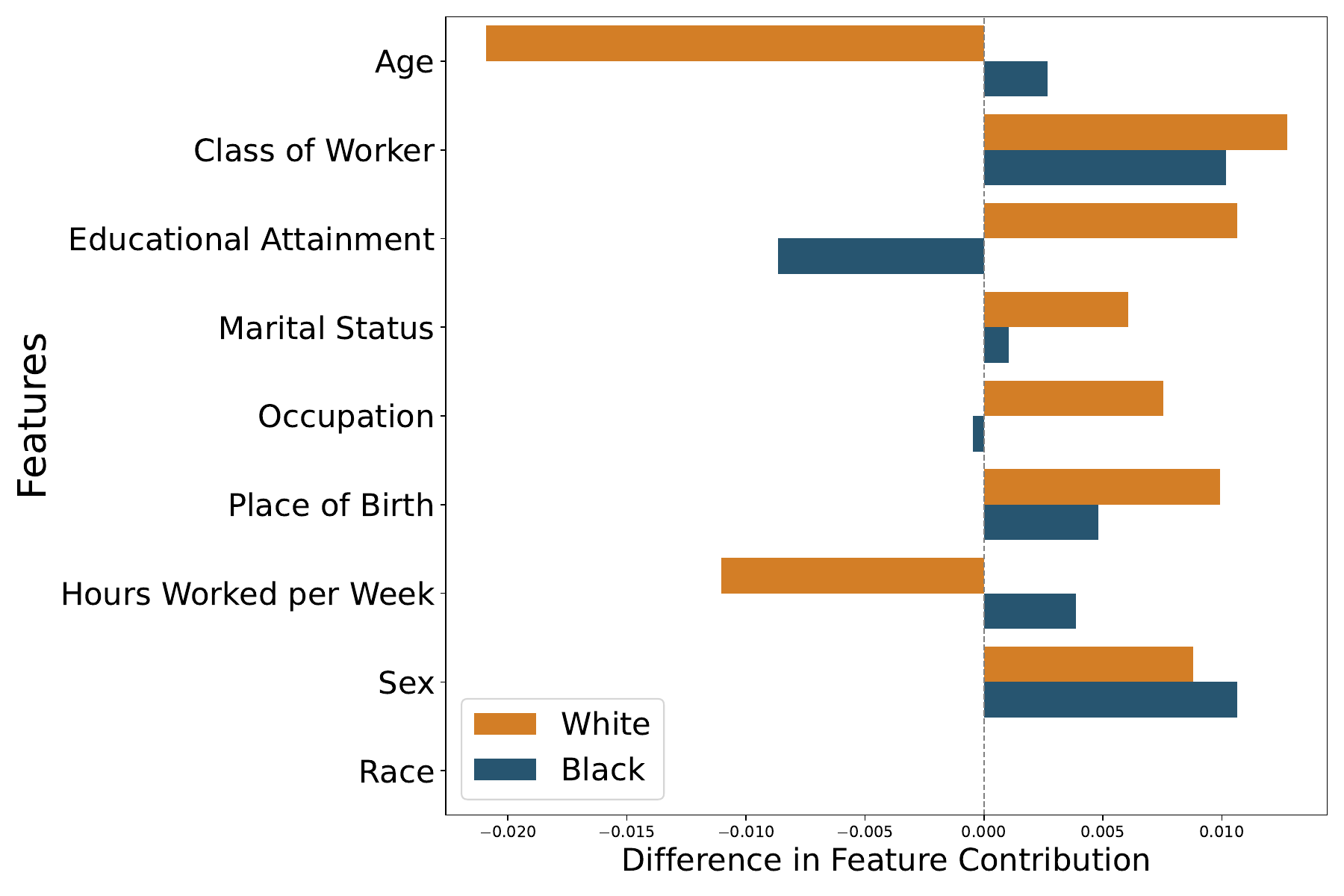}
    \end{subfigure}
    \begin{subfigure}[b]{0.32\textwidth}
        \includegraphics[width=\linewidth]{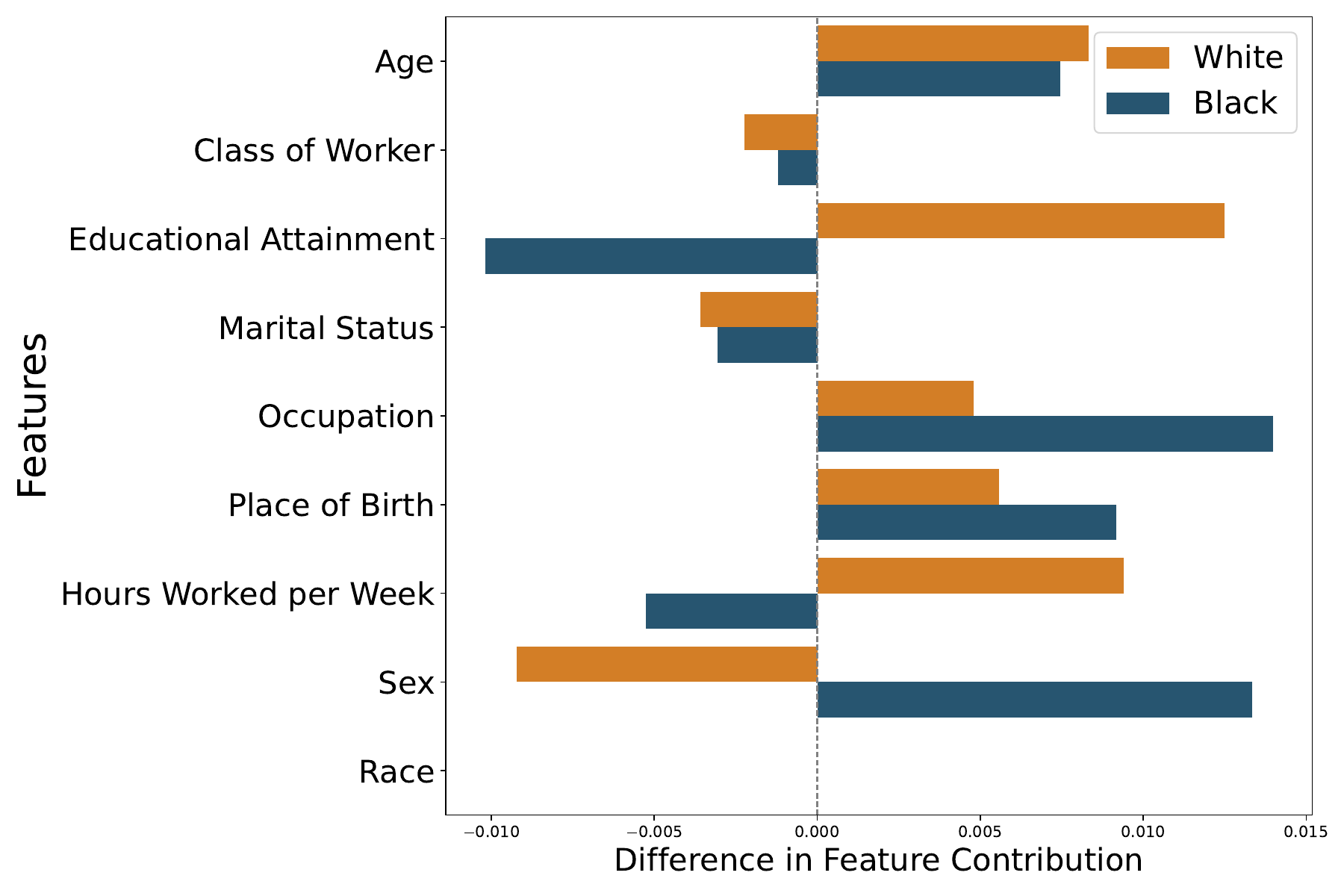}  
    \end{subfigure}
    \begin{subfigure}[b]{0.32\textwidth}
        \includegraphics[width=\linewidth]{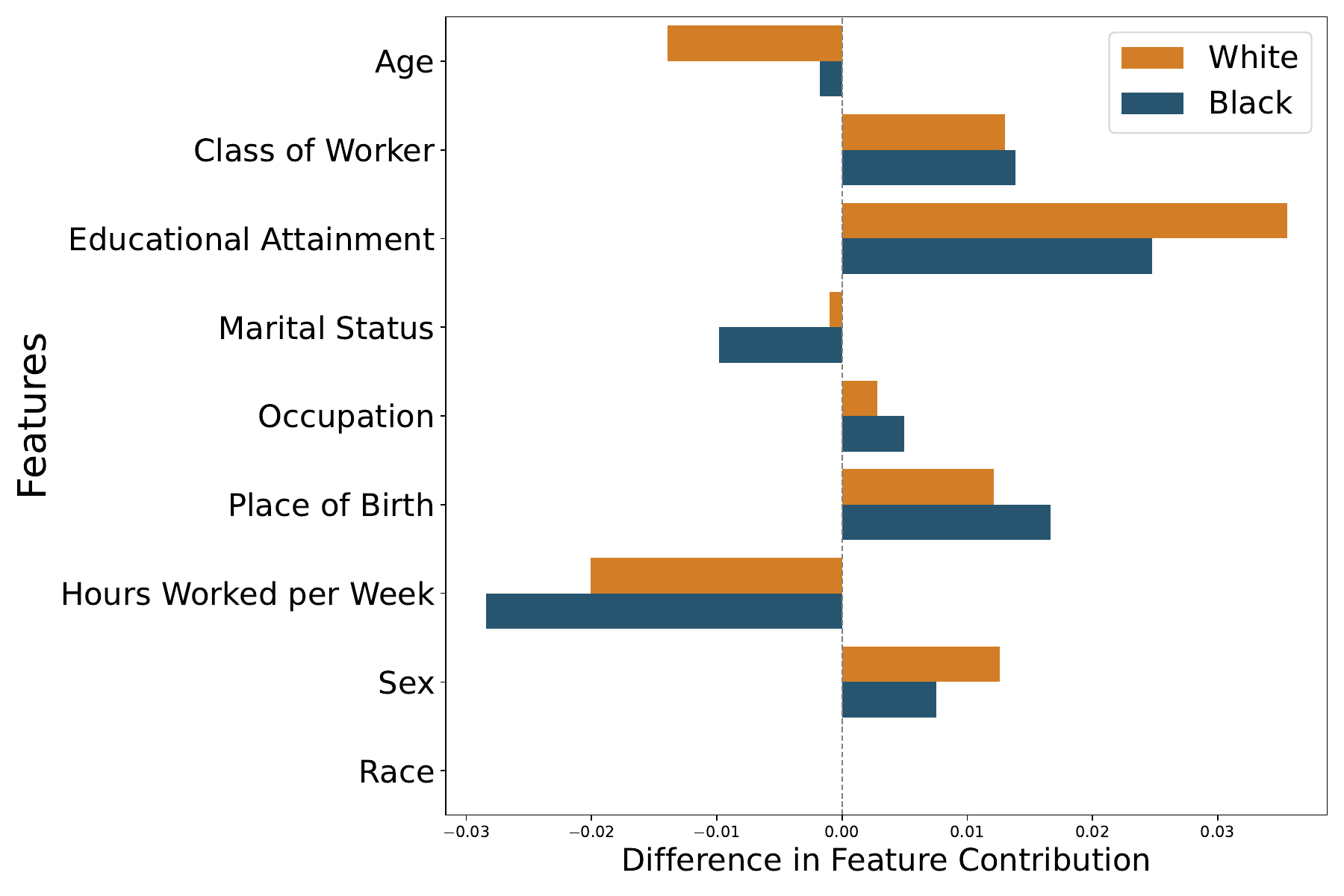} 
    \end{subfigure}
    \begin{subfigure}[b]{0.32\textwidth}
        \includegraphics[width=\linewidth]{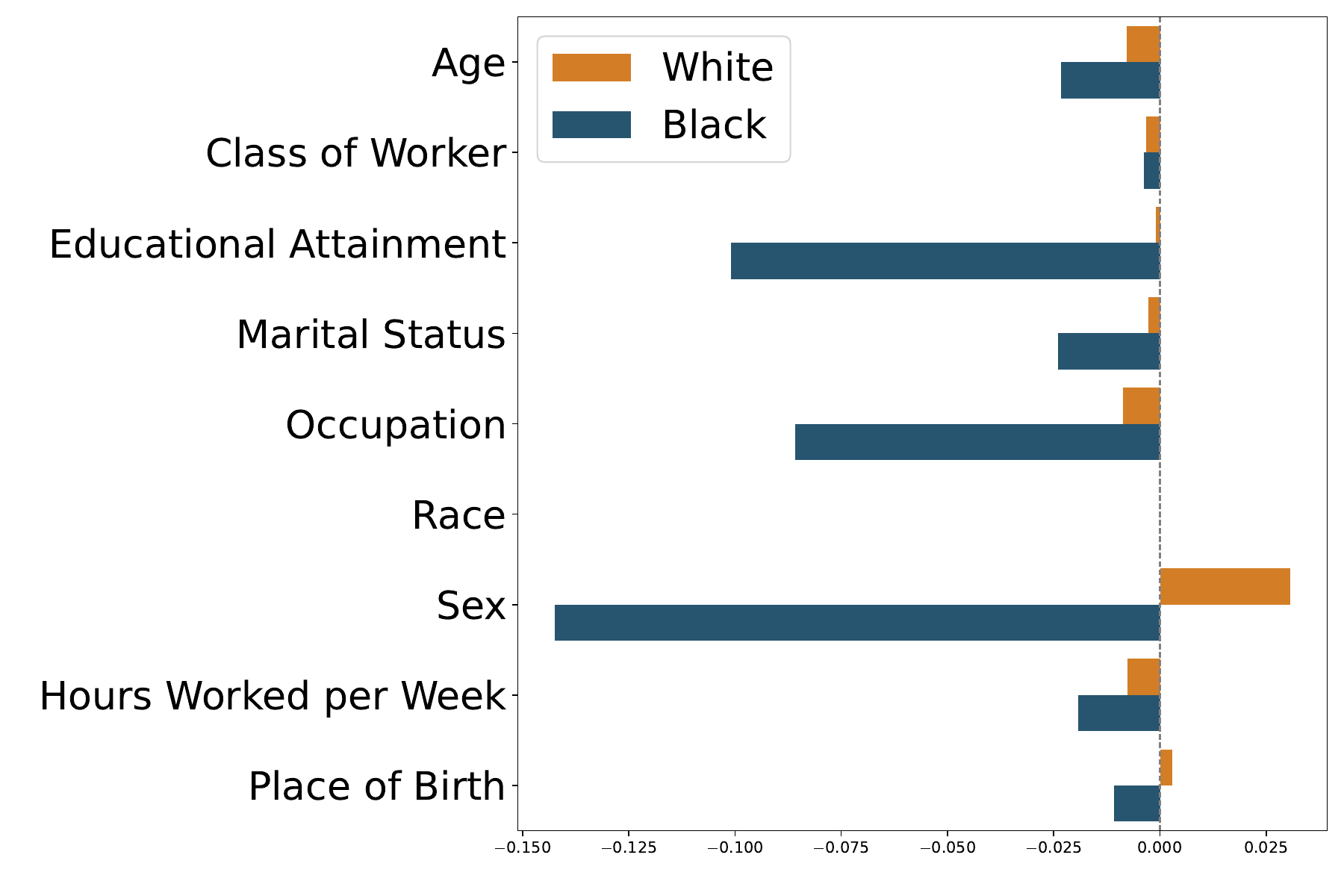}   
    \end{subfigure}
    \begin{subfigure}[b]{0.32\textwidth}
        \includegraphics[width=\linewidth]{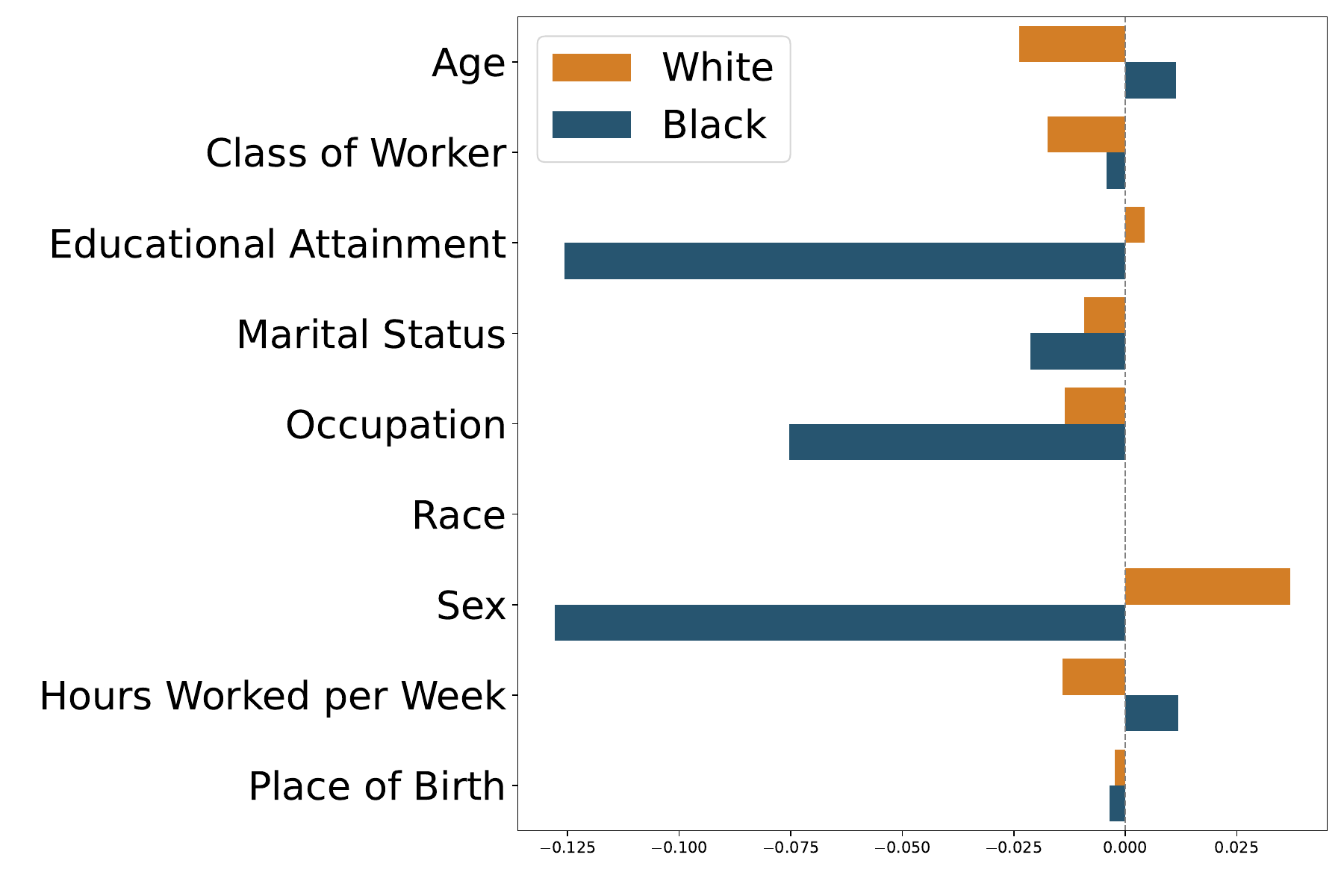}  
    \end{subfigure}
    \begin{subfigure}[b]{0.32\textwidth}
        \includegraphics[width=\linewidth]{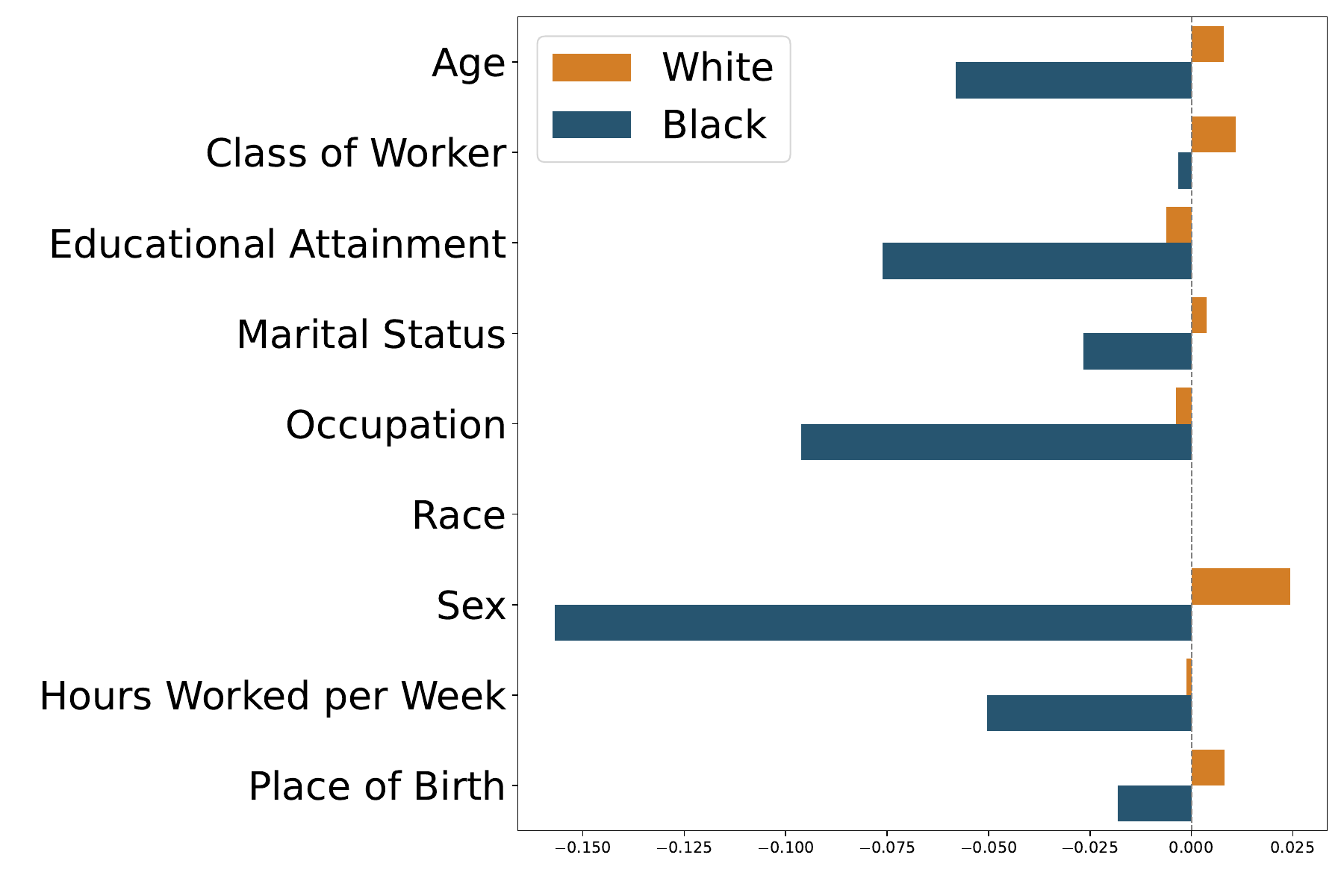}
    \end{subfigure}
    \caption{Differences of mean contributions with SHAP when the protected attribute race is removed for the Adult, AdultCA, and AdultLA datasets. Rows represent the datasets, while columns correspond to P, TP, and FP.}
    \label{fig:diff_SHAP_race}
\end{figure}

\begin{figure}[h]
    \centering
    \begin{subfigure}[b]{0.32\textwidth}
        \includegraphics[width=\linewidth]{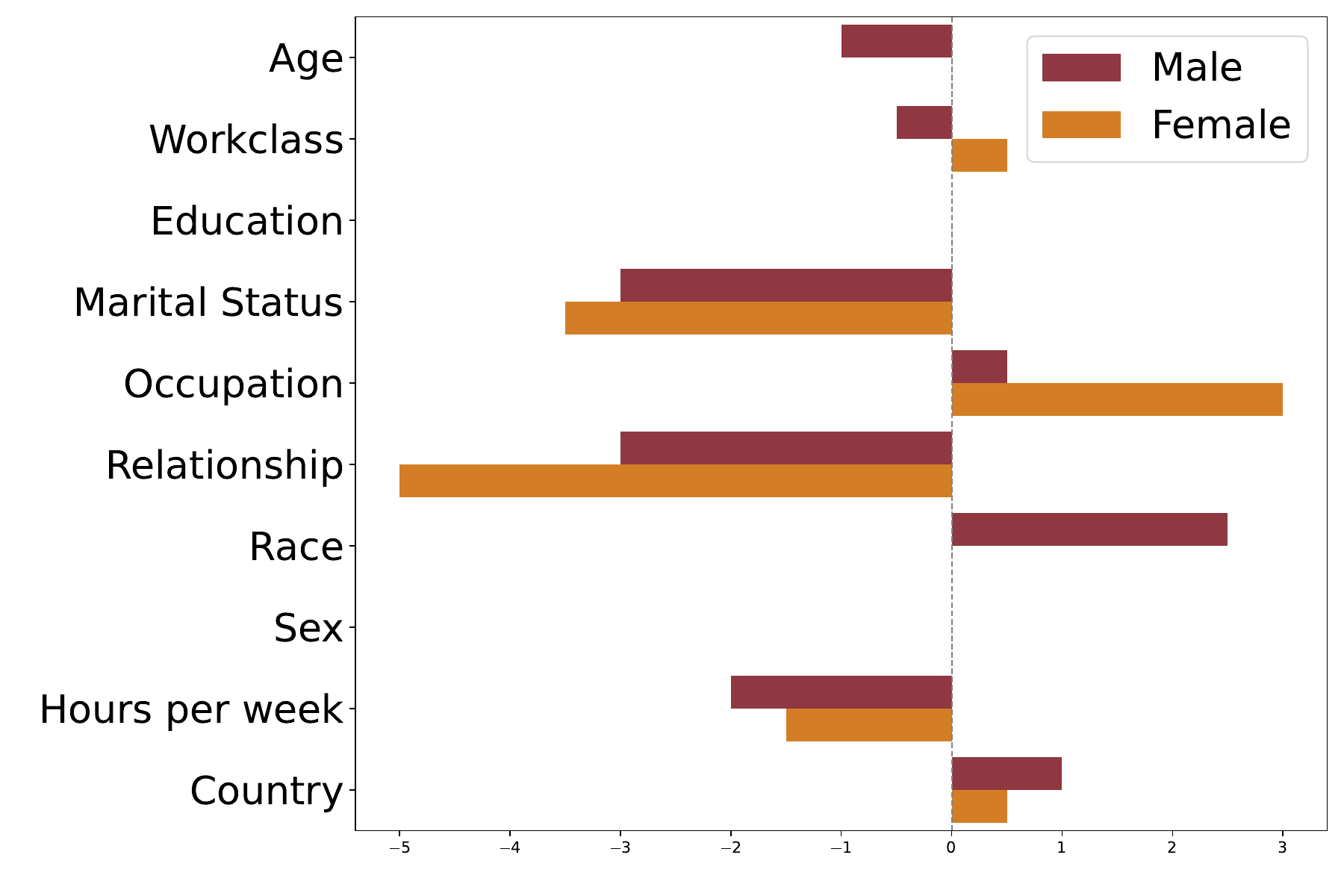}
    \end{subfigure}
    \begin{subfigure}[b]{0.32\textwidth}
        \includegraphics[width=\linewidth]{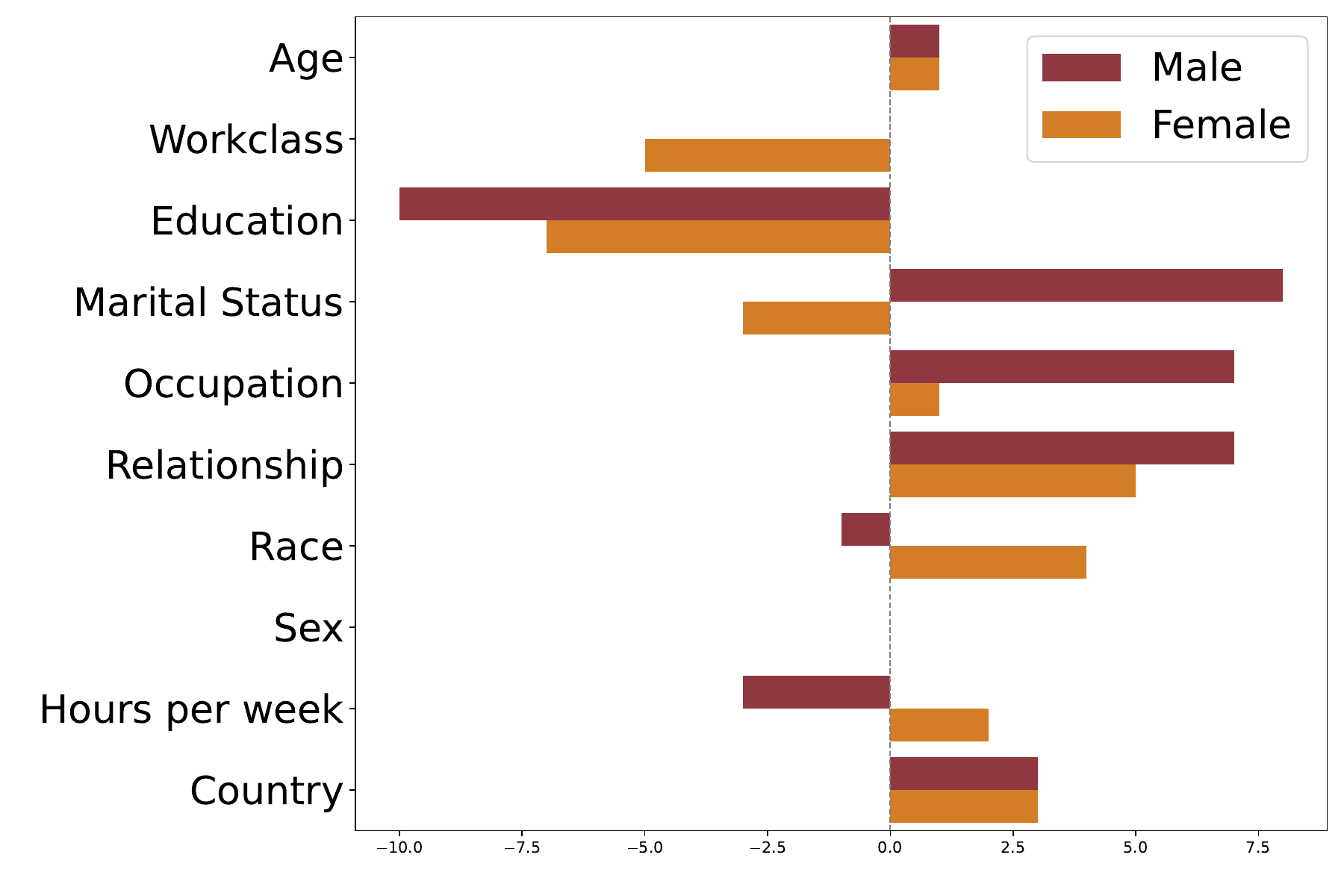}
    \end{subfigure}
    \begin{subfigure}[b]{0.32\textwidth}
        \includegraphics[width=\linewidth]{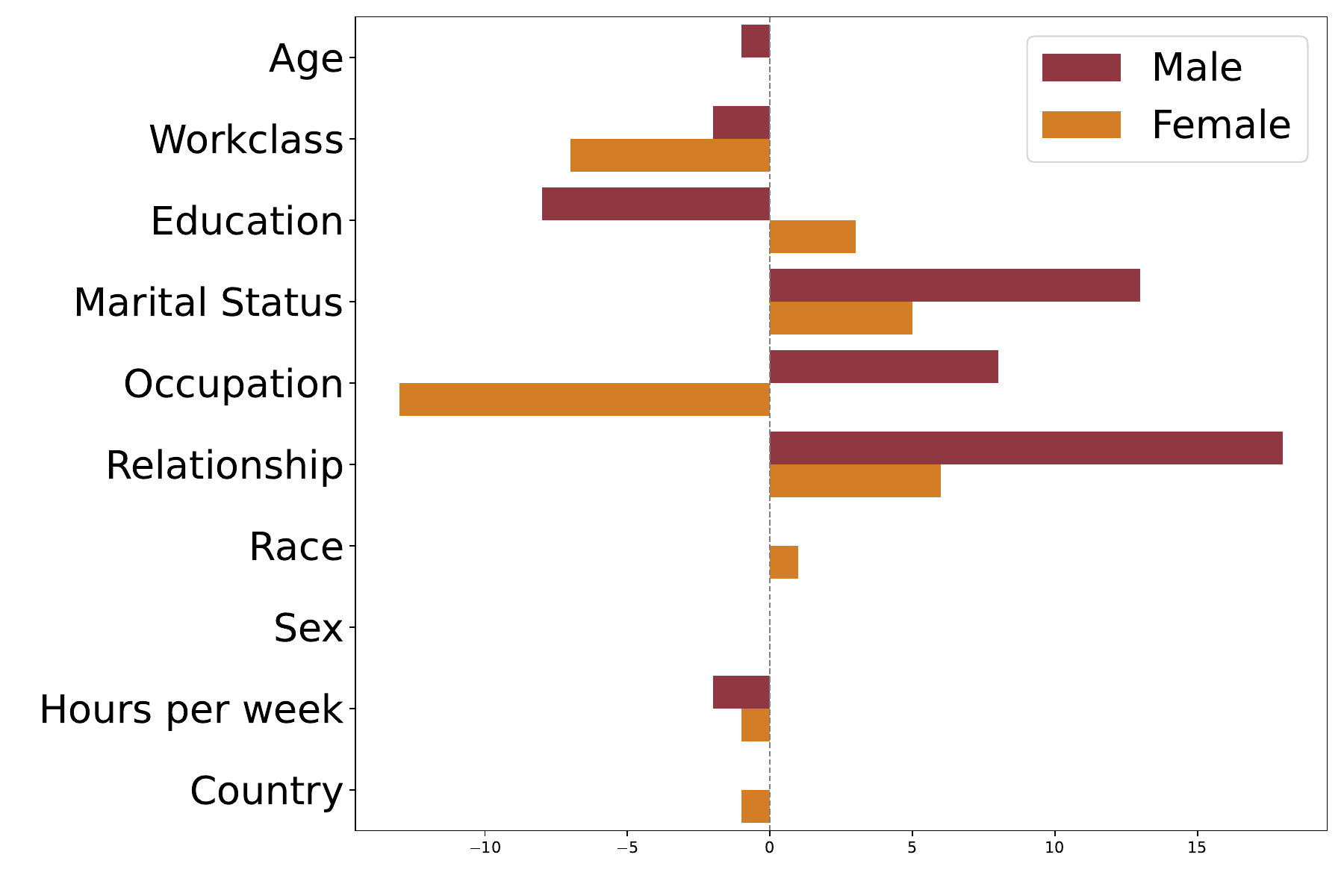}
    \end{subfigure}
    \begin{subfigure}[b]{0.32\textwidth}
        \includegraphics[width=\linewidth]{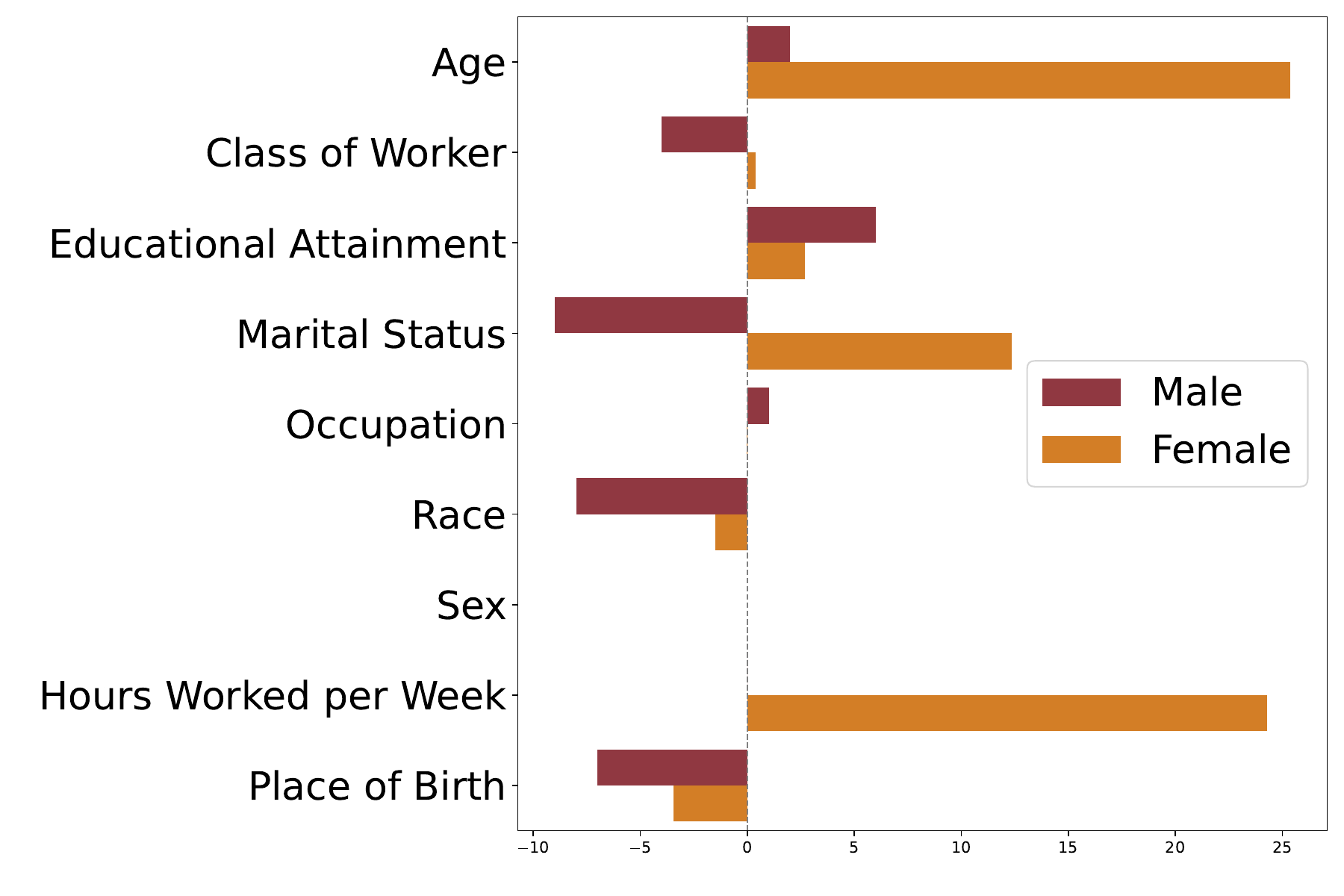} 
    \end{subfigure}
    \begin{subfigure}[b]{0.32\textwidth}
        \includegraphics[width=\linewidth]{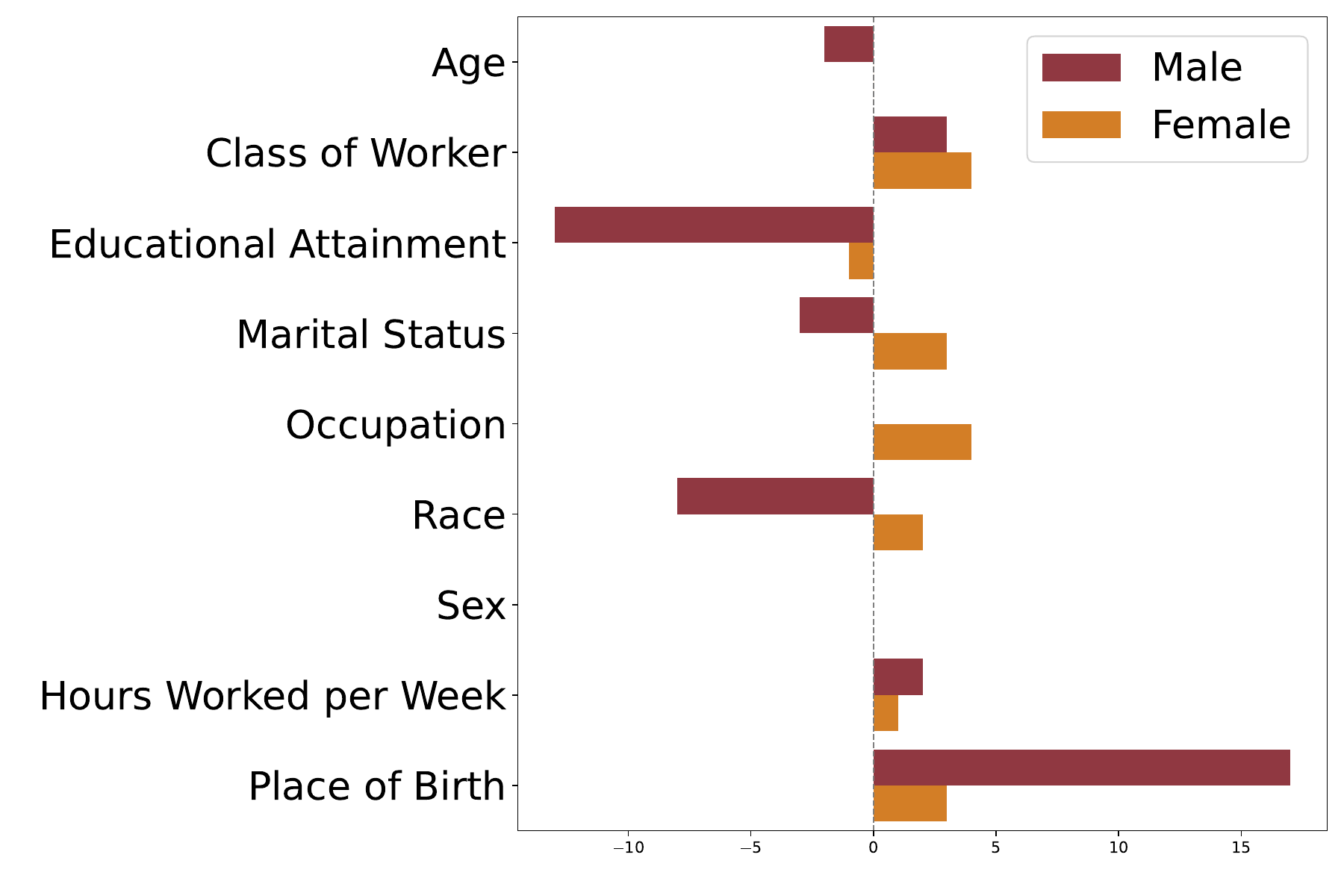}        
    \end{subfigure}
    \begin{subfigure}[b]{0.32\textwidth}
        \includegraphics[width=\linewidth]{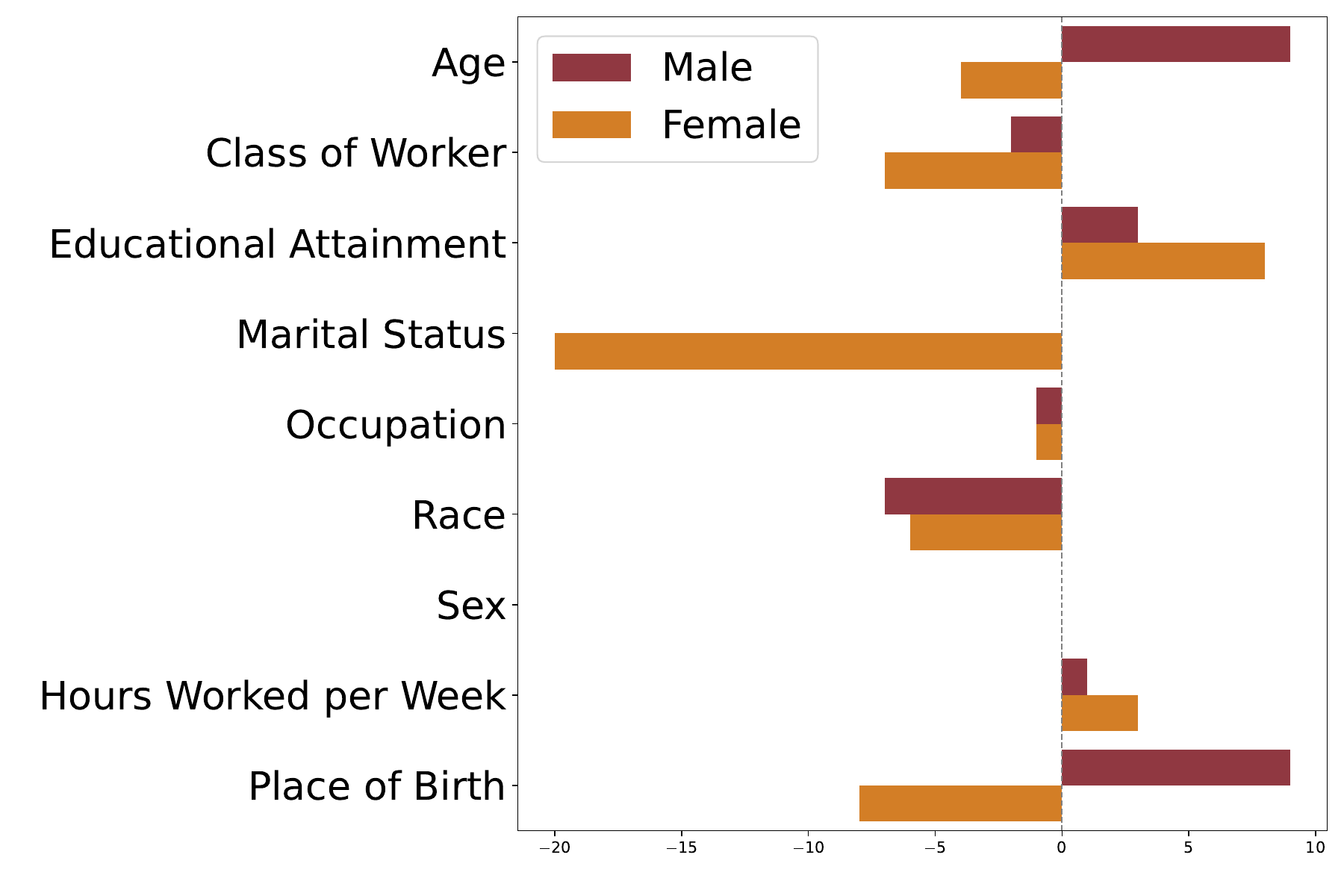}        
    \end{subfigure}  
    \begin{subfigure}[b]{0.32\textwidth}
        \includegraphics[width=\linewidth]{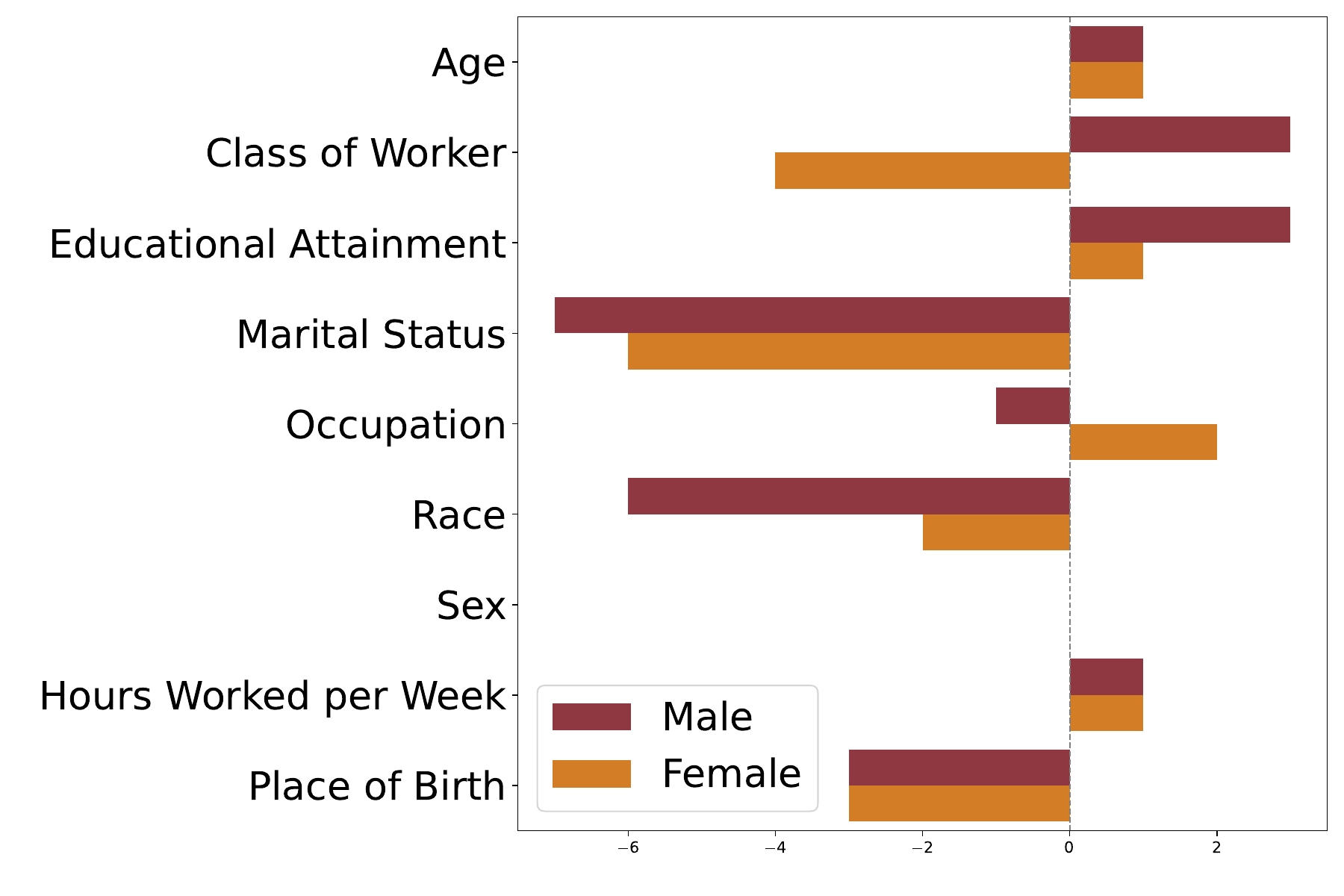} 
    \end{subfigure}
    \begin{subfigure}[b]{0.32\textwidth}
        \includegraphics[width=\linewidth]{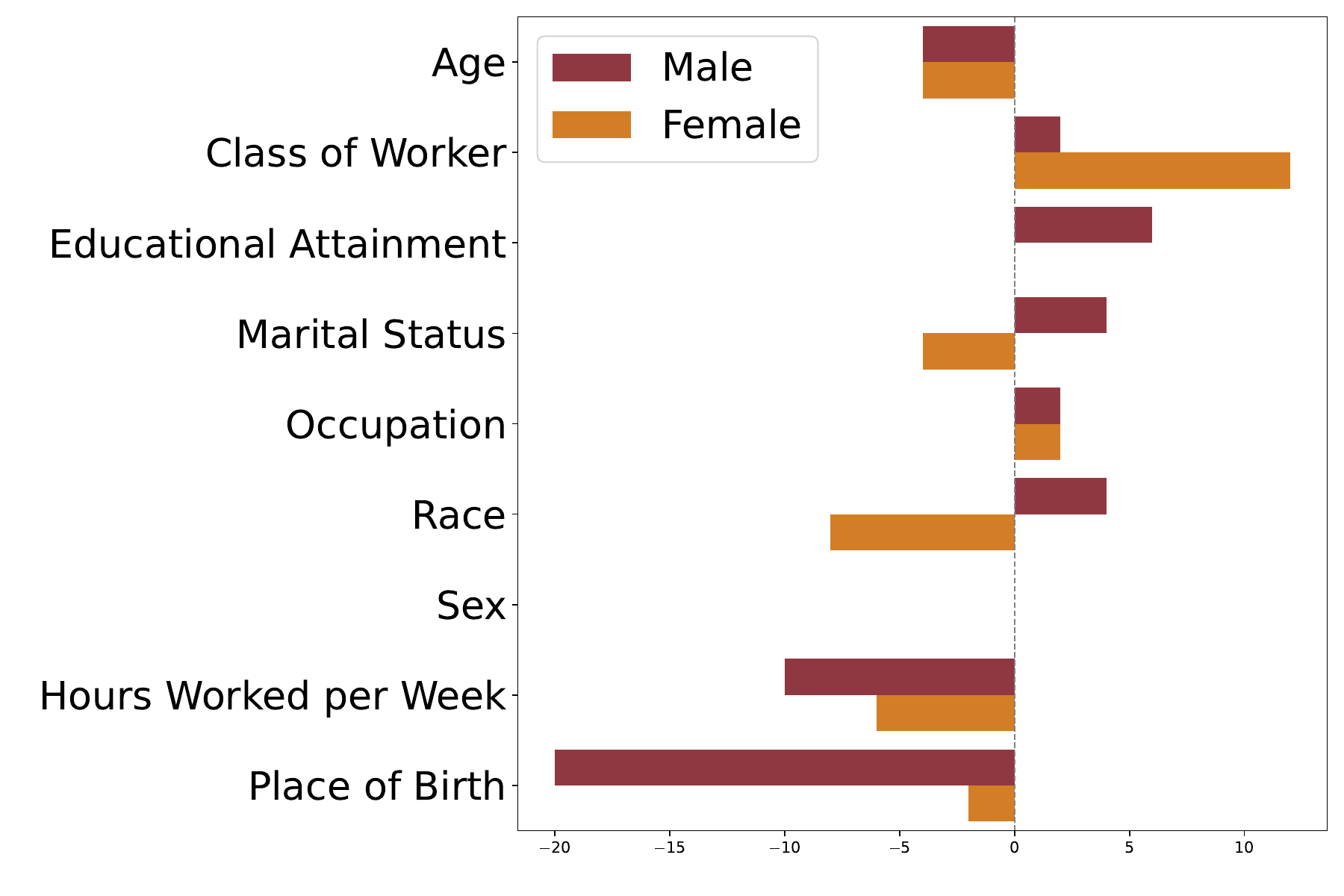}
    \end{subfigure}
    \begin{subfigure}[b]{0.32\textwidth}
        \includegraphics[width=\linewidth]{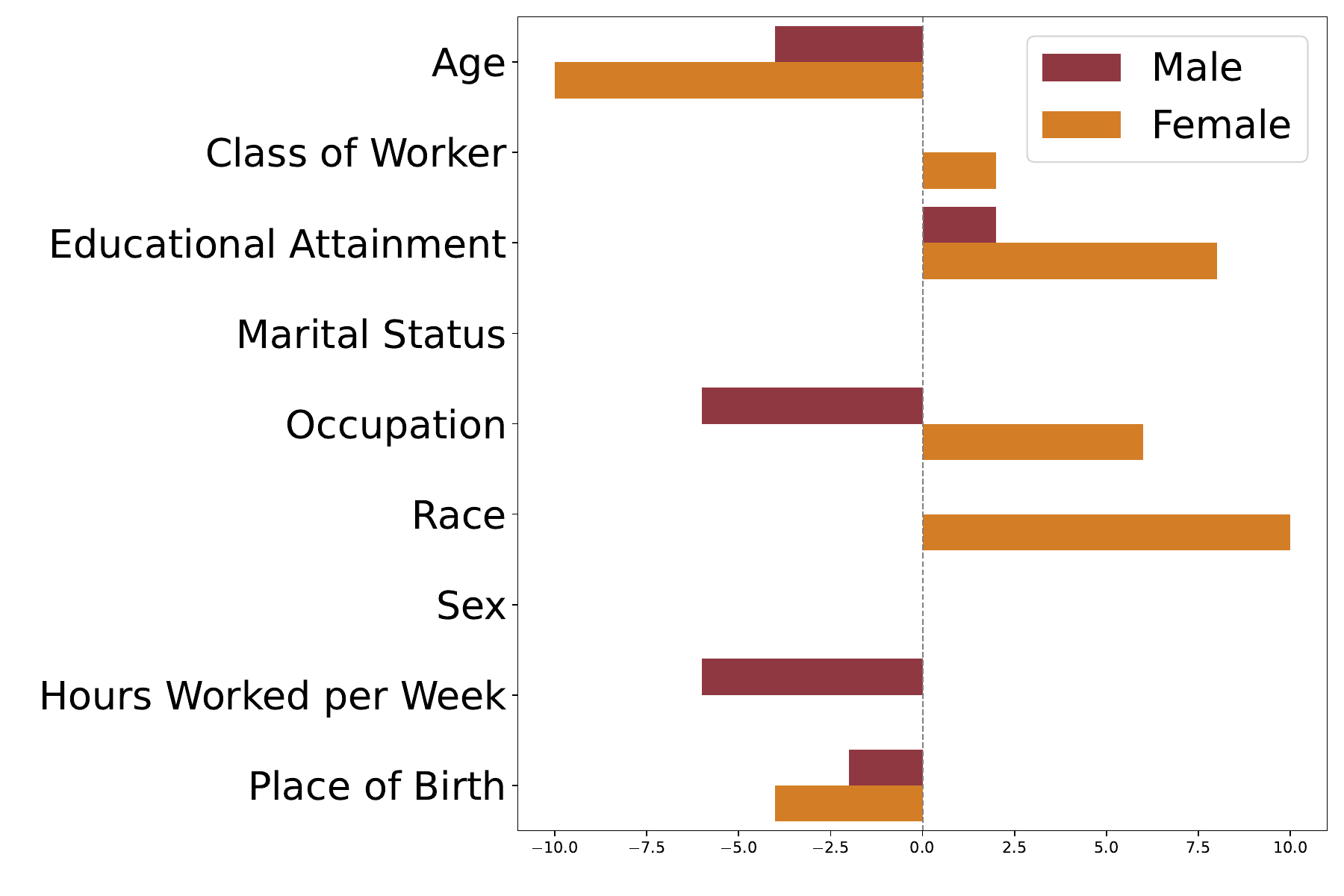}   
    \end{subfigure}
    \caption{Differences of mean contributions with DiCE when the protected attribute race is removed for the Adult, AdultCA, and AdultLA datasets. Rows represent the datasets, while columns correspond to N, FN, and TN.}
     \label{fig:diff_DiCE}
\end{figure}

\begin{figure}[h]
    \centering
    \begin{subfigure}[b]{0.32\textwidth}
        \includegraphics[width=\linewidth]{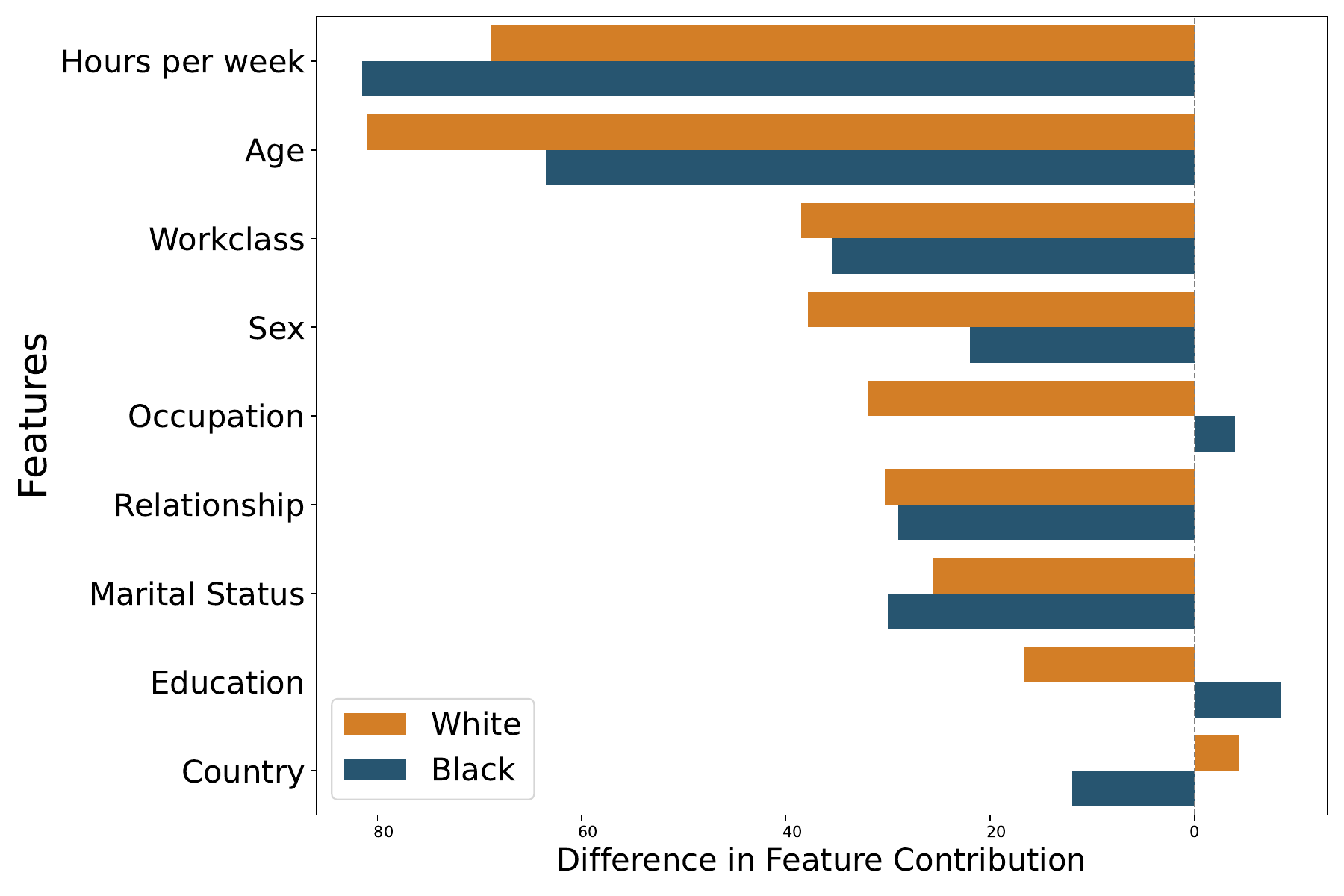}
    \end{subfigure}
    \begin{subfigure}[b]{0.32\textwidth}
        \includegraphics[width=\linewidth]{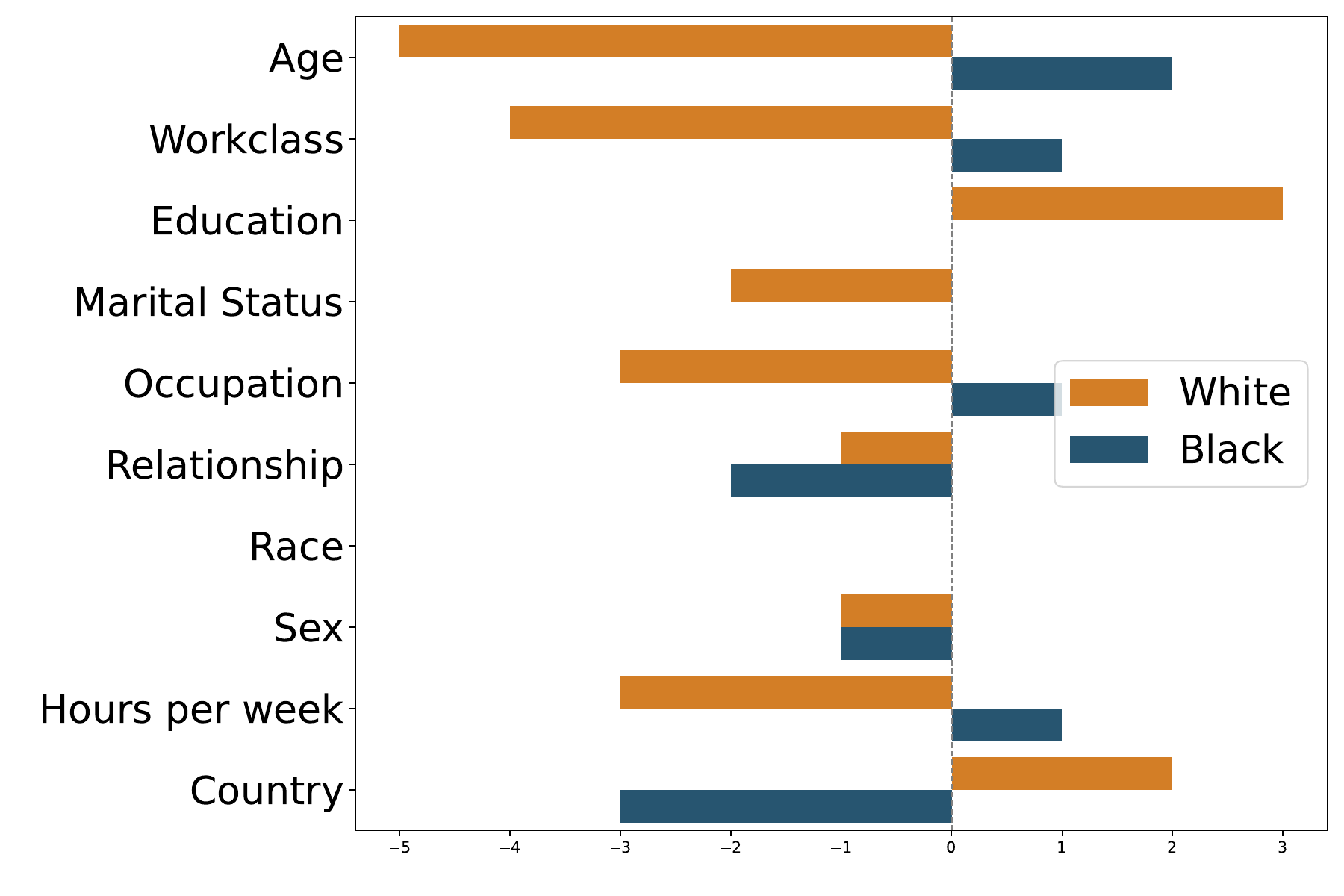}
    \end{subfigure}
    \begin{subfigure}[b]{0.32\textwidth}
        \includegraphics[width=\linewidth]{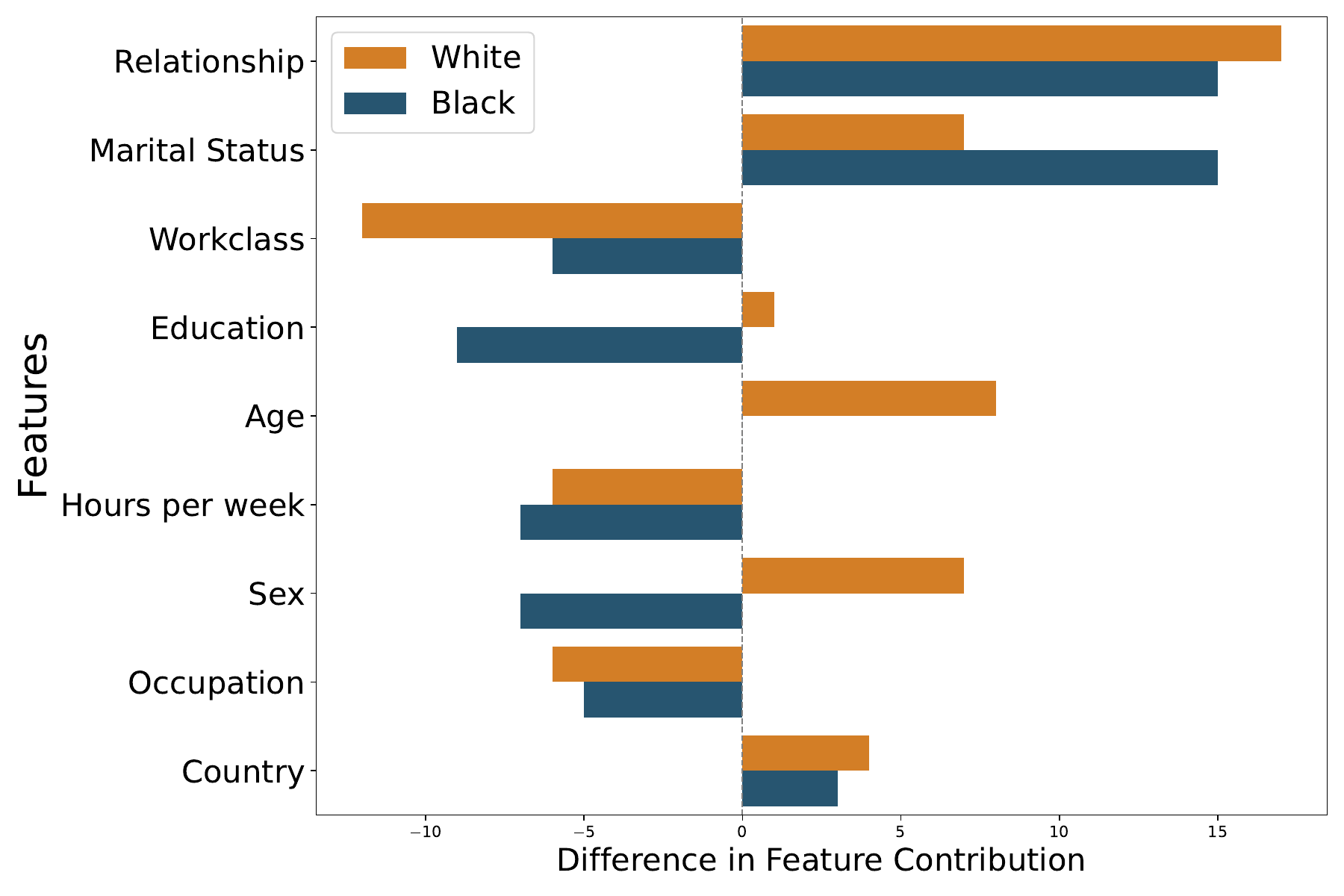}
    \end{subfigure}
    \begin{subfigure}[b]{0.32\textwidth}
        \includegraphics[width=\linewidth]{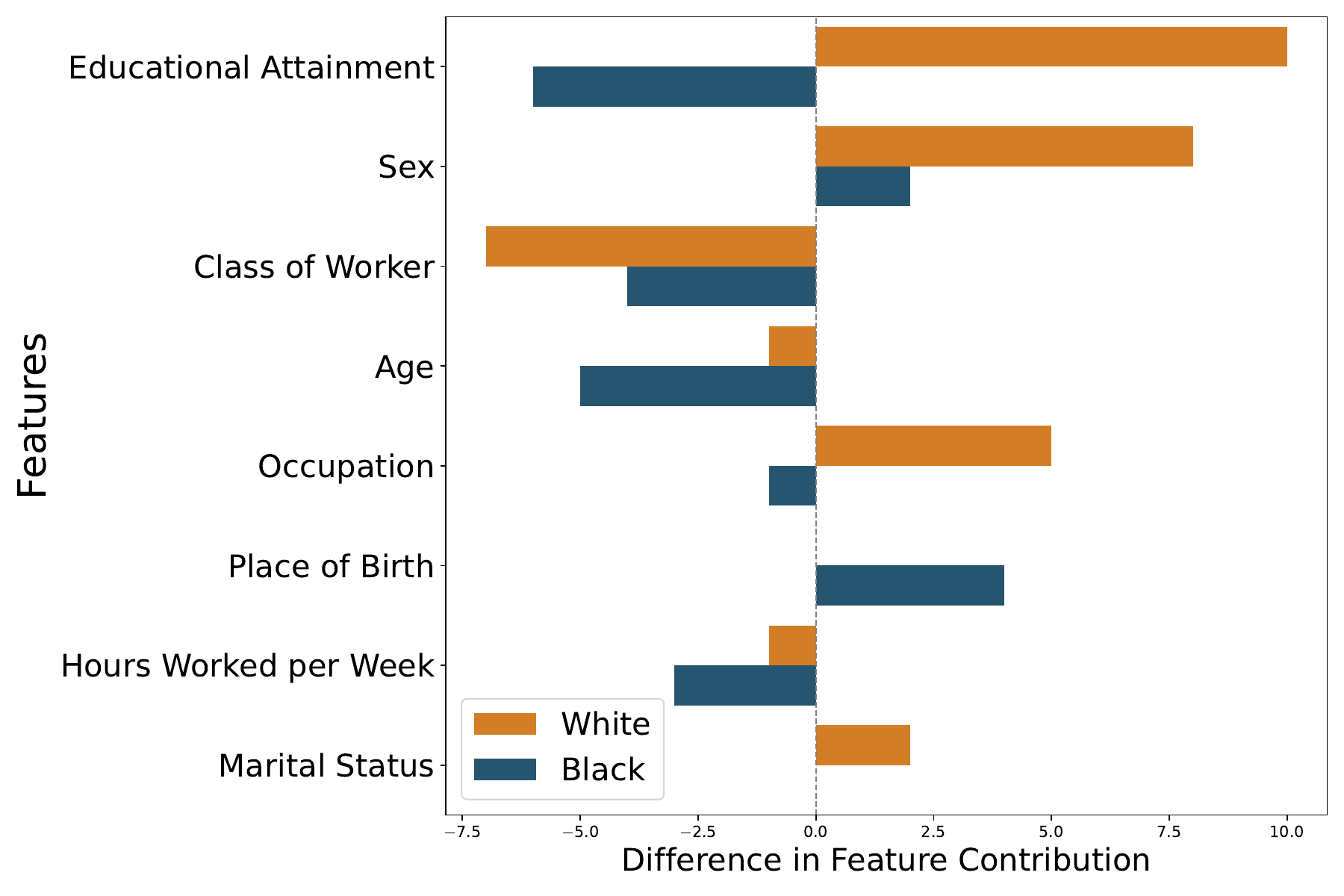}
      
    \end{subfigure}
    \begin{subfigure}[b]{0.32\textwidth}
        \includegraphics[width=\linewidth]{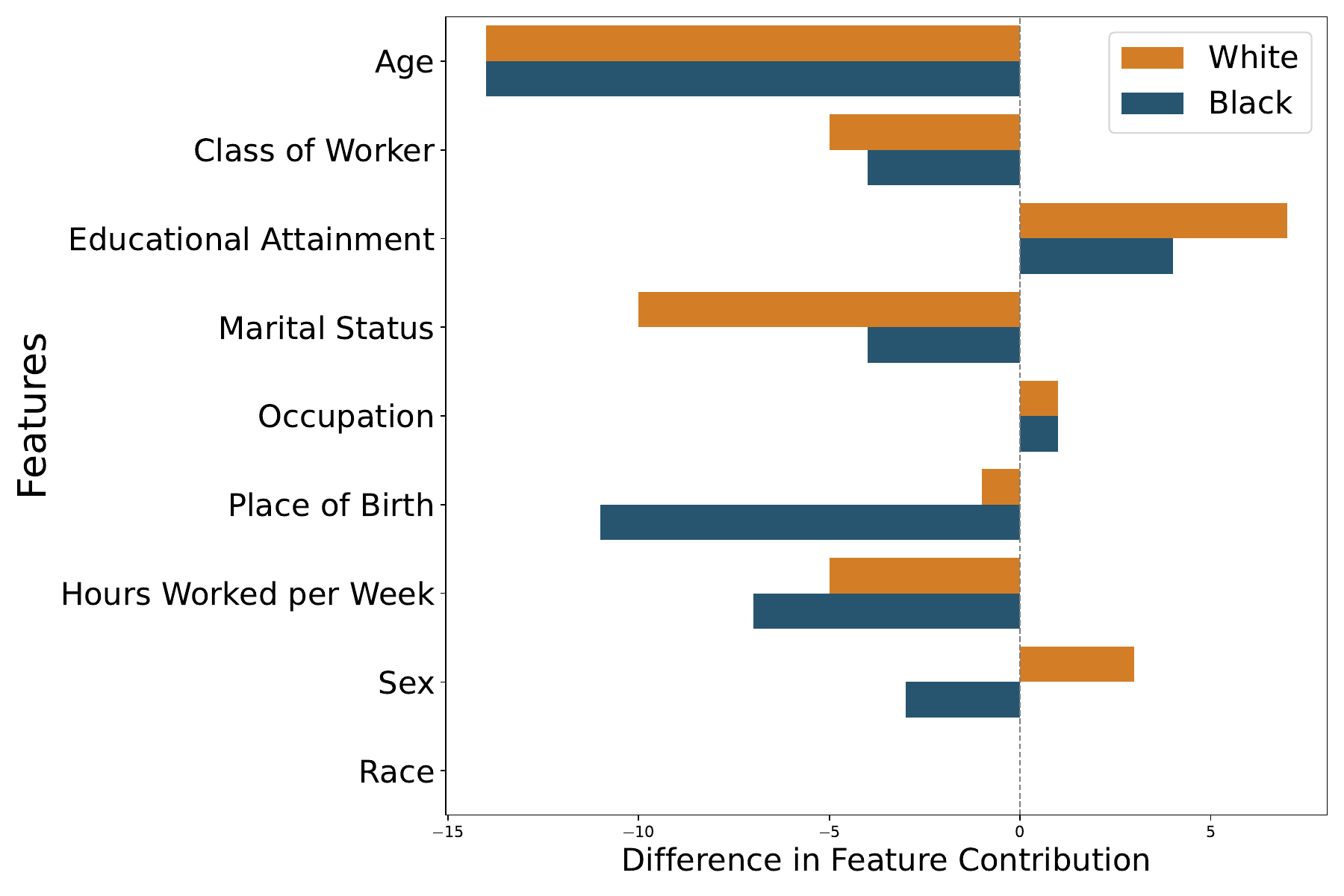}
        
    \end{subfigure}
    \begin{subfigure}[b]{0.32\textwidth}
        \includegraphics[width=\linewidth]{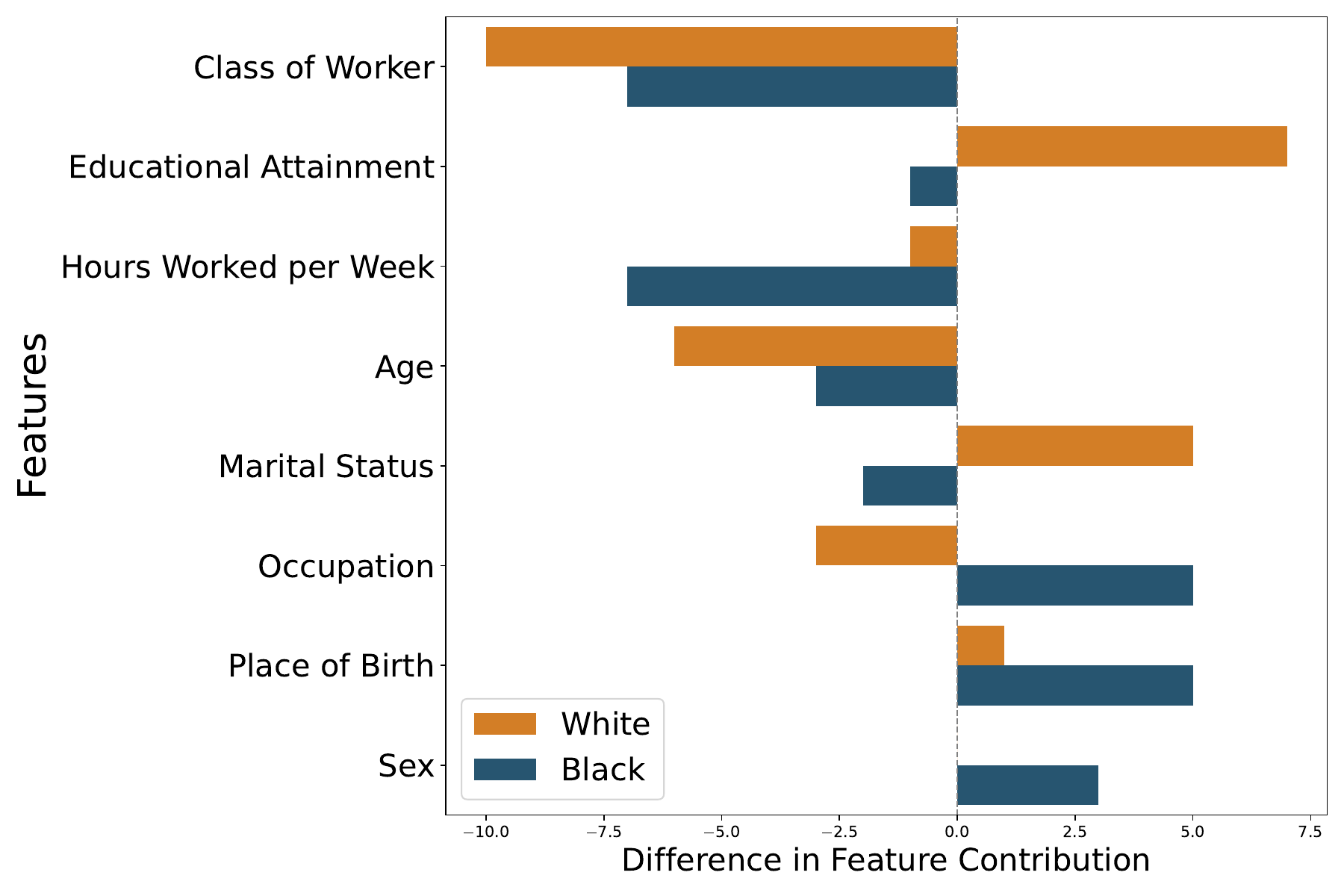}
    \end{subfigure}
    \begin{subfigure}[b]{0.32\textwidth}
        \includegraphics[width=\linewidth]{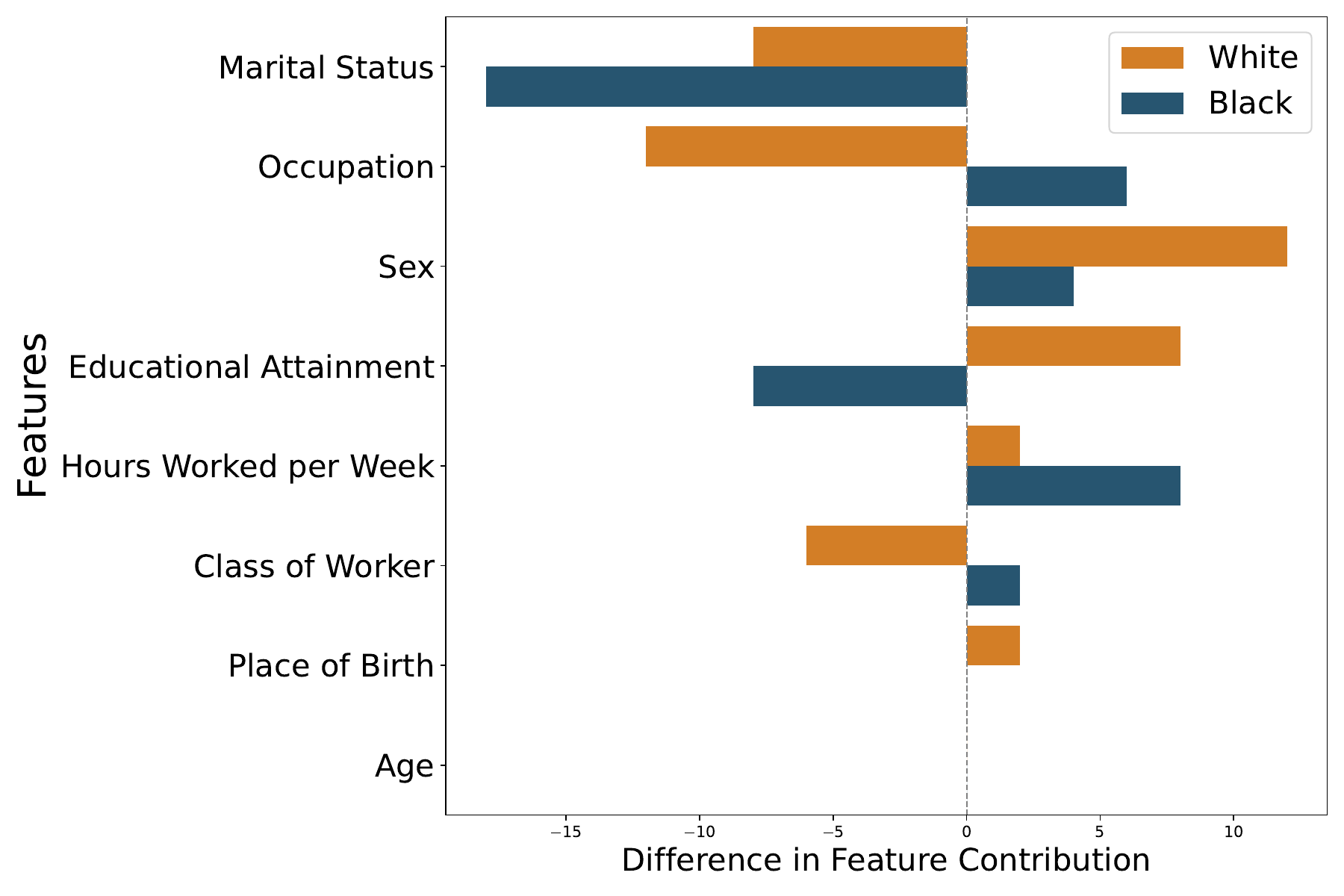}
    \end{subfigure}
    \begin{subfigure}[b]{0.32\textwidth}
        \includegraphics[width=\linewidth]{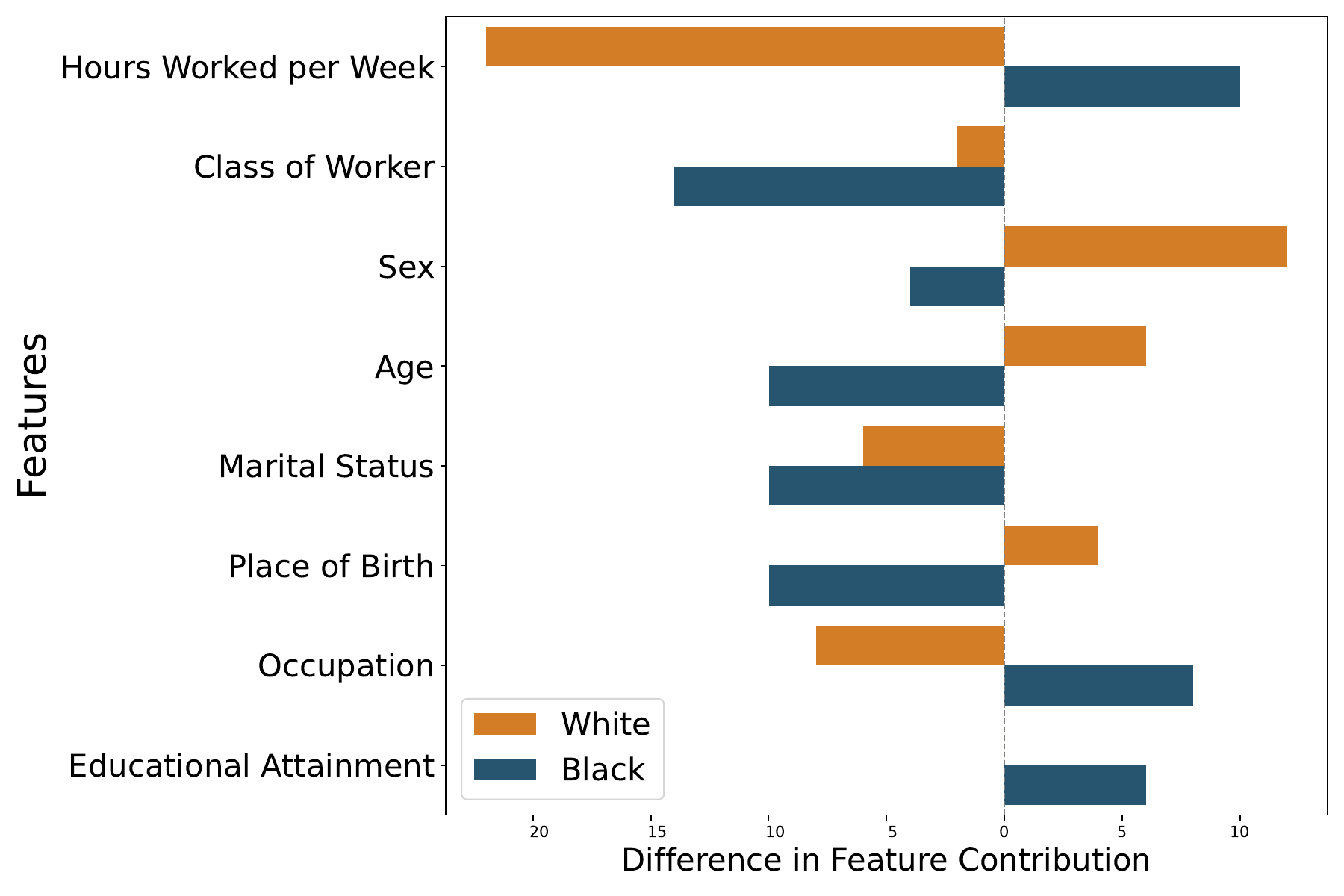}
    \end{subfigure}
    \begin{subfigure}[b]{0.32\textwidth}
        \includegraphics[width=\linewidth]{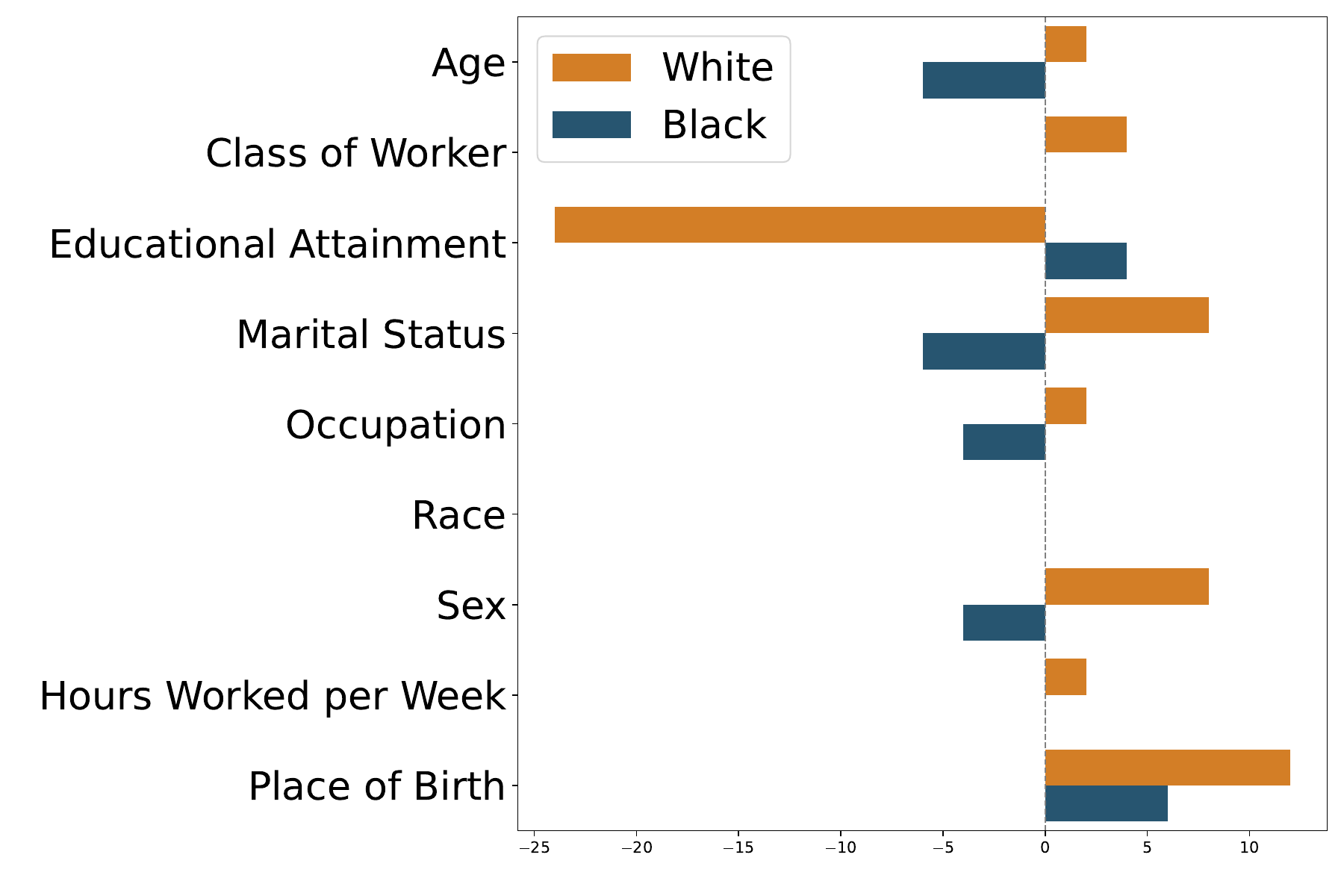}  
    \end{subfigure}
    \caption{Differences of mean contributions with DiCE when the protected attribute race is removed for the Adult, AdultCA, and AdultLA datasets. Rows represent the datasets, while columns correspond to N, FN, and TN.}
    \label{fig:diff_DiCE_race}
\end{figure}

Figure~\ref{fig:diff_LIME} shows the feature contributions from LIME for a model trained without the sex attribute. In the Adult dataset, we observe that features such as Marital Status and Relationship, both indirectly related to gender, show an increase in importance. Additionally, the contribution of race slightly increases for female instances, particularly in positive predictions. Similarly, Figures~\ref{fig:diff_LIME} and~\ref{fig:diff_SHAP_race} present the impact of removing the race attribute using LIME and SHAP, respectively. Figures~\ref{fig:diff_DiCE} and~\ref{fig:diff_DiCE_race} illustrate changes in mean contributions with DiCE when sex and race are excluded. Across all explanation methods, we consistently find that removing a protected attribute shifts the influence toward correlated or unrelated features, often reinforcing advantages for the non-protected group.

\begin{figure}[h]
    \centering
    \begin{subfigure}[b]{0.32\textwidth}
        \includegraphics[width=\linewidth]{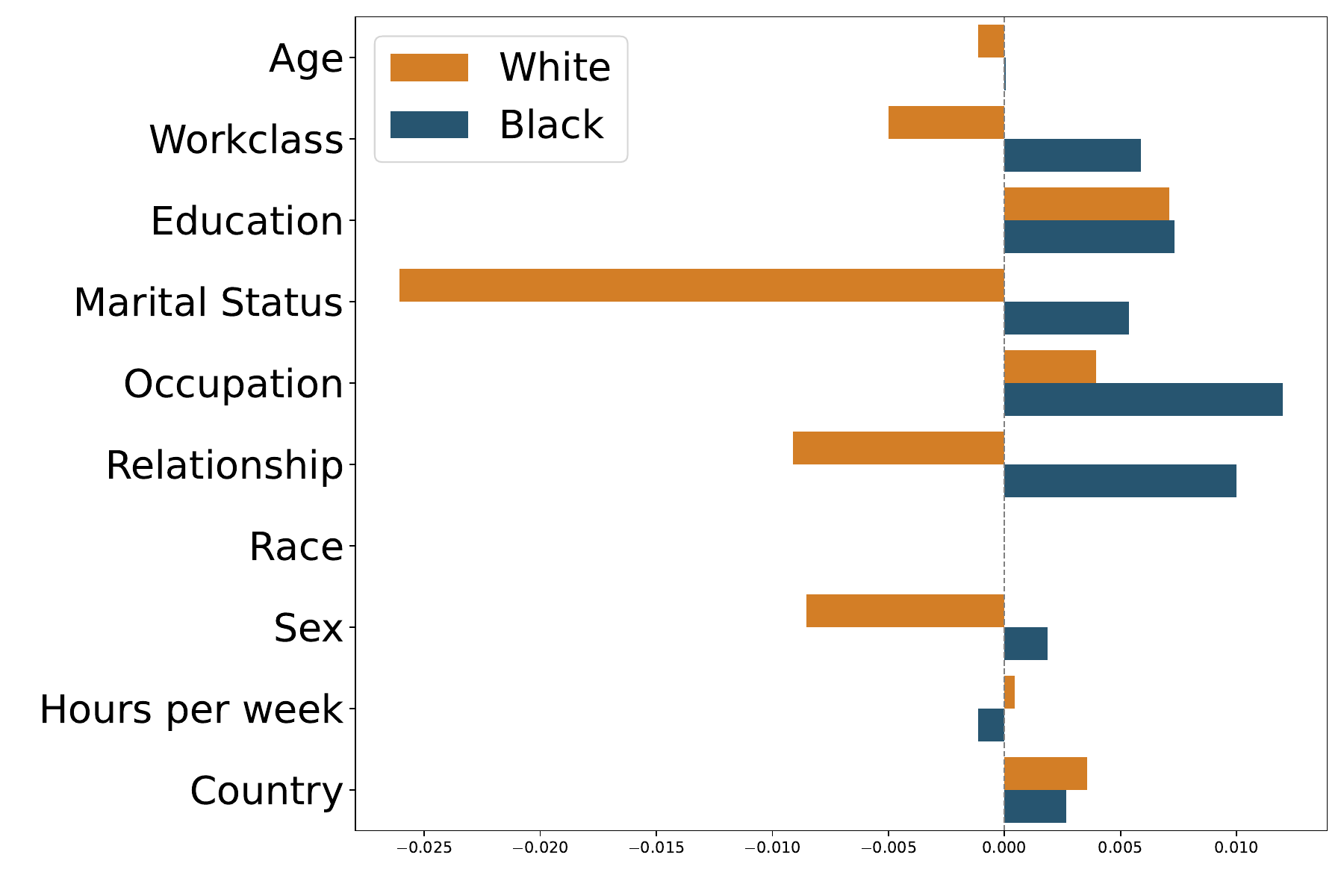}        
    \end{subfigure}
    \begin{subfigure}[b]{0.32\textwidth}
        \includegraphics[width=\linewidth]{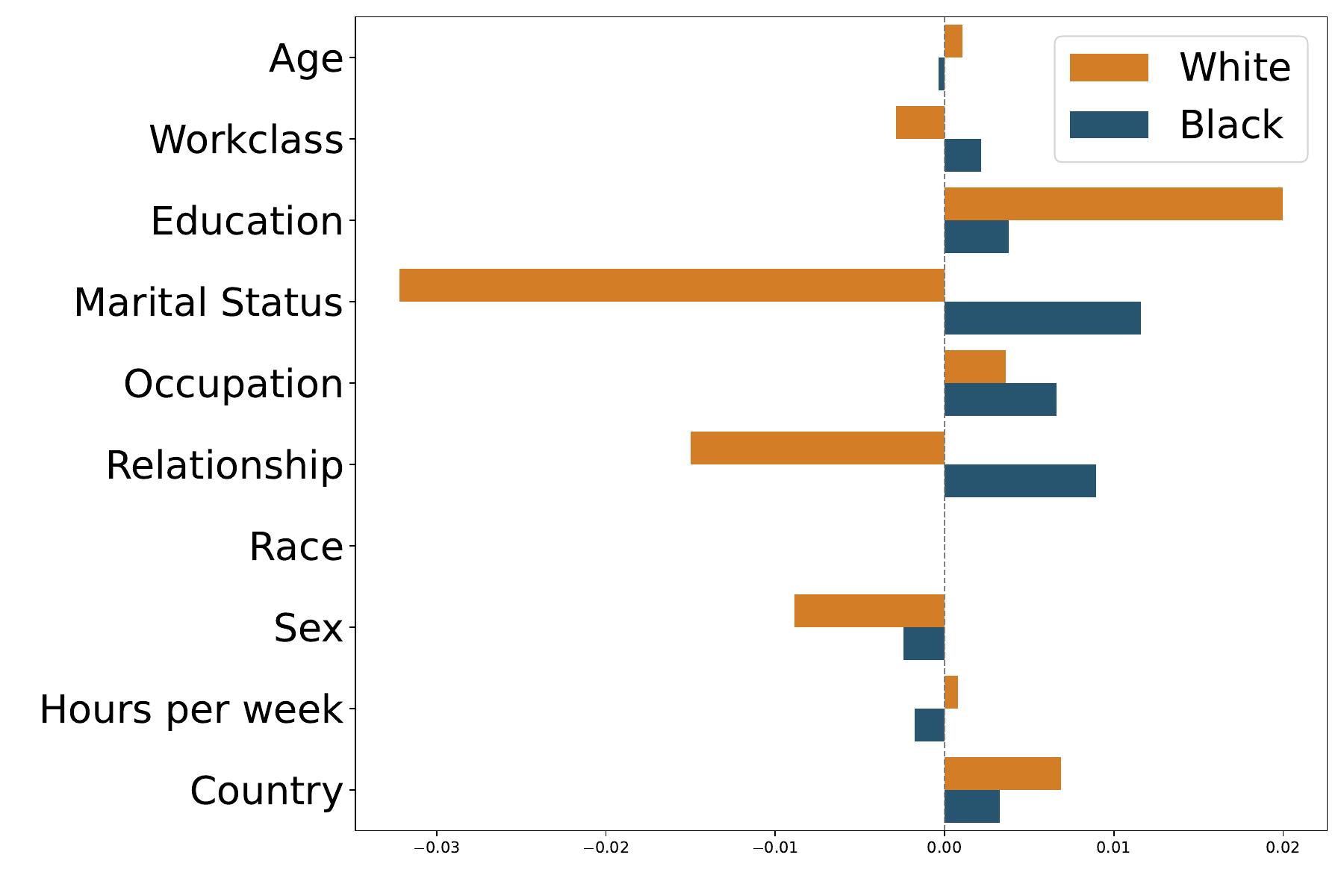}        
    \end{subfigure}
    \begin{subfigure}[b]{0.32\textwidth}
        \includegraphics[width=\linewidth]{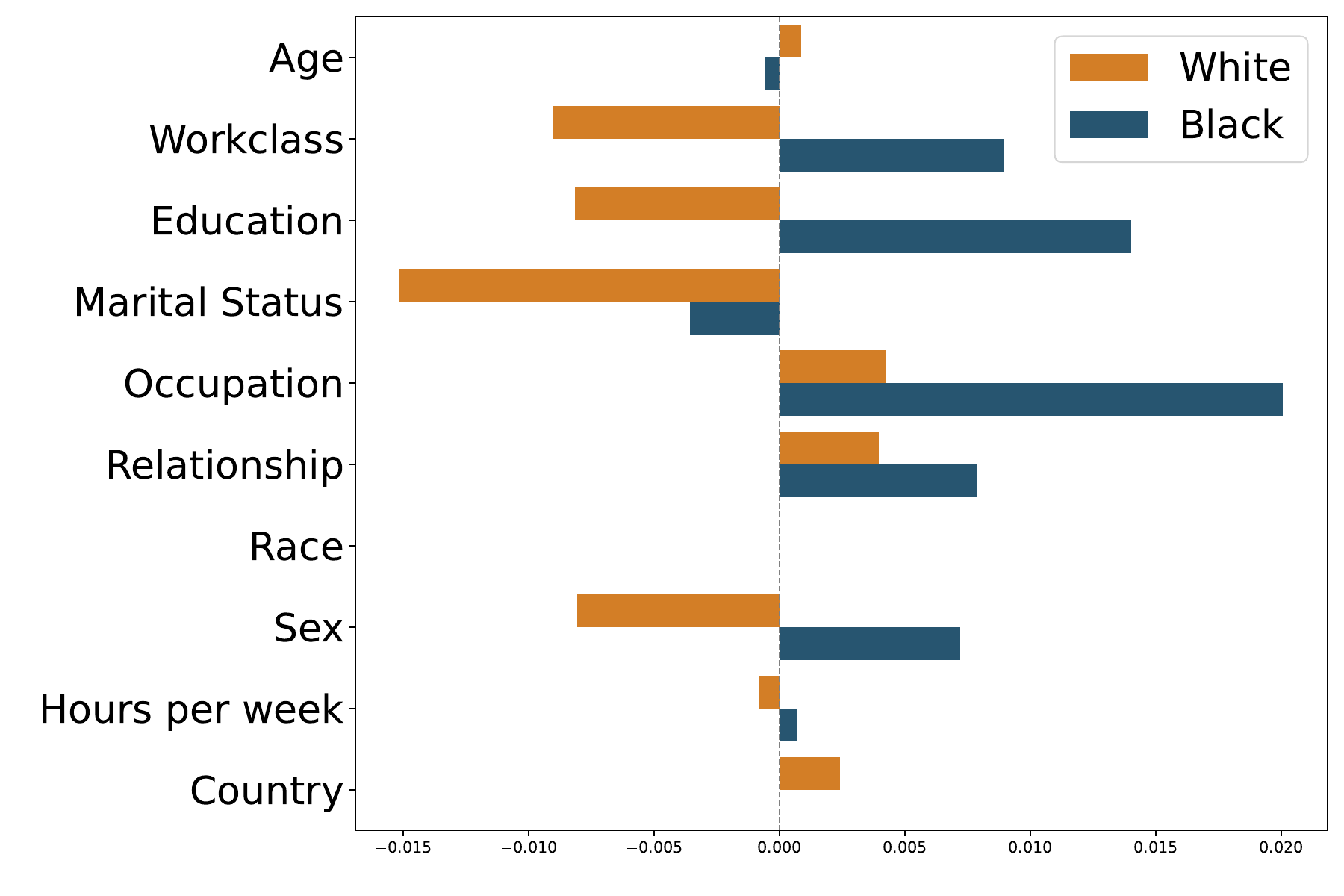}        
    \end{subfigure}   
    \begin{subfigure}[b]{0.32\textwidth}
        \includegraphics[width=\linewidth]{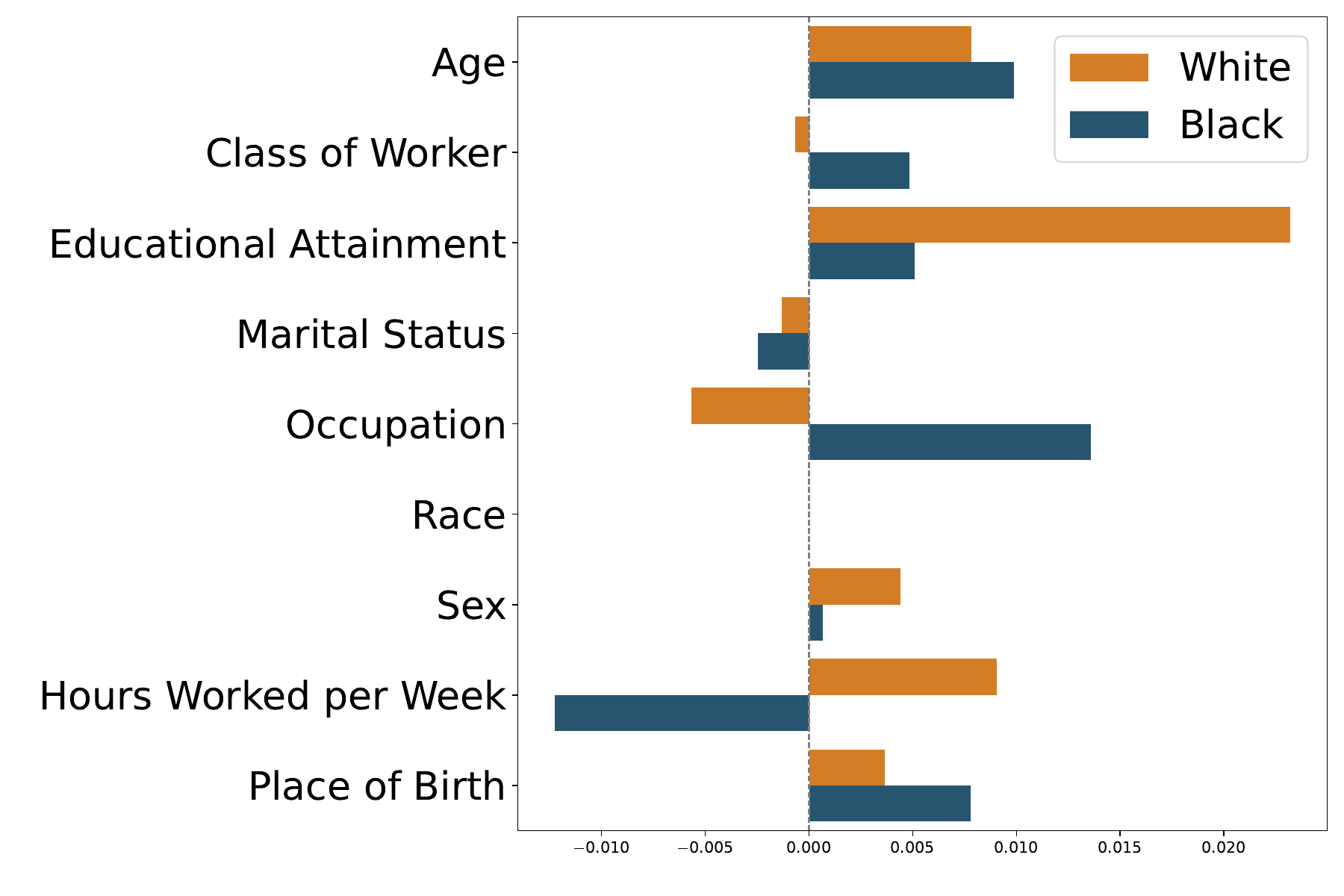}
    \end{subfigure}
    \begin{subfigure}[b]{0.32\textwidth}
        \includegraphics[width=\linewidth]{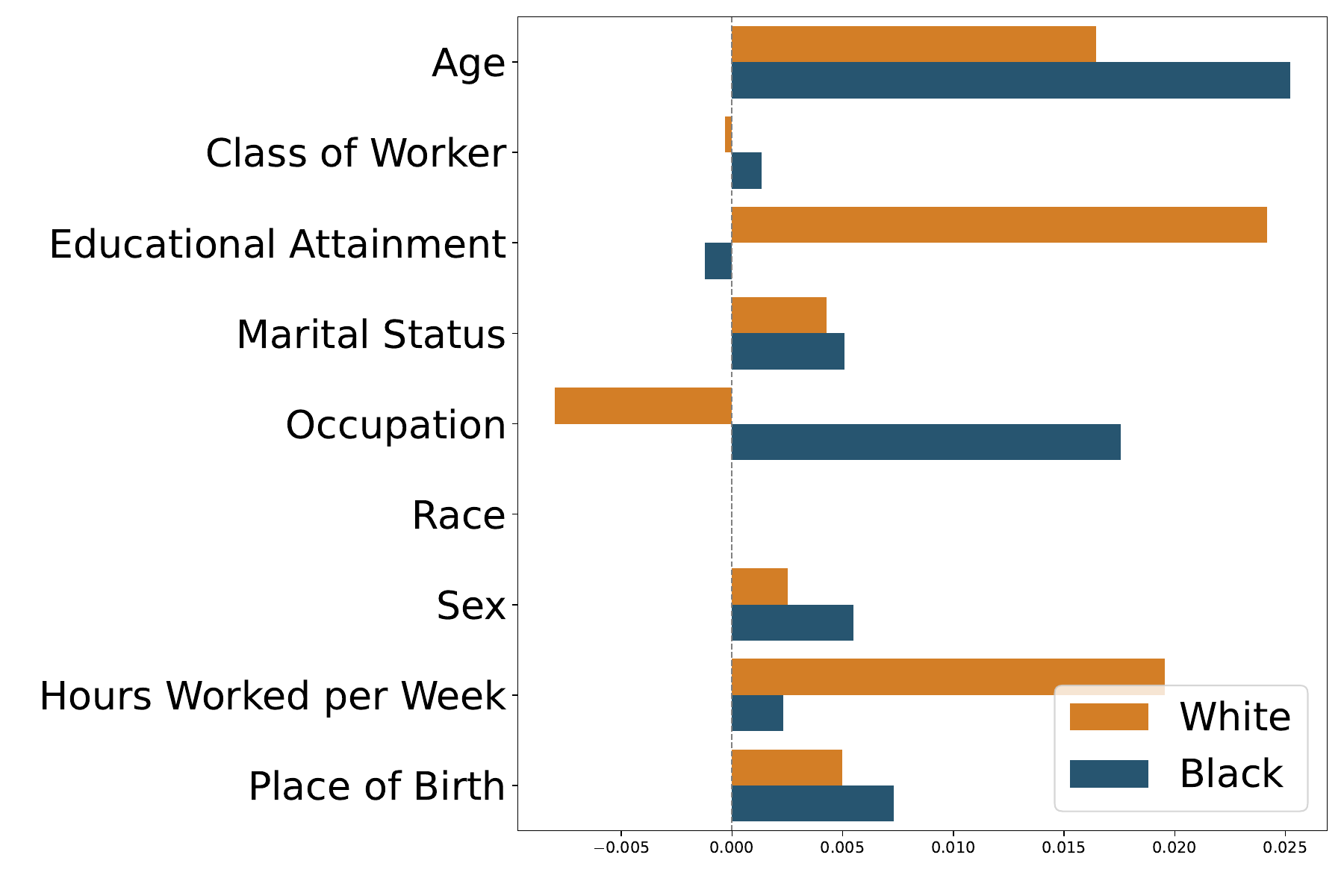}
    \end{subfigure}
    \begin{subfigure}[b]{0.32\textwidth}
        \includegraphics[width=\linewidth]{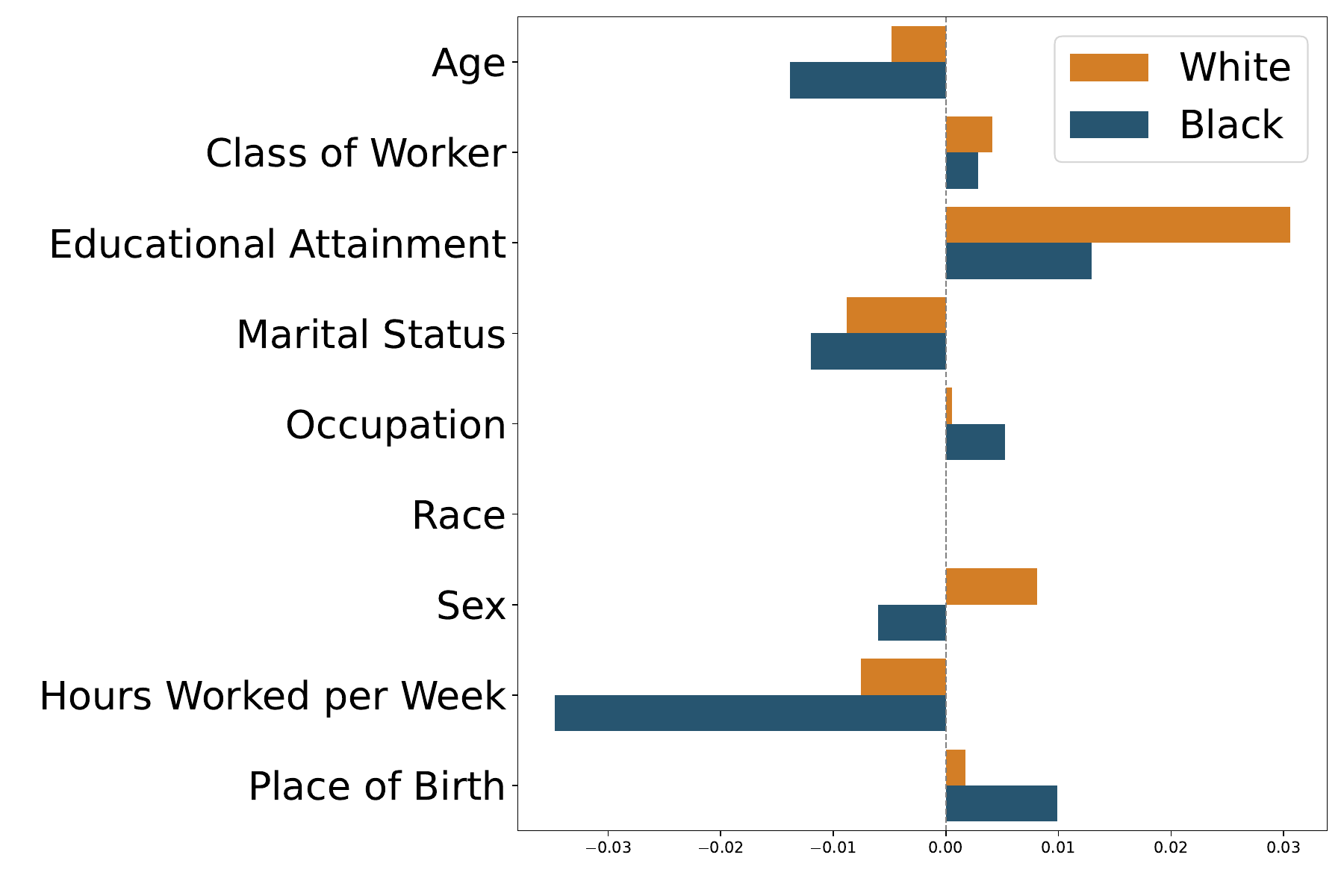}
    \end{subfigure}
    \begin{subfigure}[b]{0.32\textwidth}
        \includegraphics[width=\linewidth]{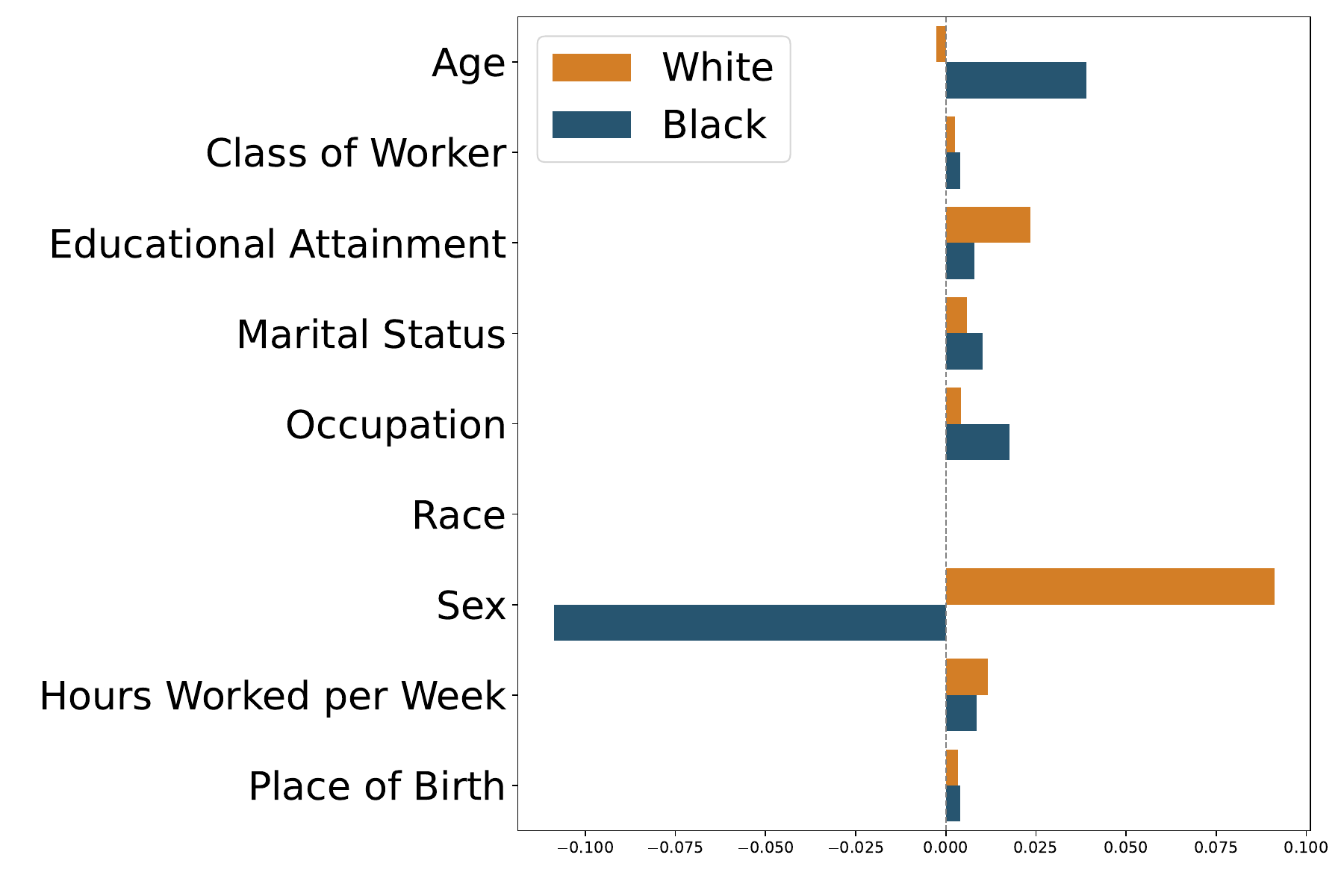}
    \end{subfigure}
    \begin{subfigure}[b]{0.32\textwidth}
        \includegraphics[width=\linewidth]{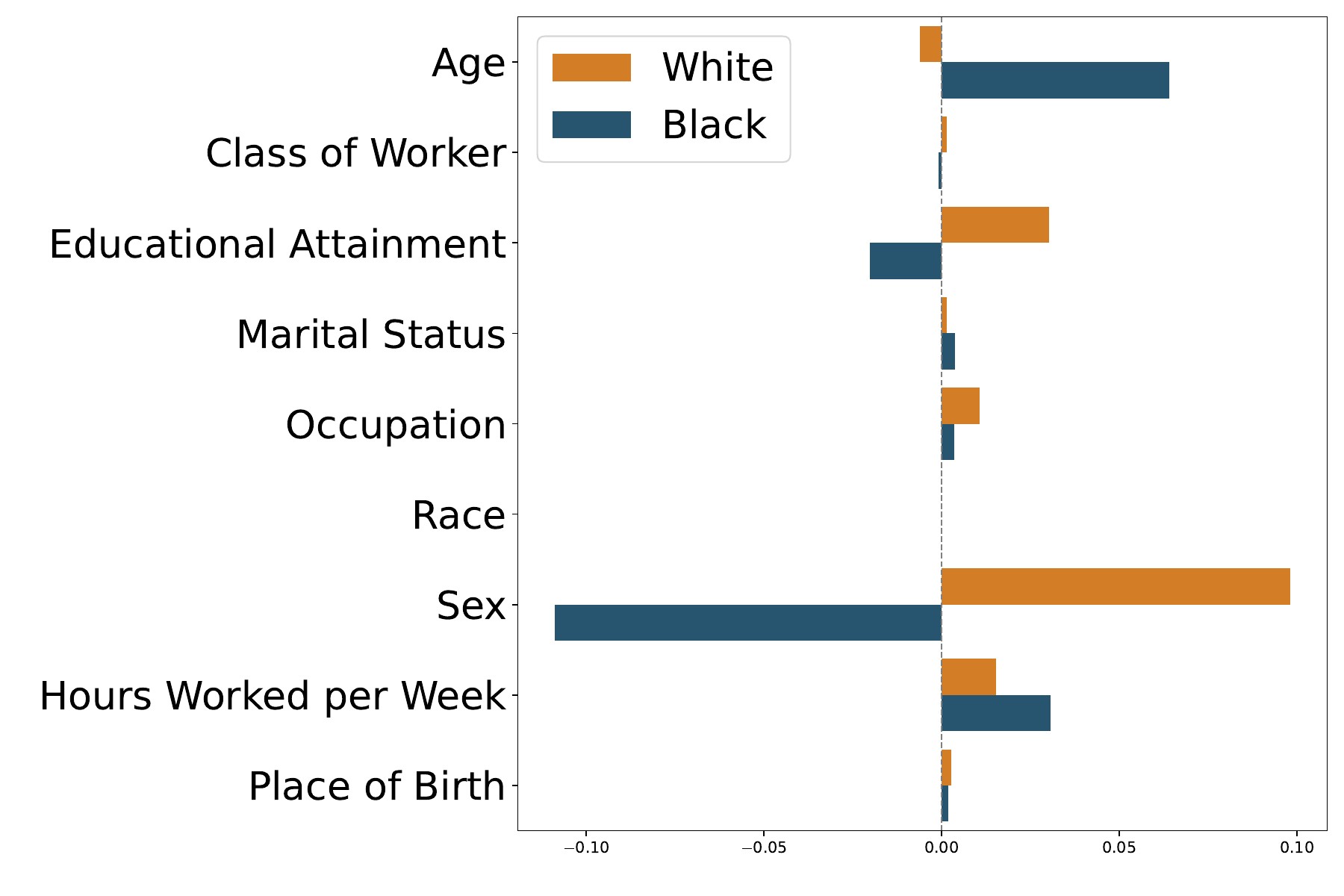}   
    \end{subfigure}
    \begin{subfigure}[b]{0.32\textwidth}
        \includegraphics[width=\linewidth]{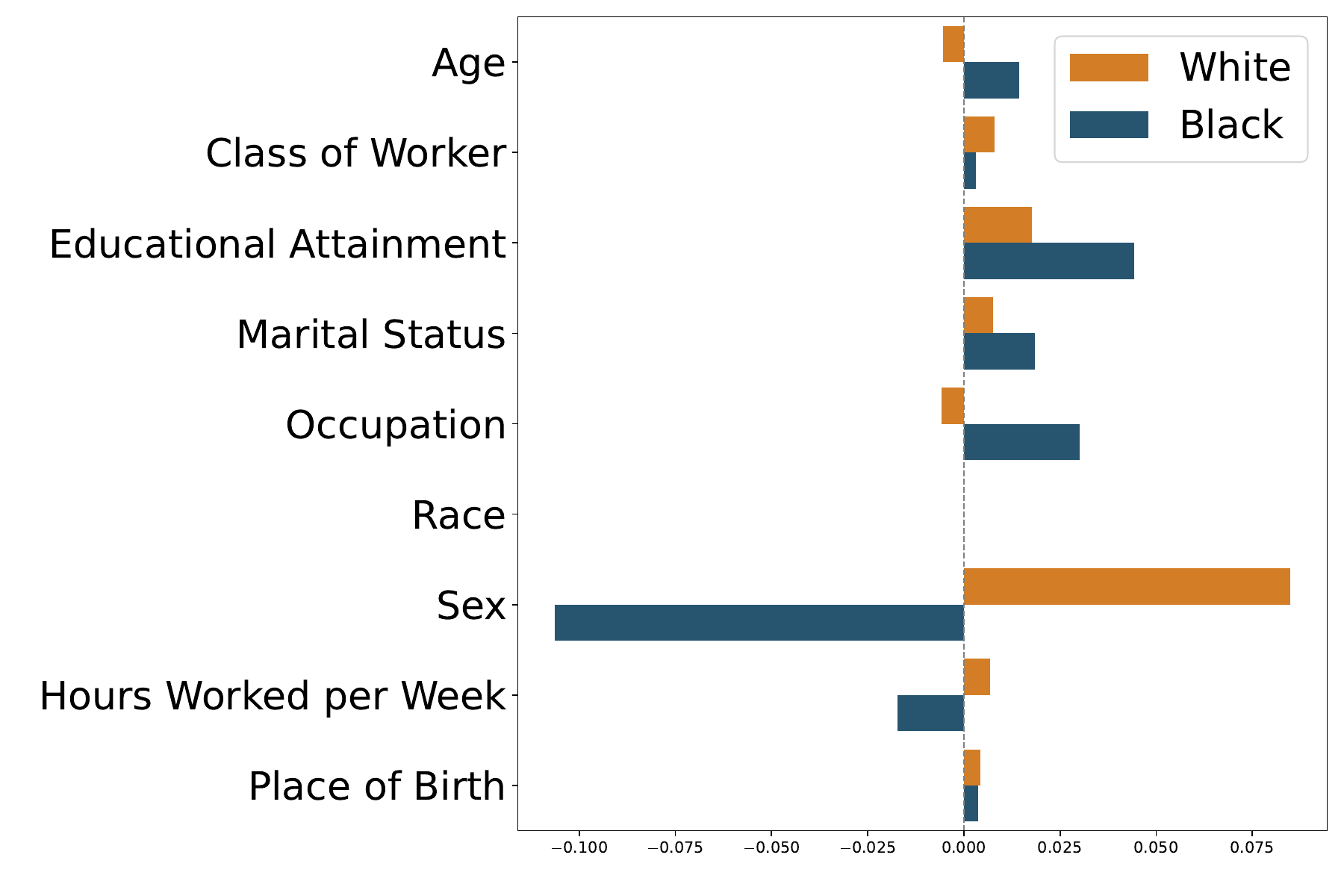}
    \end{subfigure}
    \caption{Differences of mean contributions with LIME when the protected attribute race is removed for the Adult, AdultCA, and AdultLA
datasets. Rows represent the datasets, while columns correspond to P, TP, and FP.}
 \label{fig:diff_LIME_race}
\end{figure}

\subsection{Generation of correlation matrices}
To generate the correlation matrices shown in Fig.~\ref{fig:cor_matrices}, we adopt the approach described in \cite{le2022survey}, which involves discretizing continuous features and grouping categorical features. Specifically, for the Adult dataset, we apply the following transformations: Age = \{25–60, <25 or >60\}, Hours per week = \{<40, 40–60, >60\}, Workclass = \{private, non-private\}, Education = \{high, low\}, Country = \{US, non-US\}, Race = \{White, Non-White\}, Marital Status = \{Married, Other\}, and Occupation = \{office, heavy-work, other\}. A similar approach is used for the AdultCA and AdultLA datasets, with the occupation feature grouped according to the codes provided in \cite{ding2021retiring}. In the resulting categorical space, categorical correlations between feature pairs are calculated using Cramer's V correlation \cite{cramer1946mathematical}.

\section{Appendix}
\begin{table}[H]
  \centering
  \caption{COMPAS Dataset description with features, feature descriptions, and feature types.}
  \label{tab:compas_features}
  \begin{tabular}{l l l l}
    \toprule
    Dataset & Feature & Description & Type \\
    \midrule
    \multirow{9}{*}{COMPAS} & Age & Age of defendant & Numeric \\
                             & Race & Race of defendant & Categorical \\
                             & Sex & Sex of defendant & Categorical \\
                             & JuvFelCount & Juvenile felony count & Numeric \\
                             & JuvMisdCount & Juvenile misdemeanor count & Numeric \\
                             & JuvOtherCount & Juvenile other offenses count & Numeric \\
                             & PriorsCount & Prior offenses count & Numeric \\
                             & ChargeDegree & Charge degree of original crime & Categorical \\
                             & TwoYearRecid & Whether the defendant is rearrested & Binary \\
    \bottomrule
  \end{tabular}
\end{table}
\subsection{Experiments with additional dataset}
In this section, we present additional experiments using the COMPAS recidivism dataset. The COMPAS \footnote{\href{https://raw.githubusercontent.com/propublica/compas-analysis/bafff5da3f2e45eca6c2d5055faad269defd135a/compas-scores-two-years.csv}{COMPAS}} dataset, which has released by ProPublica in 2016, contains instances from the criminal justice system and is used to predict he likelihood of recidivism. Table \ref{tab:compas_features} presents the features of COMPAS dataset, including descriptions and feature types.

\begin{figure}[H]
    \centering    
        \includegraphics[width=0.6\linewidth]
        {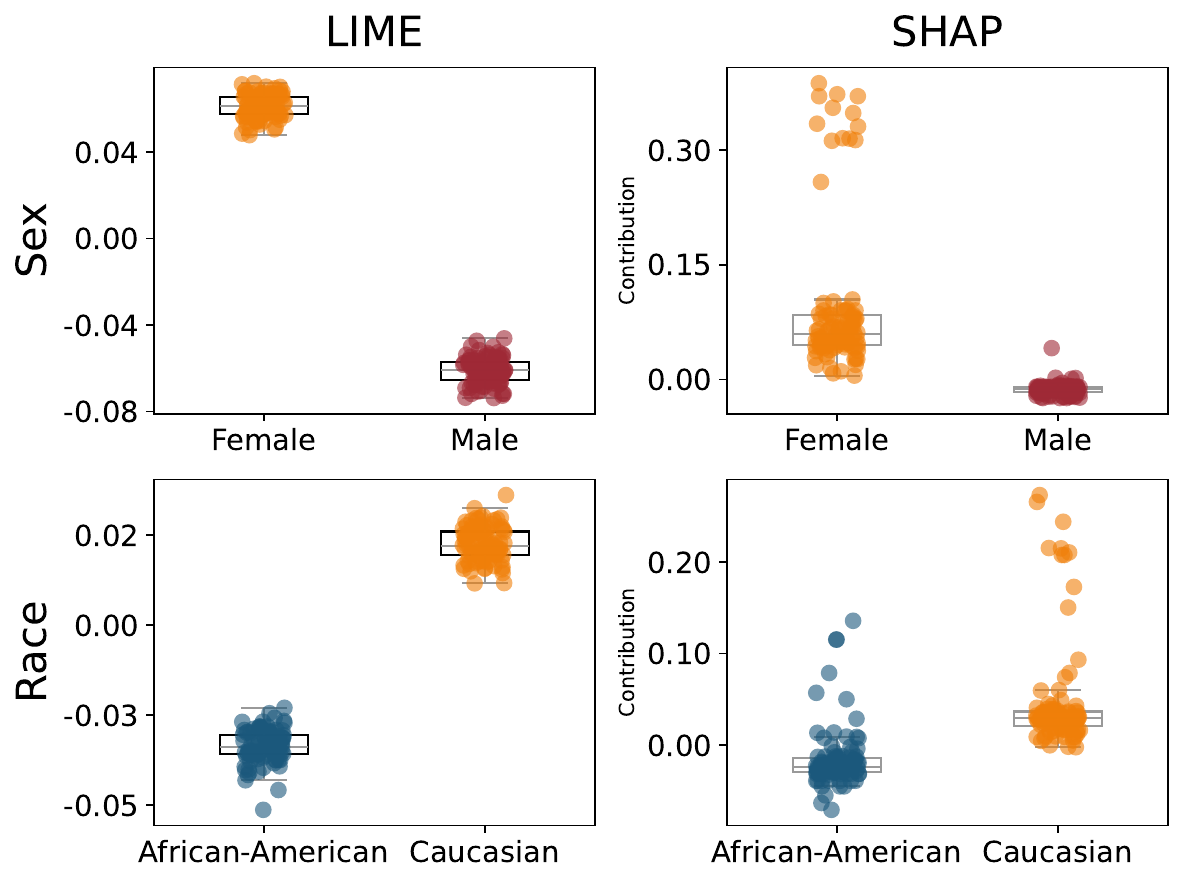} 
        \caption{LIME and SHAP feature contributions for sex and race for Compas dataset.}
        \label{fig:contributions_distributions_compas}
\end{figure}

\begin{figure}[H]
    \centering    
        \includegraphics[width=0.9\linewidth]{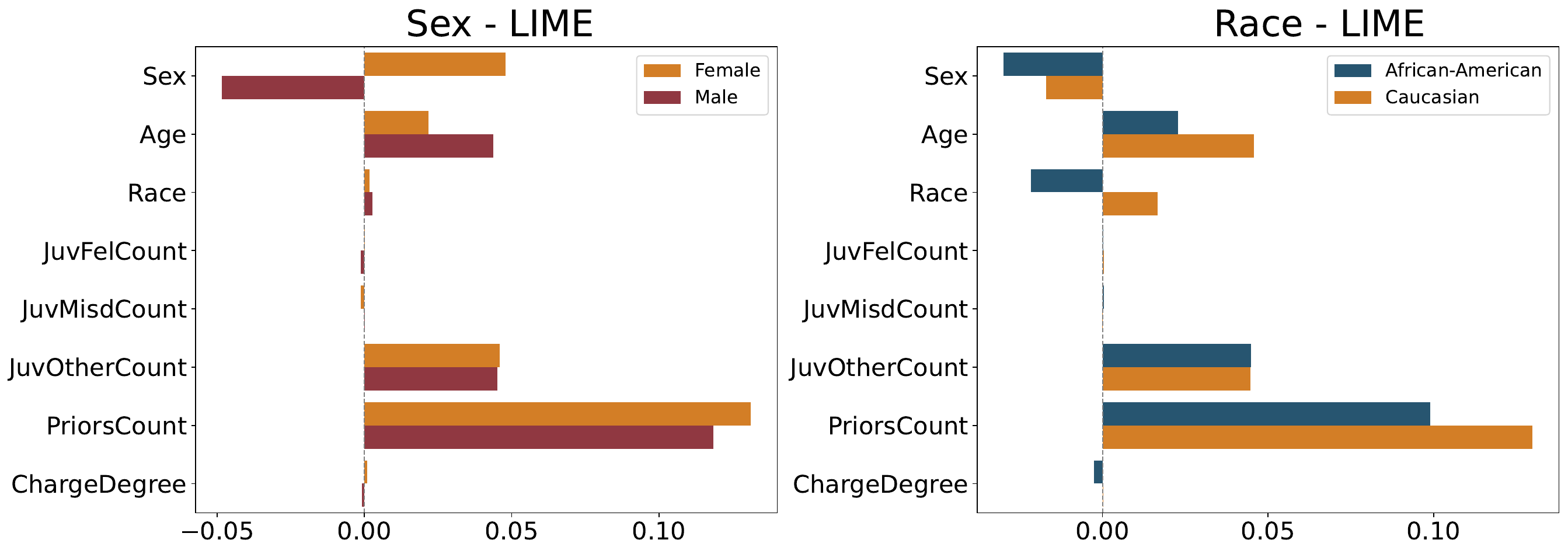} 
        \caption{LIME mean feature contributions for sex and race for Compas dataset.}
        \label{label:mean_contr_LIME_compas}
\end{figure}

\begin{figure}[H]
    \centering    
        \includegraphics[width=0.9\linewidth]{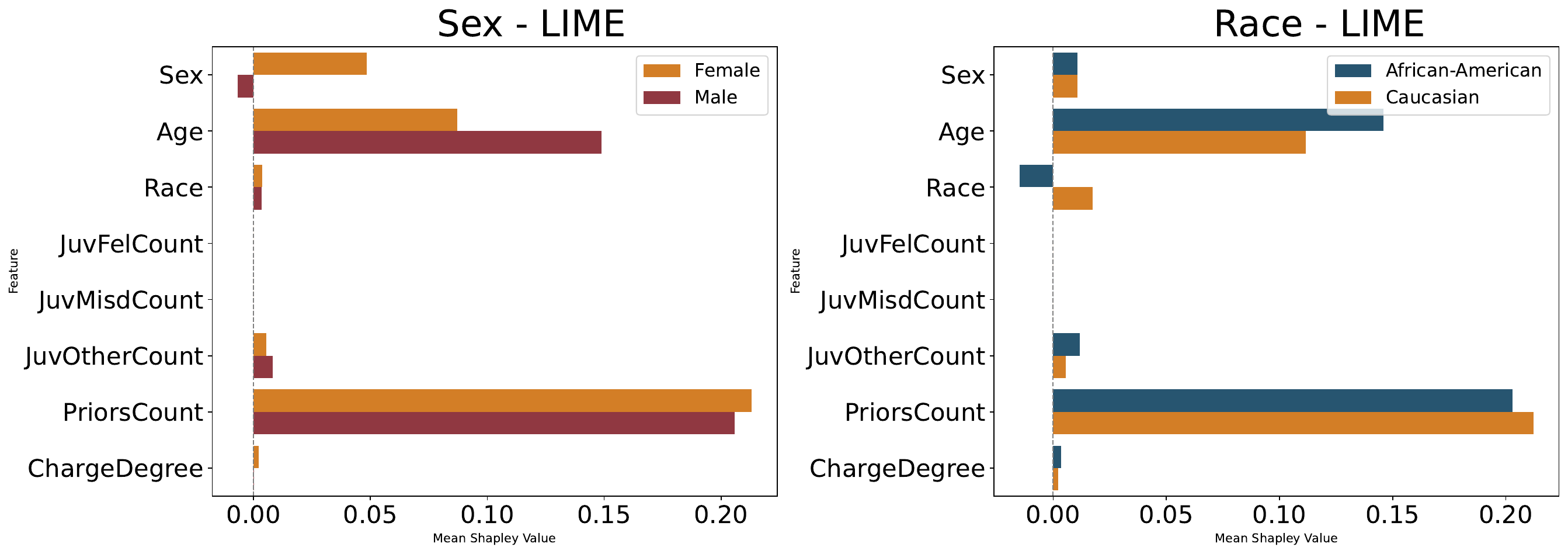} 
        \caption{SHAP mean feature contributions for sex and race for Compas dataset.}
        \label{label:mean_contr_SHAP_compas}
\end{figure}

Table \ref{tab:fairness_diff_compas} shows differences in PR, TPR, and FPR between male and female groups and Caucasian and African-American groups, with statistical z-tests for significance. The results indicate violations of all distributive fairness metrics, as evidenced by significant differences with high absolute z-test values. The negative sign in the gender comparison reflects bias against males, whereas in the racial comparison, there is bias against African-Americans. In Fig. ~\ref{fig:contributions_distributions_compas} we can look at procedural fairness by studying the contribution of the protected attribute to the positive class with LIME and SHAP methods. We observe results consistent with Fig. ~\ref{fig:contributions_distributions}, where the disadvantaged group, as defined by distributive fairness, consistently exhibits negative contributions, while the group benefiting from the bias shows positive contributions.
\begin{table}[H]
  \centering
  \small
  \caption{Differences in PR, TPR, and FPR for \texttt{sex} (male-female) and \texttt{race} (Caucasian-African American) in the COMPAS dataset, including statistical significance scores. Higher values indicate worse disparities.}
  \label{tab:fairness_diff_compas}
  \begin{tabular}{lcc}
    \toprule
    Metric & Gender & Race \\
    \midrule
    PR     & -0.22 (-6.37) & 0.325 (10.49)  \\
    TPR    & -0.162 (-4.09)  & 0.235 (6.3)  \\
    FPR    & -0.244 (-4.5) & 0.359 (7.84)   \\
    \midrule
    Accuracy & \multicolumn{2}{c}{0.68} \\
    \bottomrule
  \end{tabular}
\end{table}
In Figures \ref{label:mean_contr_LIME_compas}, \ref{label:mean_contr_SHAP_compas} we can see the features contributions using LIME and SHAP methods. We observe that among the top significant features are features not directly related to the task, such as \texttt{sex}, \texttt{age} and \texttt{race}. 

\begin{figure}[H]
    \centering
    \begin{subfigure}{0.49\textwidth}
        \centering
        \includegraphics[width=0.9\linewidth]{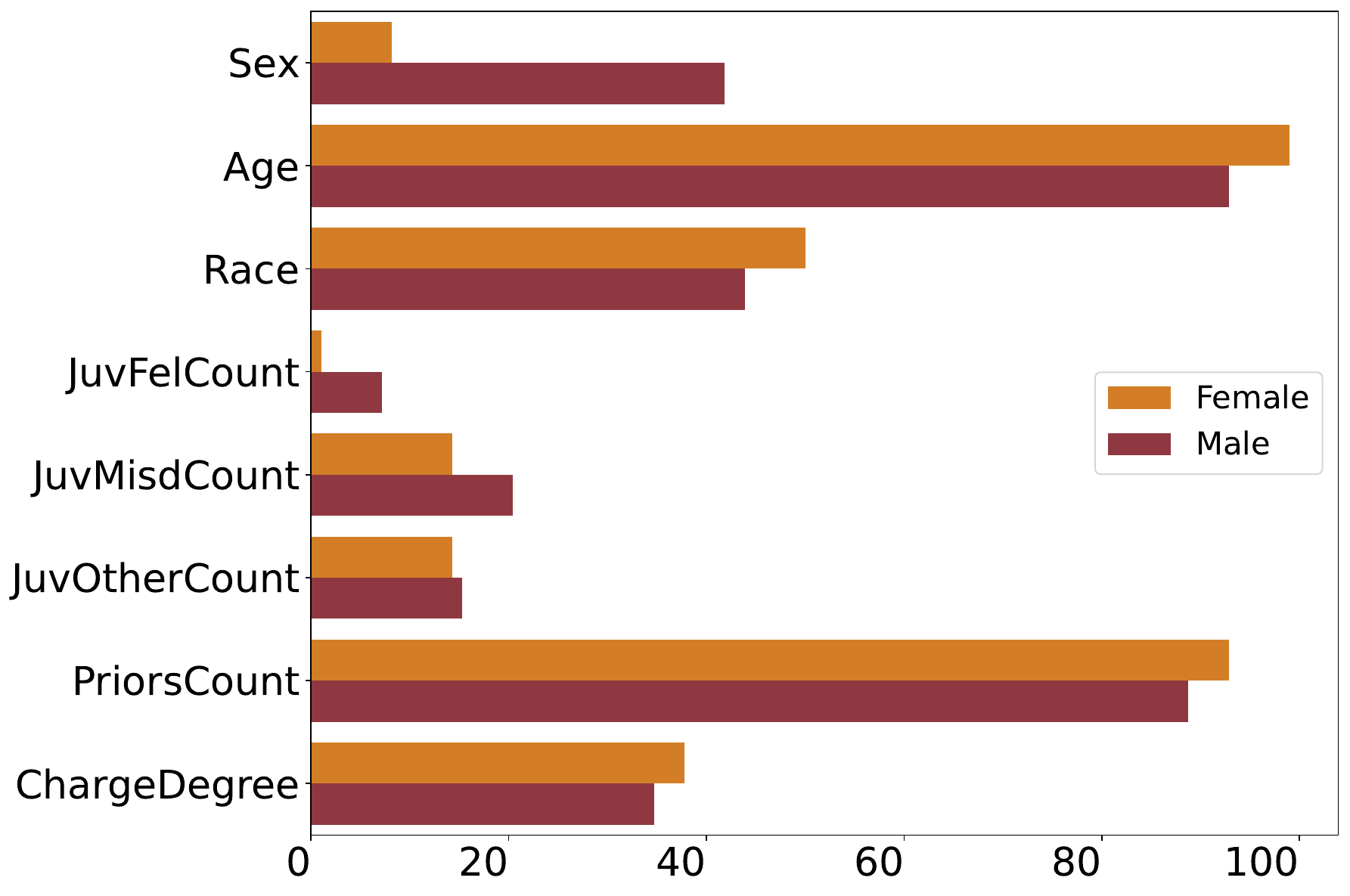} 
    \end{subfigure}
    \hspace{0.005cm}
    \begin{subfigure}{0.49\textwidth}
        \centering
        \includegraphics[width=0.9\linewidth]{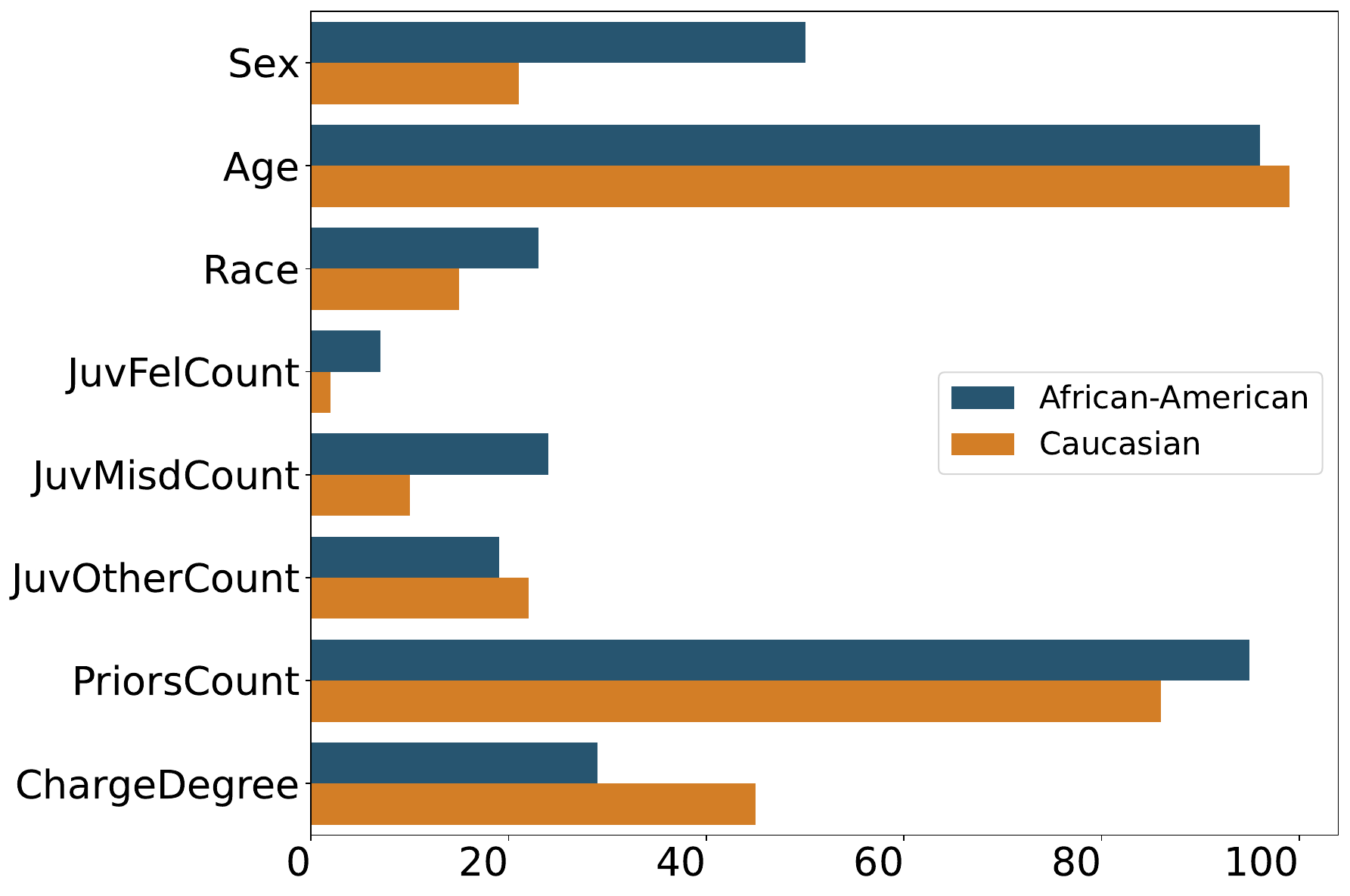} 
    \end{subfigure}
    \caption{ Percentage of feature changes from the DiCE method per group for the Compas dataset.}
    \label{fig:DICE_compas}
\end{figure}

In ~Fig \ref{fig:DICE_compas} we present the results for the DiCE explanation method. We observe that the disadvantaged groups, in this case males and African-Americans, must change their sex and race, respectively, to achieve the desired outcome.

\subsection{Experiments with additional model}
In this section, we present additional experiments using the XGBoost model on all our datasets.

\begin{figure}[H]
    \centering    
        \includegraphics[width=0.8\linewidth]{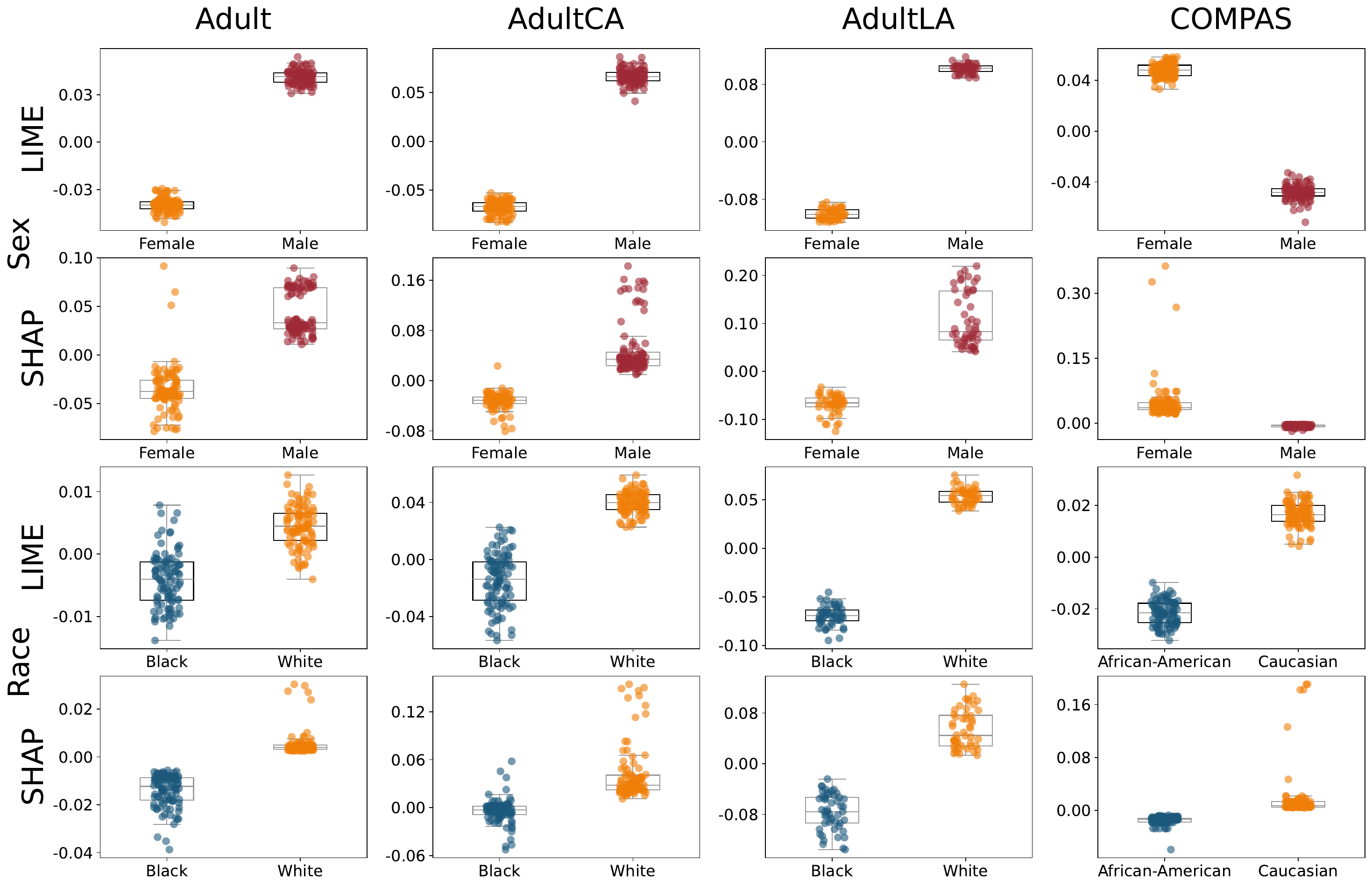} 
        \caption{LIME and SHAP feature contributions for \texttt{sex} and \texttt{race} across datasets for XGB model.}
        \label{fig:contributions_distributions_xgb}
\end{figure}

Table \ref{tab:fairness_diff_sex_race_xgb} shows the differences in PR, TPR, and FPR between male and female groups and White and Black groups, with statistical z-tests for significance when using the XGboost model.
~Fig. \ref{fig:contributions_distributions_xgb} demonstrates comparable results to those observed with the Random Forest model, further validating our hypothesis regarding the relationship between distributive and procedural fairness.

\begin{figure}[h]
    \centering    
        \includegraphics[width=\linewidth]{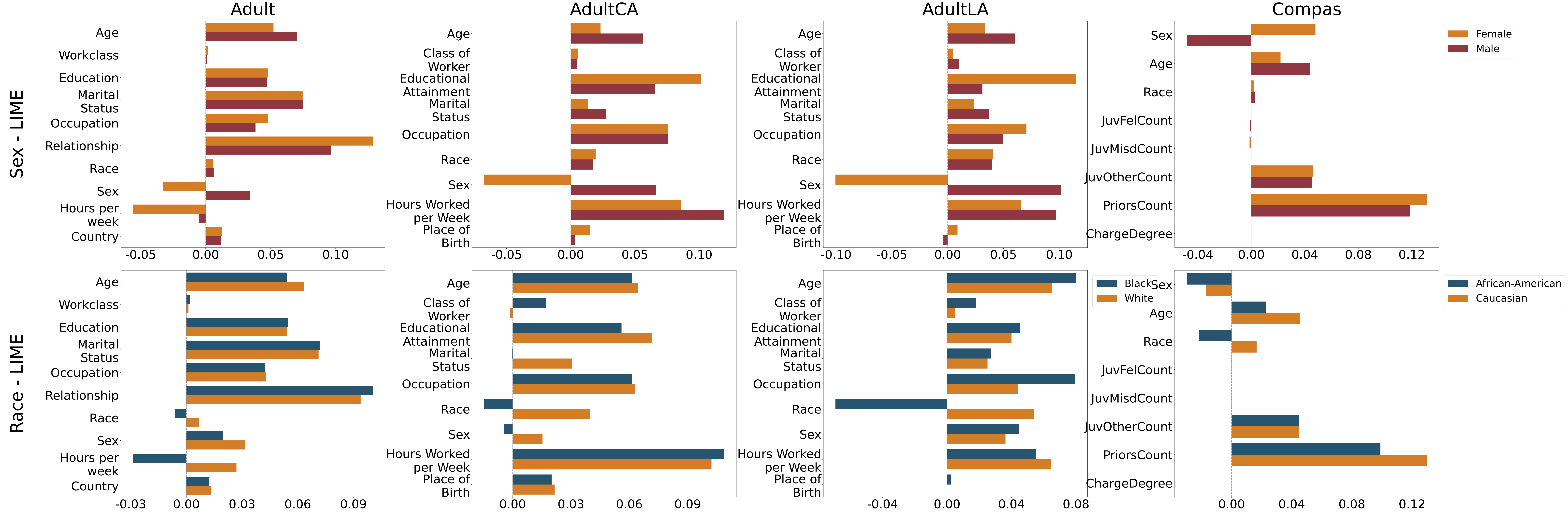} 
        \caption{LIME mean feature contributions for sex and race across datasets for XGB model.}
        \label{label:mean_contr_LIME_xgb}
\end{figure}

\begin{figure}[h]
    \centering    
        \includegraphics[width=\linewidth]{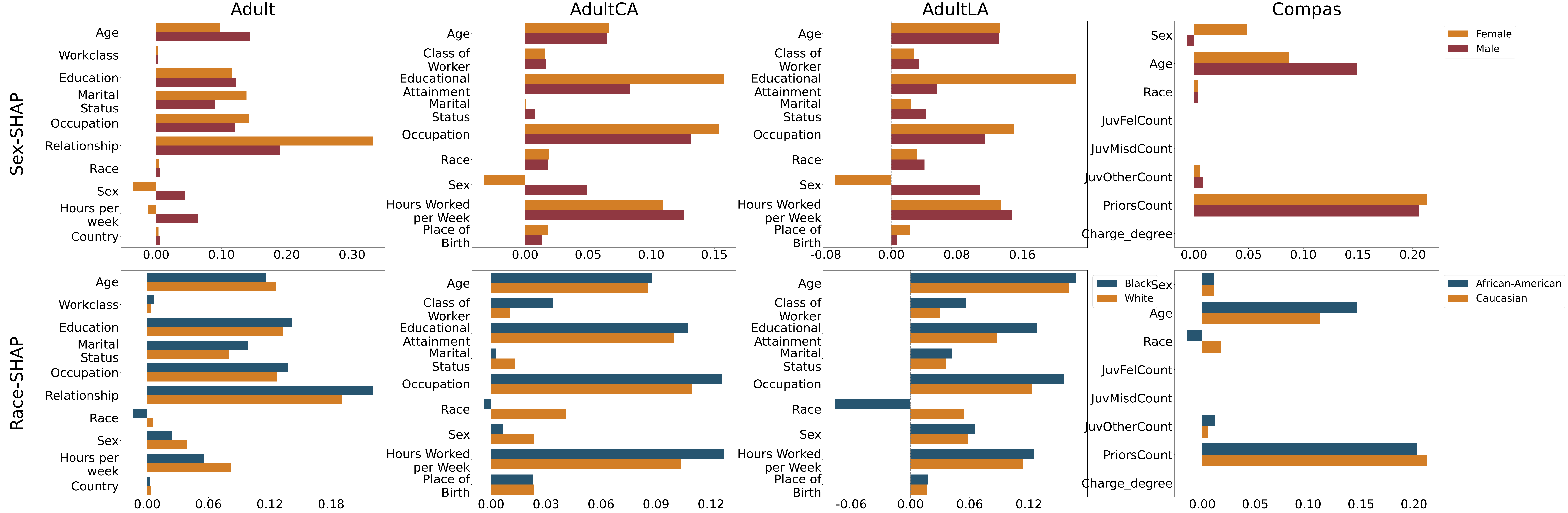} 
        \caption{SHAP mean feature contributions for sex and race across datasets for XGB model.}
        \label{label:mean_contr_SHAP}
\end{figure}

\begin{figure}[h]
    \centering
    \begin{minipage}{0.24\textwidth}
        \centering
        \includegraphics[width=\linewidth]{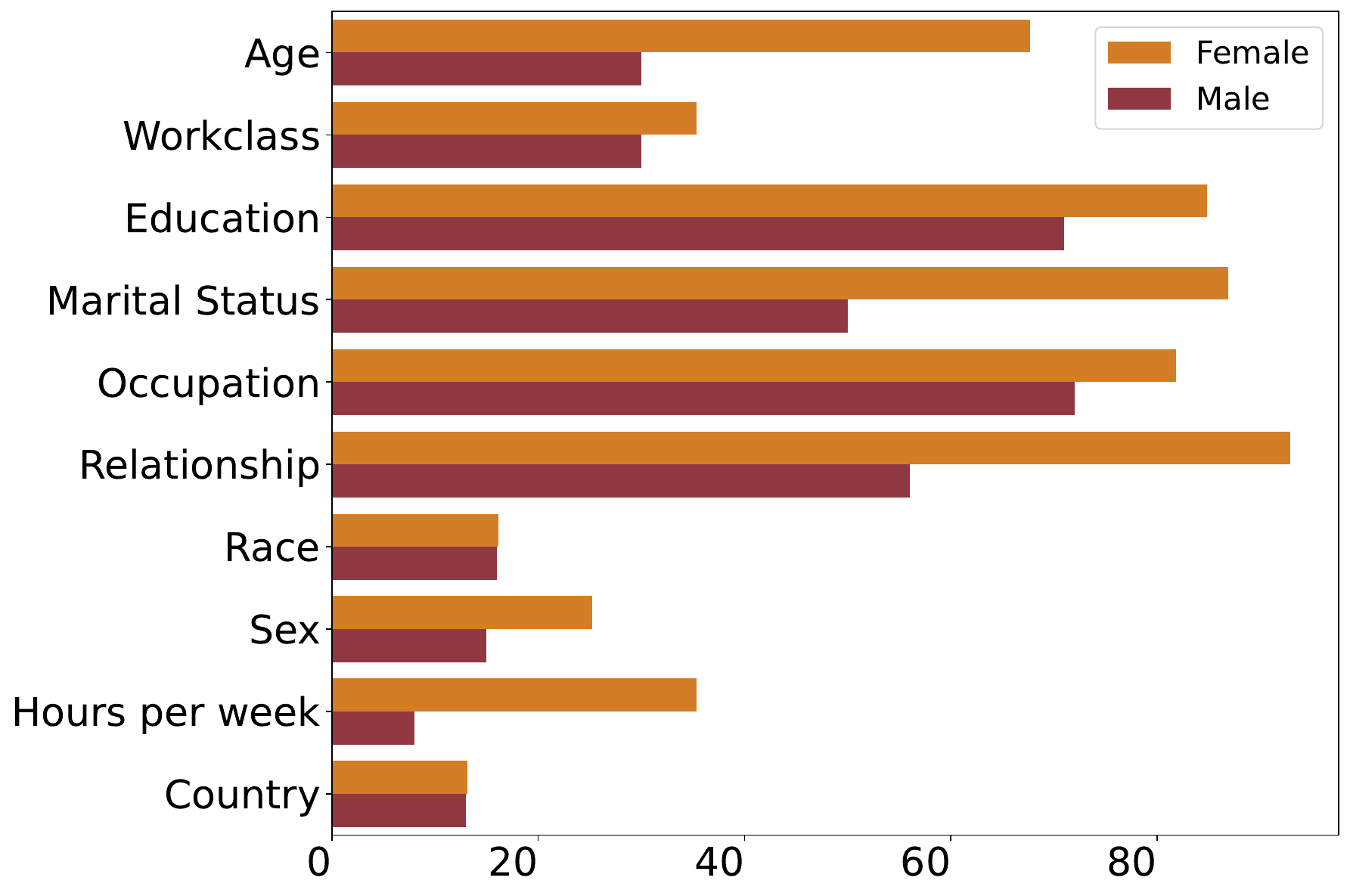}
    \end{minipage}
    \hfill
    \begin{minipage}{0.24\textwidth}
        \centering
        \includegraphics[width=\linewidth]{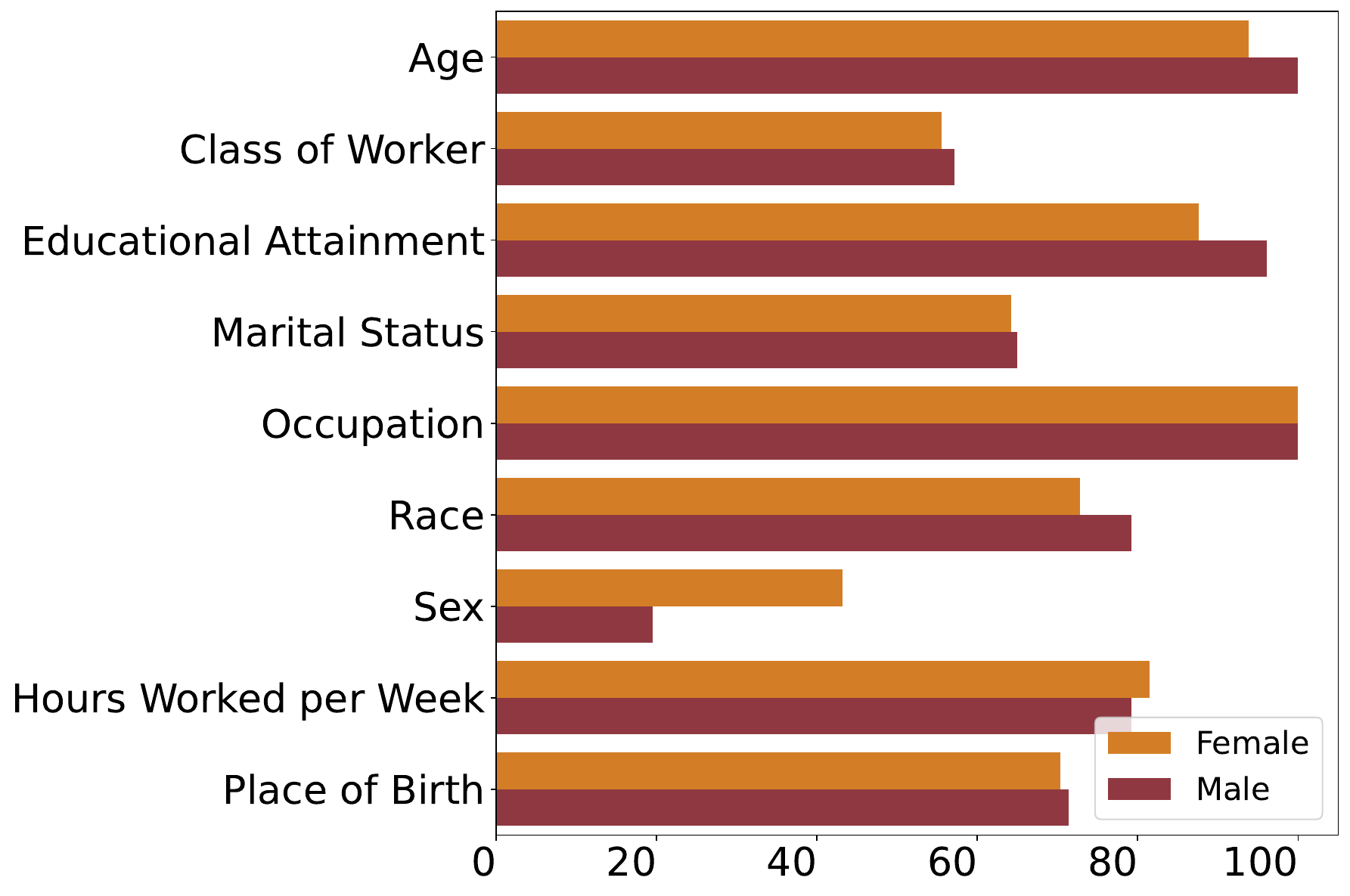}      
    \end{minipage}
    \hfill
    \begin{minipage}{0.24\textwidth}
        \centering
        \includegraphics[width=\linewidth]{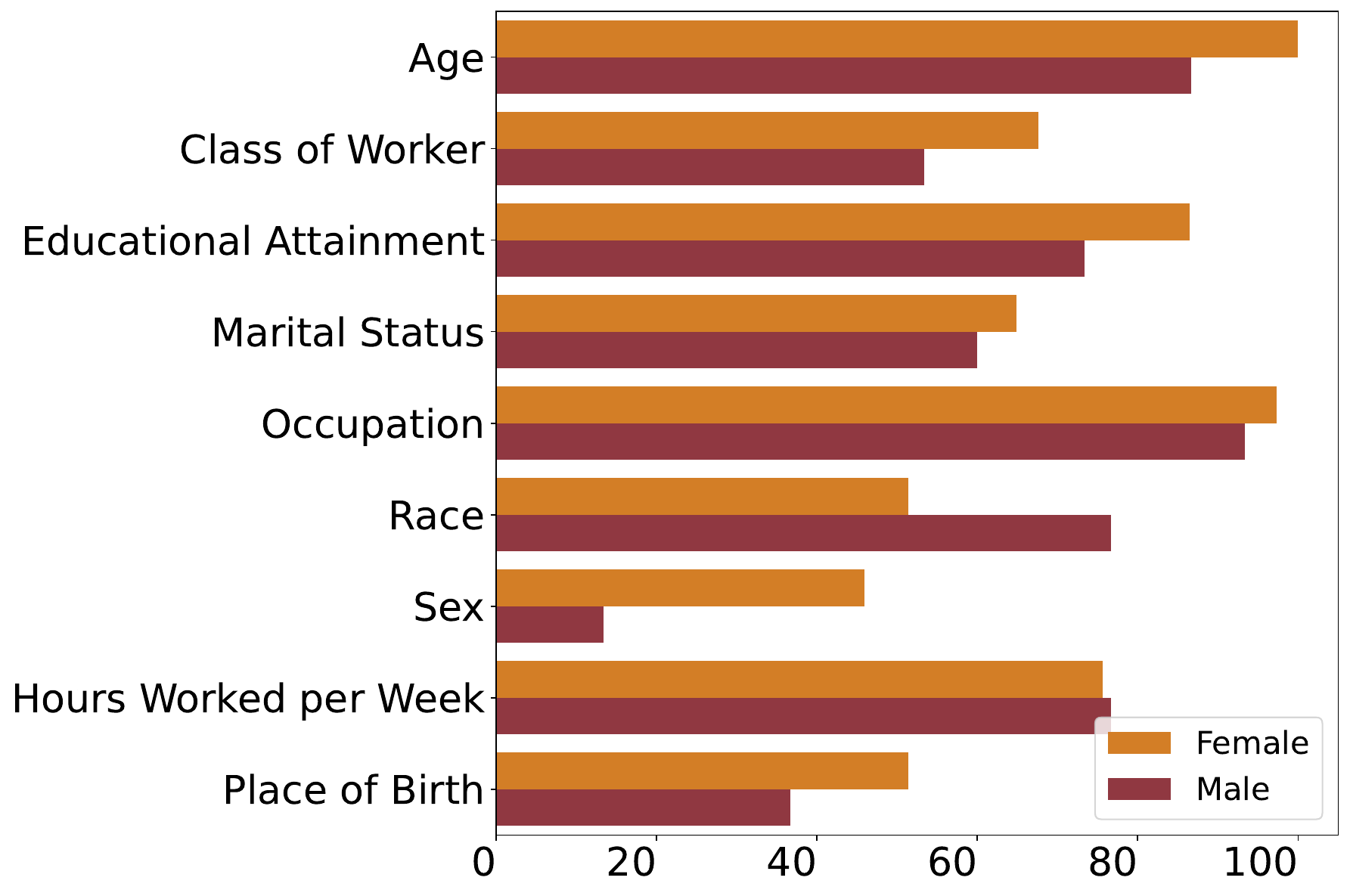}
    \end{minipage}
    \hfill
    \begin{minipage}{0.24\textwidth}
        \centering
        \includegraphics[width=\linewidth]{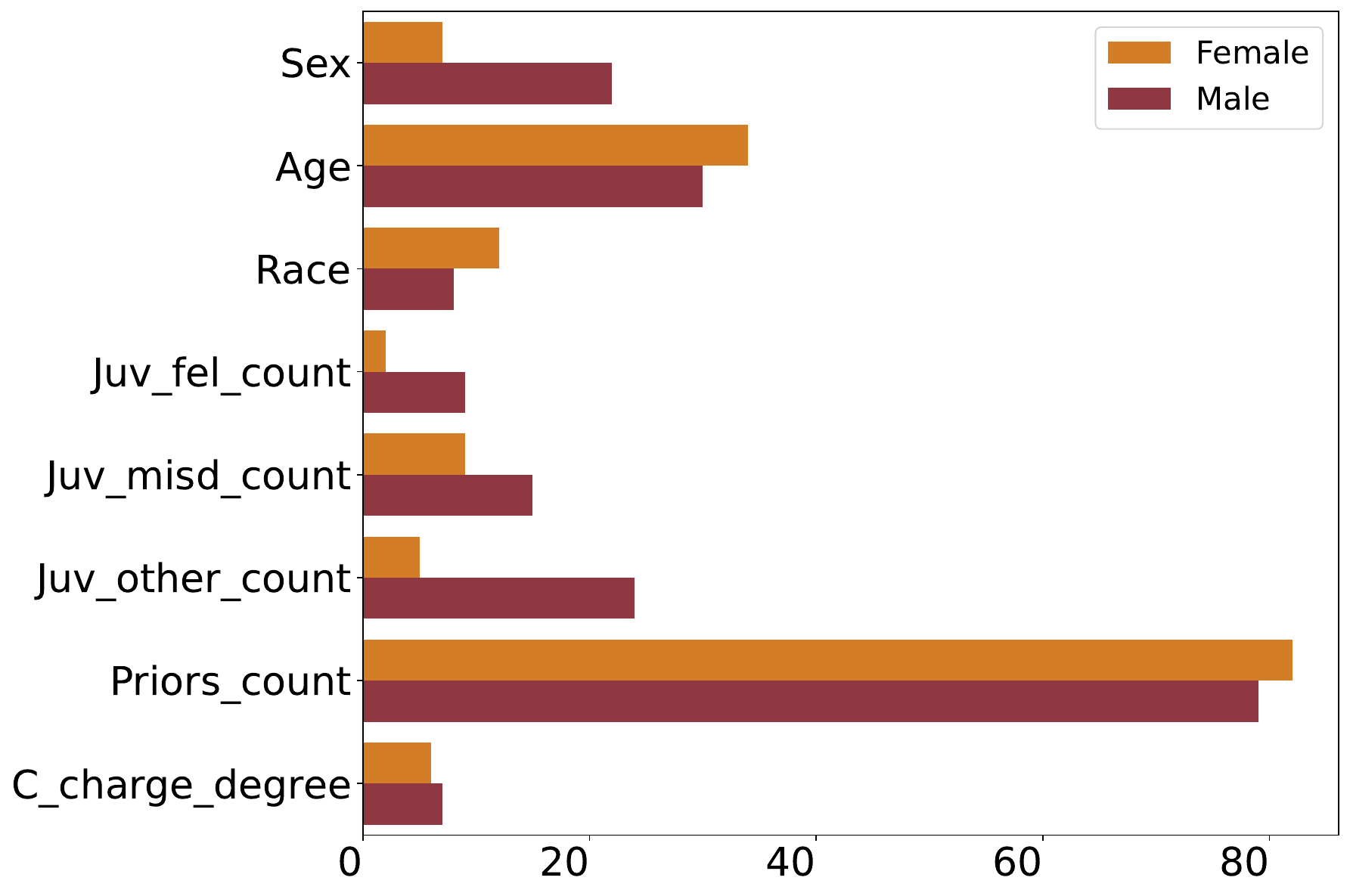}
    \end{minipage}
    
    \vspace{0.5em} 
    
    \begin{minipage}{0.24\textwidth}
        \centering
        \includegraphics[width=\linewidth]{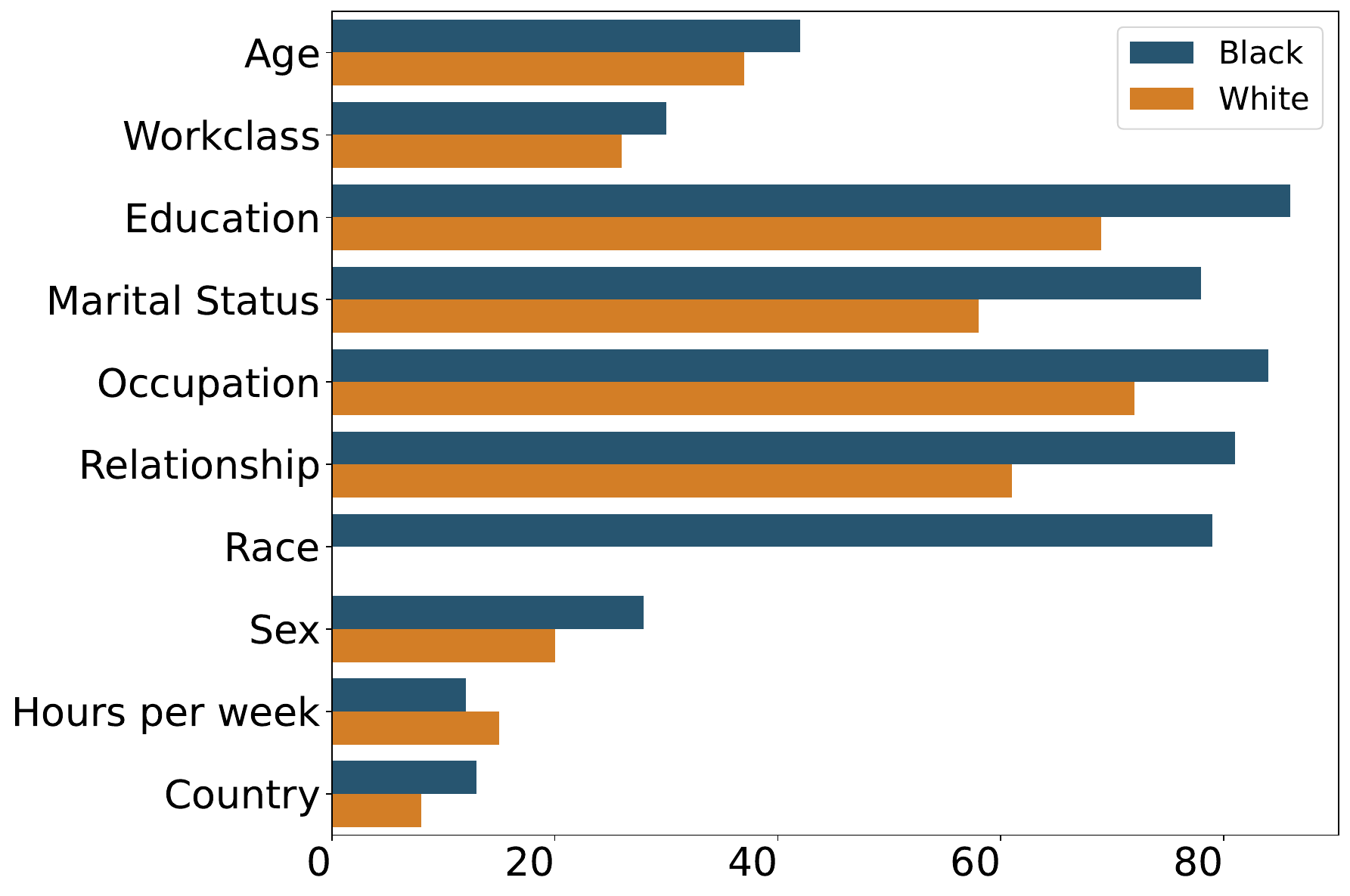}
    \end{minipage}
    \hfill
    \begin{minipage}{0.24\textwidth}
        \centering
        \includegraphics[width=\linewidth]{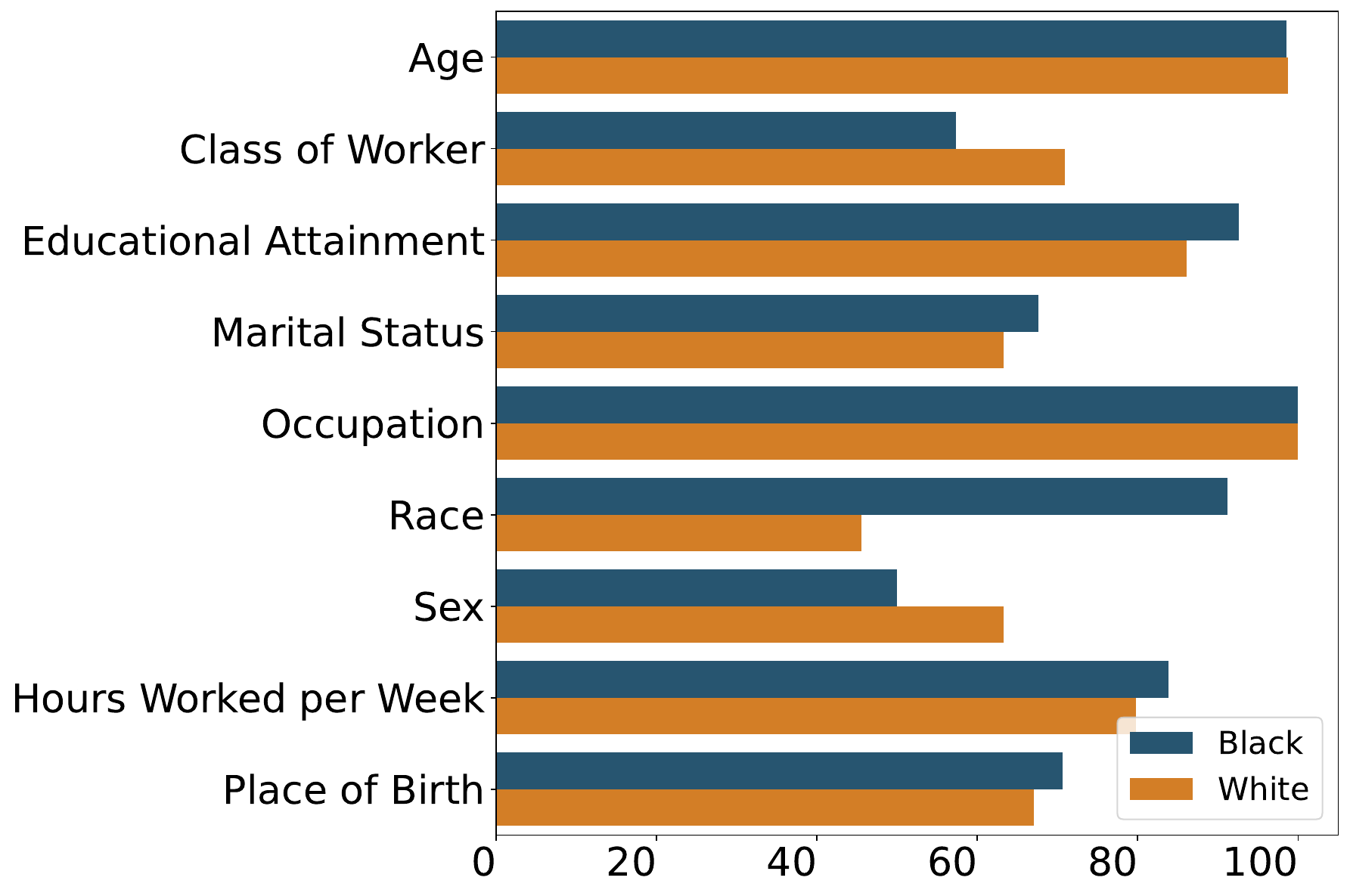} 
    \end{minipage}
    \hfill
    \begin{minipage}{0.24\textwidth}
        \centering
        \includegraphics[width=\linewidth]{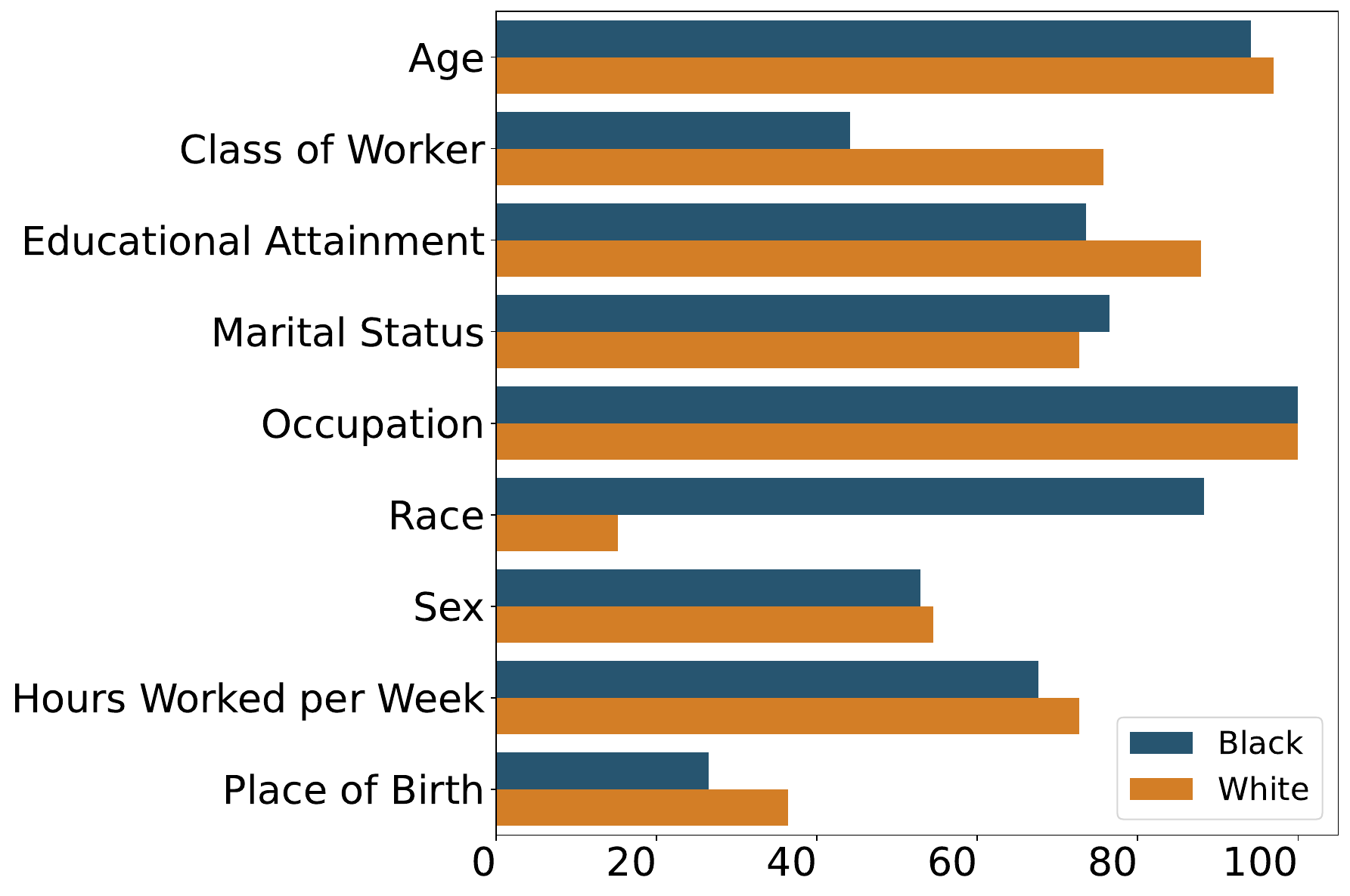}
    \end{minipage}
    \hfill
    \begin{minipage}{0.24\textwidth}
        \centering
        \includegraphics[width=\linewidth]{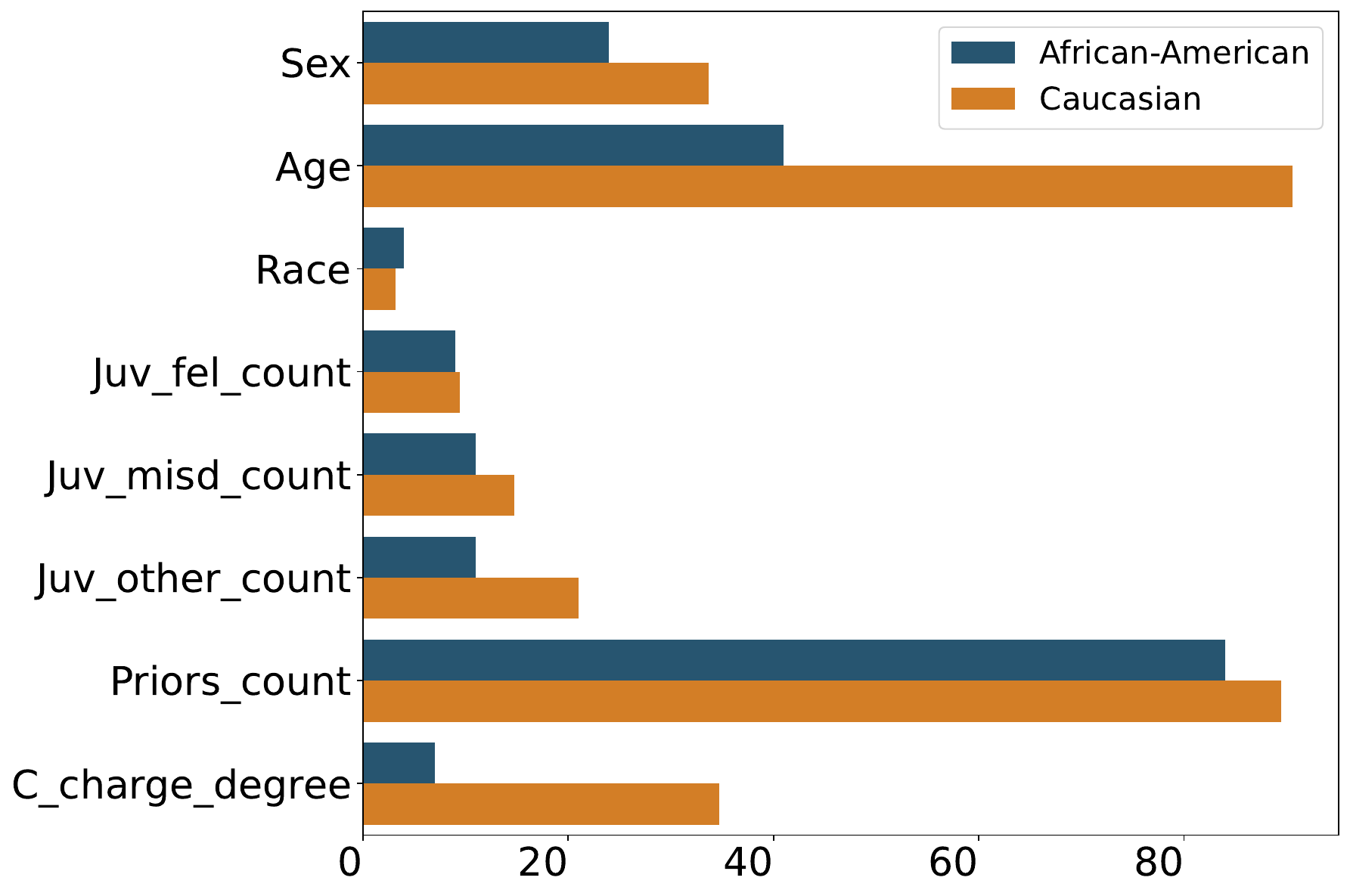}
    \end{minipage}
    
    \caption{Percentage of feature changes from the DiCE method per group across datasets for XGB model.}
    \label{fig:DICE_xgb}
\end{figure}

Figures \ref{label:mean_contr_LIME_xgb}, \ref{label:mean_contr_SHAP} show the features contributions for LIME and SHAP methods respectively, while ~Fig. \ref{fig:DICE_xgb} presents feature changes in case of the DiCE method. Aside from the attribution, the conclusions remain consistent with our previous findings.
\begin{table}[H]
  \centering
  \small
  \caption{Differences in PR, TPR, and FPR for \texttt{sex} (male-female) and \texttt{race} (White-Black) across datasets using XGB model, incl. statistical significance scores. Higher values indicate worse disparities.}
\label{tab:fairness_diff_sex_race_xgb}
  \begin{tabular}{lcccccccc}
    \toprule
    Metric & \multicolumn{2}{c}{Adult} & \multicolumn{2}{c}{AdultCA} & \multicolumn{2}{c}{AdultLA} & \multicolumn{2}{c}{Compas} \\
    \cmidrule(r){2-3} \cmidrule(r){4-5} \cmidrule(r){6-7} \cmidrule(r){8-9}
           & Gender & Race & Gender & Race & Gender & Race & Gender & Race \\
    \midrule
    PR     & 0.196 (18.9) & 0.142 (8.57)  & 0.104 (21.07) & 0.162 (12.9) & 0.264 (17.39) & 0.29 (15.0) & -0.2 (-5.76) & 0.283 (9.14) \\
    TPR    & 0.191 (5.47) & 0.125 (2.07)  & 0.034 (6.47) & 0.091 (7.53)  & 0.175 (8.23) & 0.233 (7.5)  & -0.122 (-3.11) & 0.2 (5.39) \\
    FPR    & 0.094 (12.16) & 0.063 (5.37) & 0.058 (10.47) & 0.087 (5.35)  & 0.152 (9.68) & 0.157 (8.47)   & -0.245 (-4.57) & 0.303 (6.69)  \\
    \midrule
    Accuracy & \multicolumn{2}{c}{0.83} & \multicolumn{2}{c}{0.81} & \multicolumn{2}{c}{0.79} & \multicolumn{2}{c}{0.69} \\
    \bottomrule
  \end{tabular}
\end{table}

\end{document}